\documentclass[runningheads,a4paper]{llncs}
\usepackage[top=2.35cm, bottom=2.35cm, left=2.5cm, right=2.5cm]{geometry}
\usepackage{amsmath}
\usepackage{amssymb}
\usepackage[normalem]{ulem}
\usepackage{graphicx}
\usepackage{booktabs}
\usepackage{listings}
\usepackage[unicode=true, bookmarks=true, breaklinks=false, pdfborder={0 0 1}, backref=false, colorlinks=false]{hyperref}
\hypersetup{
    linkbordercolor={1 1 1}, % White border color (effectively invisible)
}

\usepackage{xcolor}

\AtBeginDocument{%
    \hypersetup{
        citebordercolor={1 0 0} % Blue border for citations, adjust to your preference
    }
}
\let\oldhyperref\hyperref
\renewcommand{\hyperref}[2][]{\oldhyperref[#1]{\textbf{#2}}}

\usepackage{caption}
\usepackage{subcaption}

\usepackage{url}
\usepackage{color}
\usepackage{epsfig}
\usepackage{float}
\usepackage{bm}
\usepackage[backend=biber, style=numeric, sorting=none, defernumbers=true, maxnames=99]{biblatex}  
\AtEveryBibitem{\clearfield{url}}
\makeatletter
\renewcommand{\paragraph}{%
  \@startsection{paragraph}{4}%
  {\z@}{3.25ex \@plus 1ex \@minus .2ex}{-1em}%
  {\normalfont\normalsize\bfseries}%
}
\makeatother

\usepackage[nomargin,inline,marginclue,draft]{fixme}
\usepackage{cleveref}
\fxsetup{status=draft}
\fxsetup{theme=color, mode=multiuser} \FXRegisterAuthor{me}{ame}{\color{red}Me}

\title{Generalist foundation models from a multimodal dataset \\for 3D computed tomography}

\titlerunning{CT-RATE, CT-CLIP and CT-CHAT}

\author{ 
    {\small Ibrahim Ethem Hamamci\inst{1,2,3,\dag,*}}
    \and
    {\small Sezgin Er\inst{1,3,\dag}}
    \and
    {\small Chenyu Wang\inst{4}}    
    \and
    {\small Furkan Almas\inst{3}}
    \and
    {\small Ayse Gulnihan Simsek\inst{3}}
    \and
    {\small Sevval Nil Esirgun\inst{3}}
    \and
    {\small Irem Dogan\inst{3}}
    \and
    {\small Omer Faruk Durugol\inst{3}}
    \and
    {\small Benjamin Hou\inst{5}}
    \and
    {\small Suprosanna Shit\inst{1,2}}
    \and
    {\small Weicheng Dai\inst{4}}
    \and
    {\small Murong Xu\inst{1,2}}
    \and
    {\small Hadrien Reynaud\inst{6}}
    \and
    {\small Muhammed Furkan Dasdelen\inst{3}}
    \and    
    {\small Bastian Wittmann\inst{1,2}}
    \and
    {\small Tamaz Amiranashvili\inst{1,2}}
    \and
    {\small Enis Simsar\inst{7}}
    \and
    {\small Mehmet Simsar\inst{3}}
    \and
    {\small Emine Bensu Erdemir\inst{3}}
    \and
    {\small Abdullah Alanbay\inst{3}}
    \and
    {\small Anjany Sekuboyina\inst{1,2}}
    \and
    {\small Berkan Lafci\inst{1,2}}
    \and
    {\small Ahmet Kaplan\inst{3}}
    \and
    {\small Zhiyong Lu\inst{4}}
    \and
    {\small Malgorzata Polacin\inst{8}}
    \and
    {\small Bernhard Kainz\inst{6,9}}
    \and
    {\small Christian Bluethgen\inst{8}}
    \and
    {\small Kayhan Batmanghelich\inst{4}}
    \and
    {\small Mehmet Kemal Ozdemir\inst{3}}
    \and
    {\small Bjoern Menze\inst{1,2}}
}
 \authorrunning{Hamamci et al.} 
    
    \institute{{$^{1}$ Department of Quantitative Biomedicine, University of Zurich, Zurich, Switzerland \\ \quad $^{2}$ ETH AI Center, ETH Zurich, Zurich, Switzerland \\ \quad $^{3}$ International School of Medicine, Istanbul Medipol University, Istanbul, Turkey \\ \quad $^{4}$ Department of Electrical and Computer Engineering, Boston University, MA, USA \\ \quad $^{5}$ Division of Intramural Research, National Institutes of Health, MD, USA \\ \quad $^{6}$ Department of Computing, Imperial College London, London, UK \\
\quad $^{7}$ Department of Computer Science, ETH Zurich, Zurich, Switzerland \\ \quad $^{8}$ Institute for Diagnostic and Interventional Radiology, University Hospital Zurich, Zurich, Switzerland \\ \quad $^{9}$ Department Artificial Intelligence in Biomedical Engineering, FAU Erlangen-Nürnberg, Erlangen, Germany}
    ~
    \\
    ~
        \\
\small{\dag}These authors contributed equally to this work.
\small{*}Corresponding author: \email{\{ibrahim.hamamci@uzh.ch\}}}

\begin{document}
    \mainmatter
    \maketitle

\setcounter{footnote}{0} 
\begin{abstract}

Advancements in medical imaging AI, particularly in 3D imaging, have been limited due to the scarcity of comprehensive datasets. We introduce CT-RATE, a public dataset that pairs 3D medical images with corresponding textual reports. CT-RATE comprises 25,692 non-contrast 3D chest CT scans from 21,304 unique patients. Each scan is accompanied by its corresponding radiology report. Leveraging CT-RATE, we develop CT-CLIP, a CT-focused contrastive language-image pretraining framework designed for broad applications without the need for task-specific training. We demonstrate how CT-CLIP can be used in multi-abnormality detection and case retrieval, and outperforms state-of-the-art fully supervised models across all key metrics. By combining CT-CLIP’s vision encoder with a pretrained large language model, we create CT-CHAT, a vision-language foundational chat model for 3D chest CT volumes. Finetuned on over 2.7 million question-answer pairs derived from the CT-RATE dataset, CT-CHAT underscores the necessity for specialized methods in 3D medical imaging. Collectively, the open-source release of CT-RATE, CT-CLIP, and CT-CHAT not only addresses critical challenges in 3D medical imaging but also lays the groundwork for future innovations in medical AI and improved patient care. 

\end{abstract}

    % \begin{refsegment}  % Begin a new reference section for main content

    %\keywords{diagnosis, machine learning }
    % =========================
    % main part of the document
    % =========================
    \section*{Main}

Three-dimensional (3D) medical imaging, such as computed tomography (CT) and magnetic resonance imaging (MRI), provides a more detailed and comprehensive view of a patient’s health status compared to two-dimensional (2D) imaging~\cite{Mller2002ComputedTA}. Among these, CT stands out due to its speed and widespread availability, making it the preferred modality for diagnosing various abnormalities~\cite{Rubin2014ComputedTR}. As global demand for CT scans continues to grow, the need for rapid and accurate diagnostic results becomes more pressing. However, the manual generation of reports from these complex images is time-consuming and prone to errors, underscoring the urgent need for automation \cite{Claessens2015EarlyCC}. Artificial intelligence (AI) has the potential to address this challenge by efficiently identifying pathologies, reducing the workload on radiologists, and potentially saving lives in situations where immediate radiological expertise is not available.

AI has already begun to revolutionize the interpretation of medical images \cite{esteva2021deep,qin2018computer,lipkova2022deep, hamamci2023diffusion, pati2023gandlf}. Recent advances in automated diagnosis, classification, and segmentation have been driven by numerous publicly available datasets \cite{johnson2019mimic, wang2017chestx, nguyen2022vindr, irvin2019chexpert, hamamci2023dentex}. Beyond supervised methods, general AI systems have been developed using open-source image-text pairs in self-supervised learning, particularly in 2D imaging \cite{pathchat}. These versatile AI foundation models integrate comprehensive natural language annotations into the learning process without requiring task-specific training, thereby eliminating the need for explicit manual image annotations. Unconstrained by predefined classes or tasks during training, they enhance adaptability and have been successfully applied to tasks such as disease classification \cite{tiu2022expert}, case retrieval \cite{Hu2022XMIREM}, and interactive diagnostics via chat interfaces \cite{pathchat}. Despite these advancements in 2D imaging, the application of advanced AI techniques to 3D imaging remains underexplored. A major challenge is the lack of open-source 3D medical imaging datasets paired with textual reports, which require extensive computational resources, specialized knowledge, and technical expertise to create \cite{chen2022recent, zhou2021review, willemink2020preparing}. Currently, the largest 3D medical imaging datasets available contain only a few thousand cases and do not include paired textual data \cite{draelos2021machine, simpson2019large, bilic2019liver}, limiting the development of advanced AI models. Moreover, 3D models face higher computational demands and have fewer available pretrained open-source models compared to their 2D counterparts \cite{hamamci2023generatect, gao2022get3d}. Previous studies have only begun to address these challenges; for example, one involving chest CT is limited by its supervised approach, restricting detection to predefined abnormalities and offering only a small portion of its dataset publicly without paired textual reports \cite{draelos2021machine}.

To address these challenges, we introduce CT-RATE, a dataset that pairs 50,188 reconstructed 3D chest CT volumes (over 14.3 million 2D slices) from 25,692 scans of 21,304 patients with corresponding radiology reports. Each volume is annotated with 18 abnormalities, supporting extensive validation and fine-tuning. CT-RATE is fully open-source. Leveraging CT-RATE, we develop CT-CLIP, a chest CT-focused contrastive language-image pretraining framework. CT-CLIP integrates extensive natural language annotations into the learning process, enabling the model to comprehend image-based semantic knowledge and perform various tasks without specific training. Our evaluation shows CT-CLIP's ability to detect multiple abnormalities in a zero-shot setting \cite{xian2017zero}, outperforming state-of-the-art fully supervised methods on three test sets from two different countries. We extend CT-CLIP's capabilities by combining its vision encoder with a pretrained large language model to create CT-CHAT, a vision-language chat model for 3D chest CT volumes. CT-CHAT is fine-tuned with over 2.7 million question-answer pairs derived from the CT-RATE dataset. CT-CHAT outperforms open-source conversational vision-language AI assistants, such as LLaVA 1.6 \cite{llava1.6}, LLaVA-Med \cite{llavamed}, and CXR-LLaVA \cite{lee2023cxr}, underscoring the need for specialized tools in 3D medical imaging. We also benchmark CT-CHAT against state-of-the-art 3D medical report generation methods, including RadFM \cite{radfm} and CT2Rep \cite{Hamamci2024CT2Rep}, using clinically meaningful accuracy metrics. On two independent test sets, CT-CHAT outperforms these baselines by a significant margin, demonstrating both its clinical relevance and technical efficacy in report generation. In addition to quantitative benchmarks, we conduct expert evaluations with three board-certified radiologists, who consistently rated CT-CHAT’s reports as superior to those from RadFM and CT2Rep. While these results are encouraging, the experts emphasized that substantial progress is still needed for clinical deployment in 3D medical imaging. Our work marks a first step toward that goal, but important challenges remain.

    \section*{Results}
        \subsection*{CT-RATE: Pairing CT volumes with radiology text reports}

We curate and open-source CT-RATE, a dataset comprising 25,692 non-contrast 3D chest CT volumes and corresponding radiology text reports from 21,304 unique patients (Figure \hyperref[fig:fig1]{1a}). These 3D volumes, reconstructed using various methods to suit different window settings \cite{willemink2019evolution}, total over 14.3 million 2D slices. Each CT exam is accompanied by a report that includes an \emph{impression} and a \emph{findings} section, authored by a radiologist, along with additional sections detailing patient information and scan technique. CT-RATE is divided into training and internal validation sets for experiments (see \hyperref[ctrate_methods]{creating the CT-RATE dataset} section of \hyperref[methods]{Methods}). While CT-RATE is a single-center dataset, it captures substantial heterogeneity in CT scanners, acquisition protocols, and reconstruction techniques. 

To evaluate the generalization of our models (developed on CT-RATE) under distribution shifts, we incorporate two external validation sets from independent studies conducted in different countries \cite{draelos2021machine, Xu2023MedSyn} (see \hyperref[externalval_methods]{adding external validation datasets to CT-RATE} section of \hyperref[methods]{Methods}). First, we use the public RAD-ChestCT dataset, which includes multi-abnormality labels, to assess model generalizability. Second, we perform a real-world validation by inviting an independent clinical research team (uninvolved in model development) to download our open-source codebase, pretrained models, and user interface from our public repository and apply them to their own local, private dataset without any adaptation. The performance scores they obtain are presented in the following sections. In particular, none of the external validation data is used during development; all model development is conducted solely on CT-RATE.

Conventional supervised machine learning approaches for medical image classification necessitate manual annotation of large datasets, a process that requires significant time and expertise \cite{sager2021survey}. Although our self-supervised learning approach does not require abnormality labels, the baseline supervised model and finetuning methods do. To meet this need, we enrich CT-RATE with global abnormality labels extracted from the radiology reports using an automated text classifier (Figure \hyperref[fig:fig1]{1b}). This process involves finetuning a pretrained language model, RadBERT \cite{yan2022radbert}, similar to recent works \cite{minaee2021deep}. Within CT-RATE, we manually annotate 1,000 reports for 18 distinct abnormalities for this finetuning procedure. The finetuned model's precision, recall, and F1 scores, detailed in Supplementary Table \hyperref[fig:supp classifier]{2}, show robust classification across abnormalities, ensuring the CT-RATE dataset is enriched with high-quality labels.

Figure \hyperref[fig:fig3]{2} summarizes CT-RATE alongside the open-source external validation set, RAD-ChestCT. CT-RATE spans a wide age range (18-102 years) (Figure \hyperref[fig:fig3]{2a}) and includes diverse pathologies primarily related to the heart and lung (Figure \hyperref[fig:fig3]{2b}). The reports vary in length and are authored by numerous radiologists, reflecting a variety of writing styles (Figure \hyperref[fig:fig3]{2c}). CT studies in CT-RATE also vary in slice numbers (Figure \hyperref[fig:fig3]{2d}), slice thickness, axial pixel (XY) spacing (Figure \hyperref[fig:fig3]{2e}), and include different manufacturers and a balanced sex distribution (Figure \hyperref[fig:fig3]{2f}). This heterogeneity is crucial for developing robust models capable of handling out-of-distribution cases. Supplementary Table \hyperref[fig:supp dataset]{1} provides detailed statistics on abnormalities, manufacturers, and sex distributions in the training and internal validation sets. In conclusion, CT-RATE does not represent a curated study dataset but rather reflects real-world scenarios characterized by heterogeneity. This diversity ensures the models trained on CT-RATE are well-suited for deployment in clinical settings, offering reliable performance across a broad range of conditions.

\subsection*{CT-CLIP: Developing a visual-language foundation model with CT-RATE}
Unlike other methods using supervised machine learning for multi-abnormality classification of CT volumes, our study employs radiology reports, which provide a rich source of semantic knowledge. This significantly enhances the understanding of CT volumes and supports a wide range of potential zero-shot applications. Accordingly, we develop CT-CLIP, the first foundational model for chest CT-focused contrastive language-image pretraining, adapting a technique from a previous study \cite{radford2021learning}. We use both the impression and findings sections of radiology reports combined for CT-CLIP due to their superior performance compared to using either one alone, as demonstrated by the abnormality classification scores in an ablation study (Supplementary Figure \hyperref[fig:supp_ablations]{1a}). While the best performance is achieved training on both impression and findings together, using only the findings section performs better than using only the impression section, indicating that the findings section contains more clinically informative content.

CT-CLIP comprises a vision transformer and a text transformer that generate two distinct embedding vectors: one for chest CT volumes and another for reports, respectively (see \hyperref[cl_methods]{developing the CT-CLIP model} section of \hyperref[methods]{Methods}). The vision transformer is adapted from CT-ViT \cite{hamamci2023generatect}, incorporating a two-stage spatial and causal transformer, enabling 3D attention across volumes. CT-CLIP is trained to maximize the similarity between embedding vectors for paired volume-text data samples, while minimizing similarity for unpaired data. Our training approach leverages contrastive learning \cite{zhang2021cross} (Figures \hyperref[fig:fig1]{1c}, \hyperref[fig:fig1]{1d}), which enables the training of a foundation model suitable for various applications without task-specific training, including multi-abnormality detection, volume-to-volume retrieval, and report-to-volume retrieval. Subsequent sections demonstrate how CT-CLIP can be applied to these downstream tasks without additional training. To further enhance CT-CLIP's capabilities, we explore finetuning methods for detection to achieve greater accuracy and adapt CT-CLIP to create a vision-language AI assistant.

We analyze the latent spaces of reports and chest CT volumes generated by CT-CLIP, by computing t-SNE projections of these spaces. Figure \hyperref[fig:fig4]{4d} shows the distribution according to the number of abnormalities present in a case. Figure \hyperref[fig:fig4]{4e} depicts the age distribution on the t-SNE projections. The well-clustered t-SNE projections confirm CT-CLIP’s ability to accurately map latent spaces with closely aligned visual and linguistic information. For instance, in both image and text latents, the alignment between age and abnormality distributions indicates that older patients are more prone to abnormalities, which the model effectively identifies. Extended Data Figures \hyperref[fig:extfig4a]{1}, \hyperref[fig:extfig4b]{2} provide additional t-SNE projections for each abnormality using volume and report embeddings, respectively.

\subsection*{Surpassing supervised levels with CT-CLIP's zero-shot multi-abnormality detection}

CT-CLIP adopts a zero-shot classification approach for multi-abnormality detection without explicit training on specific abnormality labels. Zero-shot classification refers to the ability of a model to categorize previously unseen classes without being explicitly trained on those specific categories \cite{radford2021learning}. Instead of relying on predefined labels, CT-CLIP leverages the semantic information embedded in paired radiology reports to infer multiple abnormalities. For instance, a model trained to recognize features like “reticulation,” “consolidation,” or “nodules” in common lung diseases could potentially identify a rare lung condition it was not explicitly trained on by recognizing combinations of these features.

Zero-shot classification with foundation models is typically applied as a multi-class classification task, assigning a single label to each image \cite{radford2021learning}. However, a medical image may present multiple abnormalities, necessitating a unique approach. Inspired by recent work on 2D chest radiographs \cite{tiu2022expert}, we generate positive and negative prompts for each abnormality. The embeddings for the input 3D chest CT volume and the two prompts are projected into a shared latent space, and the contrastive loss between these projections is calculated based on the distance between the image and language spaces (see \hyperref[ctclip_methods]{CT-CLIP for zero-shot multi-abnormality detection} section of \hyperref[methods]{Methods} for more details). This process extracts and identifies the relevant features in both modalities. The model then outputs logits for each prompt, which are interpreted as the likelihood of abnormality presence, as shown in Figure \hyperref[fig:fig1]{1e}. We conduct a systematic evaluation of CT-CLIP's zero-shot detection capabilities on our internal validation set from CT-RATE and two external validation sets from Rad-ChestCT and UPMC (Figure \hyperref[fig:fig4]{4a}).

The careful engineering of prompts is instrumental in zero-shot classification tasks with vision-language models, as it significantly impacts performance \cite{yong2023prompt}. Therefore, we evaluate seven distinct prompts commonly used in our reports, with detailed information provided in the online methods section. Figure \hyperref[fig:fig4]{4b} presents the mean AUROC, F1 score, accuracy, and precision for these various prompts. Based on its consistently superior performance, we use prompt 7 (\textit{\emph{``\{Abnormality\} is \{$\bm{\mathit{\varnothing}}$/not\} present.''}}) in further experiments (see \hyperref[ctclip_methods]{CT-CLIP for zero-shot multi-abnormality detection} section of \hyperref[methods]{Methods}).

To evaluate CT-CLIP's zero-shot multi-abnormality detection, we benchmark it against CT-Net \cite{draelos2021machine}, a state-of-the-art fully supervised model for 3D chest CT volumes. CT-Net requires manual annotation for each predefined abnormality, which demands significant time and expertise \cite{sager2021survey}. We choose CT-Net as a baseline to demonstrate that CT-CLIP can surpass traditional fully supervised performance without relying on manual annotations or being restricted to predefined classes. Since the pretrained CT-Net model is not available and only a small portion of its dataset is open-source, we train CT-Net on CT-RATE using our abnormality labels for a fair comparison (see \hyperref[abnormalitydetection_methods]{benchmarking the CT-CLIP models for multi-abnormality detection} section of \hyperref[methods]{Methods}). Despite using a different dataset for training, the retrained CT-Net’s classification scores are comparable to those reported in the original study \cite{draelos2021machine}. Figure \hyperref[fig:fig4]{4a} shows the retrained CT-Net's performance metrics, while Extended Data Figures \hyperref[fig:extfig5]{3}, \hyperref[fig:extfig5external]{4}, \hyperref[fig:extfig5external_upmc]{5} and Supplementary Figures \hyperref[fig:supervisedconfusion]{2}, \hyperref[fig:supervisedconfusionexternal]{6} detail classification scores and confusion matrices.

We compare our zero-shot approach to this supervised baseline and observe significantly higher performance metrics for our approach. On the internal validation set from CT-RATE, the zero-shot CT-CLIP model achieves a mean AUROC that is 0.102 higher and a mean F1 score that is 0.050 higher than the supervised baseline. On the first external validation set, RAD-ChestCT, CT-CLIP outperforms the baseline with a mean AUROC gain of 0.085 and an F1 score improvement of 0.073. Similarly, on the second external validation set from UPMC, CT-CLIP maintains strong performance with a 0.087 increase in mean AUROC and a 0.037 boost in F1 score. Results on the external validation sets demonstrate that the zero-shot CT-CLIP model performs well not only on similar data distributions but also under distribution shifts, indicating its strong generalization to external datasets. A two-sided unpaired permutation test, using 10,000 permutations, highlights a statistically significant improvement \((p < 0.05)\) in AUROC as a zero-shot classifier over the supervised (baseline) model, detailed in Supplementary Table \hyperref[fig:supp p values]{5}.

Extended Data Figures \hyperref[fig:extfig5]{3}, \hyperref[fig:extfig5external]{4}, \hyperref[fig:extfig5external_upmc]{5} provide scores for all abnormalities on the CT-RATE internal, RAD-ChestCT external and UPMC external validation sets, respectively. Supplementary Figures \hyperref[fig:ctclipconfusion]{3}, \hyperref[fig:ctclipconfusionexternal]{7} show confusion matrices on the CT-RATE internal and RAD-ChestCT external validation sets, respectively. Notably, on the internal validation set, all abnormalities (18 out of 18) exhibit higher scores in our zero-shot approach compared to the supervised baseline in mean AUROC. Surprisingly, CT-CLIP excels by just leveraging existing radiology text reports in electronic health records, thus avoiding the need for expensive manual annotations. Moreover, CT-CLIP's zero-shot capability offers flexible predictions beyond predefined classes and adapts efficiently to changing needs post-training.

Since the availability of sufficiently large datasets is essential for machine learning, especially for transformers, which require more data than convolutional neural networks \cite{bailly2022effects, zhai2022scaling}, we explore the impact of dataset size on classification performance. We train CT-CLIP on different portions of the CT-RATE dataset, specifically 9.8\% (equivalent to the open-source portion of the RAD-ChestCT dataset \cite{draelos2021machine}, currently the largest publicly available chest CT dataset), 20\%, 40\%, 60\%, 80\%, and 100\%. Figure \hyperref[fig:fig4]{4c} reports the classification performance across different dataset sizes. Notably, a consistent trend emerges: larger dataset sizes correlate with improved scores. This finding highlights the role of dataset size in enhancing CT-CLIP's zero-shot classification performance and underscores the value of our extensive CT-RATE dataset for computational research. It also offers a clear avenue for future improvements, suggesting that further expanding the dataset or incorporating more diverse data could lead to even better performance.

\subsection*{Enhancing CT-CLIP’s detection performance with finetuning}

To enhance CT-CLIP's detection performance, we explore a task-specific finetuning method, ClassFine, inspired by conventional linear probing, by adding a new classification layer to the pretrained vision transformer (Figure \hyperref[fig:fig2]{3a}). Supplementary Figure \hyperref[fig:supp_ablations]{1b} indicates that training only the final layer may be sufficient for finetuning (see \hyperref[finetuning_methods]{finetuning CT-CLIP for multi-abnormality detection} section of \hyperref[methods]{Methods}). While the ClassFine approach improves detection performance (Figure \hyperref[fig:fig4]{4a}), it restricts the model to predefined labels as any supervised classification method. Extended Data Figures \hyperref[fig:extfig5]{3}, \hyperref[fig:extfig5external]{4}, \hyperref[fig:extfig5external_upmc]{5} present abnormality scores, with confusion matrices in Supplementary Figures \hyperref[fig:lipropconfusion]{5}, \hyperref[fig:lipropconfusionexternal]{9} on the internal and external validation sets, respectively. To address the limitations of ClassFine, we develop the VocabFine approach, a novel open-vocabulary finetuning method inspired by WISE-FT \cite{wortsman2022robust} (see \hyperref[finetuning_methods]{finetuning CT-CLIP for multi-abnormality detection} section of \hyperref[methods]{Methods}). VocabFine, like CT-CLIP’s zero-shot inference, is finetuned to improve label prediction accuracy (Figure \hyperref[fig:fig2]{3c}). Specifically, we use positive and negative prompts for abnormalities, calculate the logits with pretrained CT-CLIP, and align these logits with the ground truth labels. While the finetuned model with VocabFine retains the open-vocabulary inference capability of CT-CLIP (Figure \hyperref[fig:fig2]{3d}), it also consistently outperforms ClassFine (Figure \hyperref[fig:fig4]{4a}), further emphasizing the value of our open vocabulary finetuning. Abnormality-specific improvements are detailed in Extended Data Figures \hyperref[fig:extfig5]{3}, \hyperref[fig:extfig5external]{4}, \hyperref[fig:extfig5external_upmc]{5} while Supplementary Figure \hyperref[fig:supp_ablations]{1b} underscores the importance of training all weights in finetuning. Supplementary Figures \hyperref[fig:vocabfineconfusion]{4}, \hyperref[fig:vocabfineconfusionexternal]{8} provide confusion matrices on the internal and external validation sets, respectively. Notably, both ClassFine and VocabFine are trained solely on CT-RATE; thus, their generalization to external sets (RAD-ChestCT and UPMC) demonstrates the robustness of CT-CLIP’s pretrained representations under distribution shifts. This performance advantage across multiple test sets supports the effectiveness of our finetuning strategies. However, despite these promising results, the abnormality detection performance still falls short of clinical deployment thresholds.

Beyond its superior performance, VocabFine retains the open-vocabulary capabilities of the original CT-CLIP model, enabling inference on abnormalities that were not explicitly included during fine-tuning. This makes the model inherently more flexible and clinically useful in real-world scenarios where emerging or rare findings may need to be recognized without retraining. Additionally, unlike ClassFine, which only adds a classification head on top of the vision encoder and is therefore restricted to vision-only downstream tasks, VocabFine preserves and fine-tunes both the vision and text encoders. This dual-modality design allows it to generalize to a wider range of tasks, such as text-to-image retrieval and multimodal question answering, which rely on aligned visual and linguistic representations. In summary, our VocabFine approach improves multi-abnormality detection accuracy and also maintains the foundational model’s versatility with adaptability across diverse downstream applications.

\subsection*{Retrieving relevant cases with CT-CLIP}

In clinical practice, radiologists frequently refer to past cases with similar findings to guide diagnosis and treatment decisions. This process, known as case retrieval, enhances diagnostic accuracy and builds confidence in decision-making. However, manually searching for comparable cases is time-consuming and inefficient. An automated, robust solution can streamline case retrieval, saving both time and effort.

Beyond zero-shot detection, CT-CLIP serves as a powerful tool for image-to-image retrieval \cite{chen2022fast}, identifying the relevant volumes from a query volume by comparing cosine similarities between the embeddings (Figure \hyperref[fig:fig6]{5a}). We test on both CT-RATE internal and RAD-ChestCT external validation sets using the vision transformer of CT-CLIP, finetuned models, and the supervised baseline. We use the Mean Average Precision at K (MAP@K) metric, where K represents the number of returned 3D chest CT images (see \hyperref[retrieval_methods]{benchmarking the CT-CLIP models for CT volume retrieval} section of \hyperref[methods]{Methods}). As K increases, MAP@K scores typically decrease due to less accurate matches. Results, as in Figure \hyperref[fig:fig6]{5c}, show comparable performance among methods, with all significantly outperforming random retrieval. Although the external validation set from RAD-ChestCT has higher MAP@K scores, the fold increase over random retrieval is lower (three times on the external vs. six times on the internal set), due to differing abnormality distributions. This suggests that examining fold change offers a more accurate comparison. Beside, finetuning models does not significantly alter the vision transformer's weights, as CT-CLIP and the finetuned models perform similarly across MAP@K metrics.

CT-CLIP also functions as a text-to-image retrieval tool \cite{zhang2012query}, evaluating cosine similarities between target volume embeddings and query report embeddings (Figure \hyperref[fig:fig6]{5b}). Given the limitations of the supervised model and the ClassFine approach in encoding text, we employ CT-CLIP and the VocabFine approach for our experiments. Due to the lack of reports in the RAD-ChestCT dataset, only the internal validation set is used for report-to-volume retrieval. We evaluate performance using Recall@K, where K represents the number of returned images, reflecting the percentage of correctly matched images within the top K returns (see \hyperref[retrieval_methods]{benchmarking the CT-CLIP models for CT volume retrieval} section of \hyperref[methods]{Methods}). While Recall@K scores increase with more returned volumes, considering fold change relative to random performance is more meaningful. Figure \hyperref[fig:fig6]{5d} shows that CT-CLIP performs significantly higher than the random baseline. However, the Recall@K scores for VocabFine are substantially lower than CT-CLIP, consistent with the expectation that finetuning a foundation model for a specific task might diminish its efficacy for other tasks. The reduced performance with VocabFine in report-to-volume retrieval further indicates significant modifications to the text transformer weights due to finetuning.

\subsection*{CT-CHAT: Developing a multimodal AI assistant with CT-RATE and CT-CLIP}

Observing CT-CLIP’s remarkable capabilities for open-vocabulary inference, we build on this foundation to enable an interactive exchange between textual queries and diagnostic image information. To achieve this, we develop CT-CHAT, a multimodal generalist AI assistant specifically designed for 3D chest CT. CT-CHAT integrates both visual and language processing to perform various tasks, making it a versatile tool for clinical use. To train CT-CHAT, we create a visual question-answering (VQA) dataset, comprising over 2.7 million question-answer pairs derived from CT-RATE using Llama 3.1 7B \cite{llama3.1} (see \hyperref[ctvqa_methods]{creating the dataset for CT-CHAT} section of \hyperref[methods]{Methods}). This dataset consists of several components, including long-answer questions, short-answer questions, multiple-choice questions, and report generation tasks (Extended Data Figure \hyperref[fig:vqa_dataset]{8}), reflecting the diverse types of inquiries encountered in clinical practice.

CT-CHAT’s architecture leverages the strengths of CT-CLIP by using its pretrained vision transformer to encode 3D chest CT volumes. These encodings are then passed to a large language model (LLM) through a multimodal projector, as shown in Figure \hyperref[fig:fig6ctchat]{6a}. We perform an ablation study evaluating CT-CHAT's performance with 4 different state-of-the-art LLMs, as detailed in Extended Data Figure \hyperref[fig:vqa_extended]{6}. We select 70B Llama 3.1 \cite{llama3.1}, outperforming others in key metrics, for the optimal configuration of CT-CHAT.

We create CT-CHAT by finetuning the multimodal projector and Low-Rank Adaptation (LORA) weights \cite{lora} for the LLM (see \hyperref[ctchat_methods]{developing the CT-CHAT model} section of \hyperref[methods]{Methods}). Prior to this, we perform a medical concept alignment step, where the LLM is frozen, and the multimodal projector is trained solely to generate radiology reports. This step slightly enhances performance by aligning the multimodal projector with biomedical concepts (Supplementary Table \hyperref[fig:ctchat_ablations_supp]{10b}). The ablation study for the volume provision proves CT-CHAT extracts relevant information from the given 3D volumes (Supplementary Table \hyperref[fig:ctchat_ablations_supp]{10a}). Although CT-CHAT shows the ability to retrieve information from the given image, it is also expected to hallucinate some pathologies when the input volume is not provided. Thus, we implement a guardrail after the ablation study, with the system prompt (Supplementary Table \hyperref[fig:ctchat_system]{9}). Supplementary Figure \hyperref[fig:guardrail_examples]{11} provides an example of the guardrail in case the volume is not provided.

Finally, CT-CHAT generates contextually relevant and clinically accurate responses across a variety of tasks, from simple question-answering to the generation of detailed radiology reports as shown in Extended Data Figure \hyperref[fig:ctchat_examples_supp]{9}. The model is also capable of dynamic task switching, facilitated by special tokens that guide the LLM in generating the appropriate type of response based on the specific task. This flexibility makes CT-CHAT a powerful tool for enhancing radiologists' workflows, reducing the time and effort required for interpretation, and improving diagnostic accuracy in 3D medical imaging.

\subsection*{Surpassing other multimodal AI assistants with CT-CHAT}

Given the absence of generalist models for 3D imaging, we evaluate CT-CHAT against several state-of-the-art open-source vision-language AI models, including LLaVA 1.6 (Mistral 7B and Vicuna 13B) \cite{llava1.6}, LLaVA-Med \cite{llavamed}, and CXR-LLaVA \cite{lee2023cxr}. To ensure a fair comparison, these models, designed for 2D imaging tasks, are tested using Digitally Reconstructed Radiographs (DRRs) of 3D chest CT volumes, a method that ensures accurate representation in a 2D context, as suggested by \cite{hou2024shadow}, rather than using random or central 2D slices of a 3D volume (see \hyperref[benchmarking_ctchat_methods]{benchmarking the CT-CHAT model} section of \hyperref[methods]{Methods}). In contrast, CT-CHAT directly processes 3D chest CT, giving it a significant advantage in understanding and interpreting the complex anatomical and pathological information within these scans.

CT-CHAT generates clinically accurate answers to questions about the given CT, as in Figure \hyperref[fig:fig6ctchat]{6b}. To evaluate, we employ multiple metrics, including BLEU, METEOR, ROUGE-L, and CIDEr, which assess the quality of the generated text. Also, inspired by previous work \cite{llavamed}, we use an LLM-based metric, Llama score, to evaluate the clinical accuracy of responses, with the models’ outputs compared against ground truths (see \hyperref[benchmarking_ctchat_methods]{benchmarking the CT-CHAT model} section of \hyperref[methods]{Methods}). The results indicate that CT-CHAT, being capable of evaluating the full information of 3D volumes, consistently outperforms the other open-source vision-language AI assistants across all tasks. It shows superior ability in generating long and short answers, handling multiple-choice questions, and producing detailed radiology reports that align closely with clinical expectations, as shown in Figure \hyperref[fig:fig6ctchat]{6c}. Extended Data Figure \hyperref[fig:vqa_extended]{6} provides a detailed breakdown, demonstrating that CT-CHAT significantly outperforms others in all tasks. There are two key reasons why CT-CHAT significantly outperforms 2D-based models: (1) Image: CT-CHAT leverages the 3D spatial information processed by CT-CLIP’s pretrained vision transformer, which captures much richer and more detailed anatomical structures than 2D models, a critical advantage for CT interpretation. (2) Text: CT-CHAT is trained on 3D CT volumes paired with radiology-specific linguistic data from CT-RATE, enabling it to generate highly precise and clinically relevant responses, far surpassing those produced by models trained on general-purpose texts. This powerful combination of CT-CLIP's capacity to handle complex 3D imaging data and the domain-specific paired image-text data from CT-RATE allows CT-CHAT to push the boundaries of multimodal tasks in 3D medical imaging. In summary, CT-CHAT not only surpasses existing 2D tools but also provides a scalable solution for enhancing chest CT interpretation, underscoring the need for generalist models in 3D medical imaging.

In the absence of existing vision-language baselines for 3D medical imaging, we extend our evaluation of CT-CHAT beyond comparisons with 2D multimodal AI assistants by benchmarking its report generation capabilities against state-of-the-art 3D radiology report generation models, specifically RadFM~\cite{radfm} and CT2Rep~\cite{Hamamci2024CT2Rep}. This task is particularly important as it closely reflects the daily workflow of radiologists and provides a more clinically relevant benchmark than VQA. To robustly assess report generation accuracy, we conducted expert evaluations with three board-certified radiologists who independently rated 150 reports for each model (CT-CHAT, RadFM, and CT2Rep). To ensure a fair and comprehensive assessment, we did not select cases randomly; instead, we used the Llama score on CT-CHAT outputs to stratify cases into low (0–3), medium (4–6), and high (7–10) clinical accuracy ranges, selecting 50 cases from each range. Each radiologist then scored the reports on a scale from 0 to 10 based on predefined clinical criteria. The mean expert scores demonstrate CT-CHAT’s clear superiority: Radiologist~1 rated CT-CHAT at 3.44 (vs. RadFM~0.55, CT2Rep~2.66); Radiologist~2 at 4.10 (vs. RadFM~0.91, CT2Rep~2.52); and Radiologist~3 at 4.52 (vs. RadFM~0.86, CT2Rep~2.74). These results are presented in Figure~\hyperref[fig:fig6ctchat]{6d}.

To further strengthen report generation, we enhanced CT-CHAT with structured 3D segmentation inputs by providing detailed nodule information,such as lobar location, centroid, diameters, and volumes, extracted using a state-of-the-art 3D nodule segmentation algorithm on the CT-RATE training set. For implementation details on how we integrate nodule segmentation information, see the \hyperref[ctchat_methods]{developing the CT-CHAT model} section of \hyperref[methods]{Methods}. This additional structured input significantly improved report accuracy, as shown in Figure~\hyperref[fig:fig6ctchat]{6e}, demonstrating that leveraging high-quality 3D segmentation models meaningfully boosts clinical performance. Currently, we incorporate only nodule information; incorporating structured details for other abnormalities would likely enhance performance further, but robust 3D segmentation models for these additional findings are not yet available.

For quantitative evaluation, we exclude language generation metrics because they are limited in capturing clinical fidelity. Instead, we use clinical efficacy metrics (precision, recall, F1, and CRG score) that better reflect diagnostic value. Labels are extracted with our automated report labeler, the same model used to annotate CT-RATE, and CT-CHAT is benchmarked using its 8B Meta Llama~3.1 configuration. Although the RAD-ChestCT dataset lacks text reports, its binary abnormality labels allow us to apply the text labeler and compute metrics consistently. CT-CHAT outperforms RadFM and CT2Rep on the CT-RATE internal and RAD-ChestCT external validation sets across all metrics, as shown in Figure~\hyperref[fig:fig6ctchat]{6e}. CT-CHAT’s superior performance over CT2Rep is particularly notable, given that both models are trained on CT-RATE. Unlike CT2Rep, which does not leverage pretraining, CT-CHAT benefits from CT-CLIP’s pretrained vision encoder, demonstrating the strength of our foundational approach. Our results show that integrating outputs from other advanced 3D algorithms, such as robust segmentation tools, can further improve tasks like report generation. These findings highlight that continued development and integration of high-quality 3D models are essential for clinically deployable AI systems. By providing first-of-their-kind resources and benchmarks, especially through CT-RATE, our work takes an important step toward this goal, though significant challenges remain for routine clinical use.

    \section*{Discussion}
\begin{figure}[p]
    \centering
    \includegraphics[width=\textwidth]{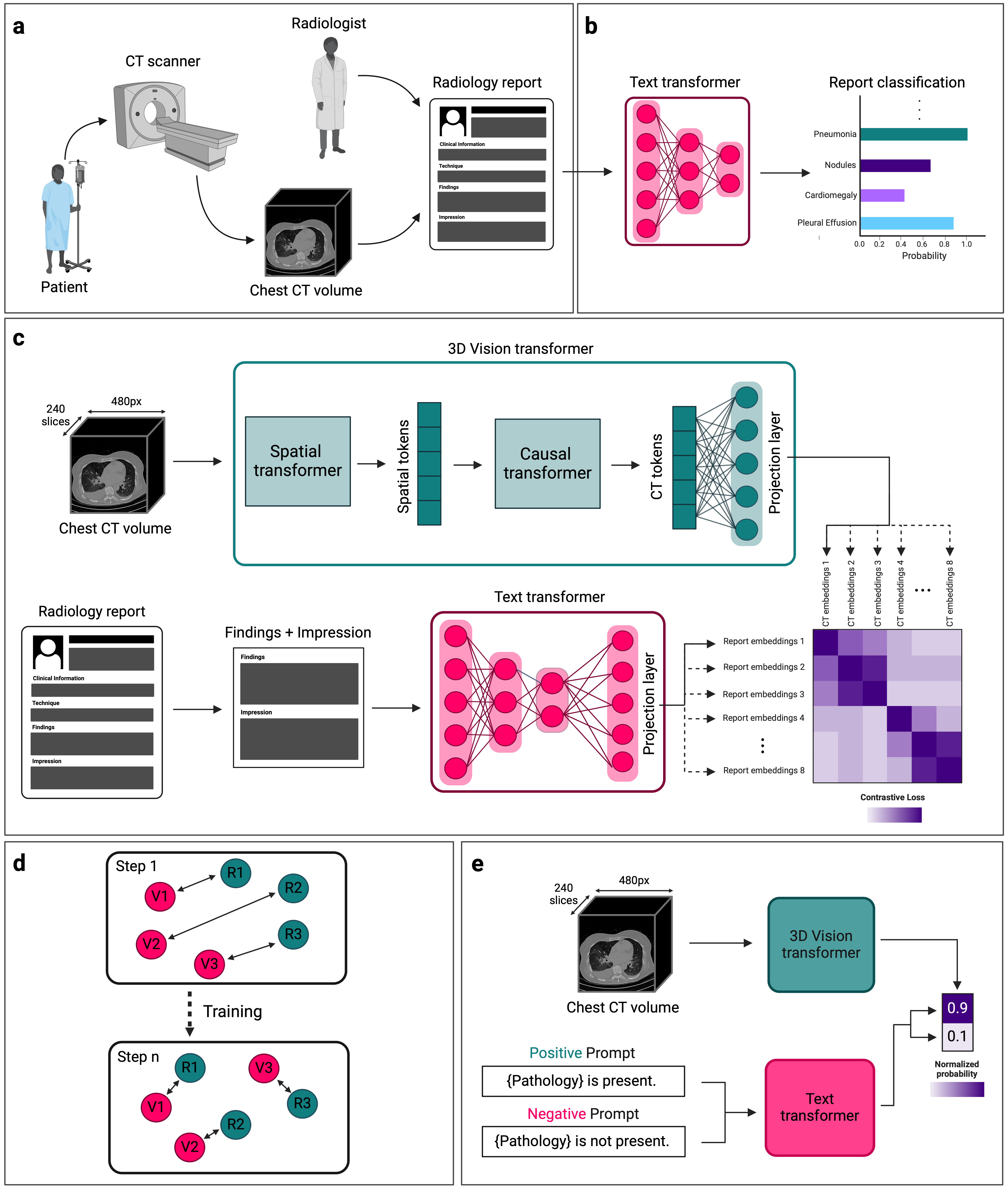}
    \caption{\textbf{Figure 1: Overview of our dataset and CT-CLIP.} \textbf{a.} Data acquisition protocol for the CT-RATE dataset, comprising non-contrast 3D chest CT volumes and corresponding radiology text reports written by radiologists. \textbf{b.} Extraction of multi-abnormality labels from radiology reports by using a finetuned text encoder. \textbf{c.} Training of our CT-CLIP framework, which pairs 3D chest CT volumes with radiology reports from the CT-RATE dataset through a contrastive learning approach. \textbf{d.} Illustration of the contrastive learning procedure, employed in the correlation matrix during training to ensure precise match recognition and mismatch differentiation. $\text{V}_\text{i}$ shows volume embeddings and $\text{R}_\text{i}$ shows report embeddings. \textbf{e.} Inference for zero-shot multi-abnormality detection uses normalized probabilities derived from softmax-applied contrastive loss for each positive and negative prompt.}
    \label{fig:fig1}
\end{figure}

        Radiology reports provide comprehensive descriptions of medical images without adding extra labeling burden, as they are already part of the daily routine for radiologists \cite{hartung2020create}. Recent studies show that foundational models, trained on paired radiology reports and 2D medical images, perform comparably to fully supervised methods in various tasks \cite{tiu2022expert}. However, these studies have not extended to 3D medical images, primarily because a publicly available 3D medical imaging dataset paired with text reports is absent. In this study, we introduce CT-RATE, a comprehensive dataset, which we first use to develop CT-CLIP, a foundation model for 3D medical imaging using radiology text reports (Figure \hyperref[fig:fig1]{1c}).

CT-CLIP provides a versatile solution for various tasks without requiring task-specific training. Unlike supervised methods, it eliminates the need for explicit labels and enables predictions beyond predefined classes (Figure \hyperref[fig:fig1]{1e}), making it particularly valuable when learning objectives evolve post-training. CT-CLIP's zero-shot multi-abnormality detection outperforms the state-of-the-art supervised method across all metrics (Figure \hyperref[fig:fig4]{4a}). Importantly, this experiment suggests that explicit labels are not necessary for high performance in 3D medical image interpretation when corresponding radiology reports are available for training. Furthermore, we externally validate CT-CLIP on two independent datasets, Rad-ChestCT and UPMC, from a different country (Figure \hyperref[fig:fig2]{3e}). The model's ability to generalize to different distributions, a primary challenge in deploying medical AI \cite{kelly2019key}, is confirmed as it outperforms the fully supervised method in all metrics in the external validation sets as well (Figure \hyperref[fig:fig4]{4a}). To establish a fully supervised baseline and finetune CT-CLIP, we extract multi-abnormality labels from reports, making them publicly available. Following recent examples, we develop an automatic labeler by finetuning a language model (Figure \hyperref[fig:fig1]{1b}) \cite{boecking2022making}. However, developing such a system requires time, expertise, and domain knowledge, as evidenced by our effort in manually annotating a small portion of the reports. CT-CLIP, as a zero-shot classifier, overcomes this bottleneck by detecting abnormalities at a supervised level without the need for an automatic labeler. Finetuning experiments further confirm CT-CLIP's representational learning capabilities, demonstrating superior performance (Figure \hyperref[fig:fig4]{4a}). Despite being trained without explicit supervision, CT-CLIP’s vision encoder captures clinically meaningful features that are highly transferable to downstream tasks. These finetuning strategies, one tailored for classification with predefined labels (ClassFine), and the other enabling open-vocabulary inference (VocabFine), build upon the same pretrained encoder and consistently yield superior results compared to a fully supervised baseline. This indicates that the encoder does not rely on memorizing specific abnormality classes but instead learns a more general and semantically aligned representation of the 3D chest CT volumes. Such flexibility underscores the foundational nature of CT-CLIP and its potential for broad applicability in medical imaging tasks, even beyond those explicitly considered during training. Finally, and indicating a direction for future developments, the effective clustering observed in t-SNE projections (Figure \hyperref[fig:fig4]{4d-e}) unveils another aspect of CT-CLIP’s clinical potential. The successful clustering of abnormalities (Extended Data Figures \hyperref[fig:extfig4a]{1} and \hyperref[fig:extfig4b]{2}) suggests that CT-CLIP could extend its use beyond multi-abnormality detection to assist in prognosis, disease grading, and the calculation of lung-related parameters in future research.

Analyzing CT-CLIP’s open-vocabulary inference capabilities, we observe potential for developing multimodal algorithms that address broader and more diverse clinical applications. To expand these capabilities, we finetune four state-of-the-art LLMs through visual instruction tuning with the CT tokens created by CT-CLIP, resulting in the creation of CT-CHAT. For finetuning, we create and open-source a visual question-answering dataset derived from CT-RATE (Extended Data Figure \hyperref[fig:vqa_dataset]{8}). As a result, CT-CHAT outperforms existing 2D models, underscoring the urgent need for generalist models specialized in 3D medical imaging (Figure \hyperref[fig:fig6ctchat]{6c}). We find that the choice of LLM (Supplementary Table \hyperref[fig:ctchat_ablations_supp]{10a}) and the use of concept alignment pretraining (Supplementary Table \hyperref[fig:ctchat_ablations_supp]{10b}) have minimal impact on CT-CHAT's accuracy. In contrast, including the volume as input is crucial, particularly in report generation (Supplementary Table \hyperref[fig:ctchat_ablations_supp]{10b}). This highlights the value of the CT-RATE dataset and CT-CLIP in developing a generalist AI assistant for 3D medical imaging. Also, we demonstrate that engineering the system prompt of CT-CHAT (Supplementary Table \hyperref[fig:ctchat_system]{9}) helps mitigate pathology hallucinations when the volume is not provided (Supplementary Figure \hyperref[fig:guardrail_examples]{11}). Our model, with new special tokens, can respond in various modes, including long and short answers, multiple-choice, and report generation (Extended Data Figure \hyperref[fig:ctchat_examples_supp]{9}). 

To further evaluate CT-CHAT’s clinical use, we benchmark its report generation performance against state-of-the-art 3D medical models RadFM \cite{radfm} and CT2Rep \cite{Hamamci2024CT2Rep}. Since radiology report generation closely mirrors the daily workflow of radiologists, it serves as a more clinically relevant benchmark than VQA. Instead of relying solely on conventional text similarity metrics, which are limited in assessing the clinical accuracy of generated content, we employ clinical efficacy metrics, using our in-house automated report labeler, which was also used to annotate CT-RATE (see \hyperref[ctrate_methods]{creating the CT-RATE dataset} section of \hyperref[methods]{Methods} for more details). CT-CHAT consistently outperforms both RadFM and CT2Rep on the CT-RATE internal and RAD-ChestCT external validation sets across all clinical metrics (Figure \hyperref[fig:fig6ctchat]{6e}). Moreover, we demonstrate that integrating structured outputs from other advanced 3D algorithms can further improve report generation. By incorporating detailed nodule segmentation information, such as lobar location and size, extracted using a state-of-the-art 3D segmentation model, we substantially enhanced CT-CHAT’s report accuracy (Figure~\hyperref[fig:fig6ctchat]{6e}). This experiment emphasizes that leveraging robust 3D models for different abnormalities could further boost performance, yet such high-quality models are still scarce. Continued development and open-source release of more domain-specific 3D segmentation algorithms are essential for future progress. Our expert evaluation adds a robust layer of validation, consistently confirming CT-CHAT’s superior performance (Figure~\hyperref[fig:fig6ctchat]{6d}). This direct clinical assessment highlights the real-world value of our approach. Notably, CT-CHAT’s advantage over CT2Rep, despite both models being trained on the CT-RATE dataset, demonstrates the effectiveness of pretraining with CT-CLIP. These results reinforce the strength of our foundational model, CT-CLIP, in learning clinically meaningful representations and enabling high-quality performance in downstream tasks. That said, it remains far from ready for clinical deployment, as emphasized by our radiologists. As with all AI systems, challenges such as robustness, interpretability, regulatory compliance, and prospective validation must be addressed before integration into routine radiology workflows. To help broaden accessibility, we introduce a user-friendly graphical interface (via our GitHub repository) that allows non-experts to interact with the CT-CHAT model. This interface not only makes CT-CHAT a valuable tool for educational and clinical research settings but also invites a wider audience to engage with and contribute to the evolution of interactive AI-driven inference in clinical practice.

Despite these advancements, our CT-RATE dataset and the CT-CLIP and CT-CHAT models have certain limitations. First, biases inherent to our single-institution dataset may influence our self-supervised training. Although we curate a large and diverse set of volume-report pairs, the data originates from a single geographic and clinical setting, which may limit the generalizability of our models to other populations. We conduct an ablation study to assess the impact of training set size on CT-CLIP's performance, revealing that performance increases with larger training sets (Figure \hyperref[fig:fig4]{4c}). This suggests that incorporating data from multiple institutions and more diverse patient populations could further enhance model robustness. Second, CT-CLIP depends on the specific terminology used in reports and the choice of prompts during zero-shot detection. Variability in how diseases are described across radiology reports can affect detection accuracy. To assess the importance of prompt engineering, we evaluate zero-shot multi-abnormality detection using seven different prompts (Figure \hyperref[fig:fig4]{4b}). While we select the best-performing prompt from commonly used structures in reports, continued optimization may further improve performance. Third, the evaluation of CT-CHAT faces limitations due to the lack of pretrained 3D medical multimodal AI assistants for benchmarking. As a result, we compare CT-CHAT against 2D models, which may not fully capture the 3D spatial information inherent in CT volumes. To address this, we use digitally reconstructed radiographs (DRRs) generated from our dataset \cite{hou2024shadow}. This approach is suggested by prior work showing that models trained with DRRs can yield comparable results to those trained with 3D chest CT volumes \cite{hou2024shadow}. However, the absence of true 3D benchmarks highlights the need for further development and validation of multimodal AI assistants in 3D medical imaging. Due to the absence of 3D VQA baselines, we also benchmark CT-CHAT’s report generation function against state-of-the-art 3D radiology report generation models. Despite CT-CHAT being trained for VQA, CT-CHAT outperforms these models; however, the quality of its generated reports remains insufficient for clinical use, underscoring the need for further improvements in performance and clinical validation. Finally, our current models are limited to chest CT volumes. To broaden their clinical applicability, future work should focus on extending CT-RATE, CT-CLIP, and CT-CHAT to include other 3D imaging modalities such as MRI and PET scans, as well as other anatomical regions. Expanding to different modalities and body parts would enhance the use of our models across a wider range of medical imaging tasks and clinical scenarios.

In conclusion, the development of CT-CLIP and CT-CHAT marks a major step forward in 3D medical imaging. Enabled by CT-RATE, an open-source 3D dataset paired with radiology reports, these models set a benchmark for multimodal AI in this domain. Their open availability provides a strong foundation for future research and, ultimately, for the development of clinically deployable systems.

%    \end{refsegment}

% \printbibliography[segment=1, heading=bibliography, title={Main References}]

\begin{figure}[p]
    \centering
    \includegraphics[width=\textwidth]{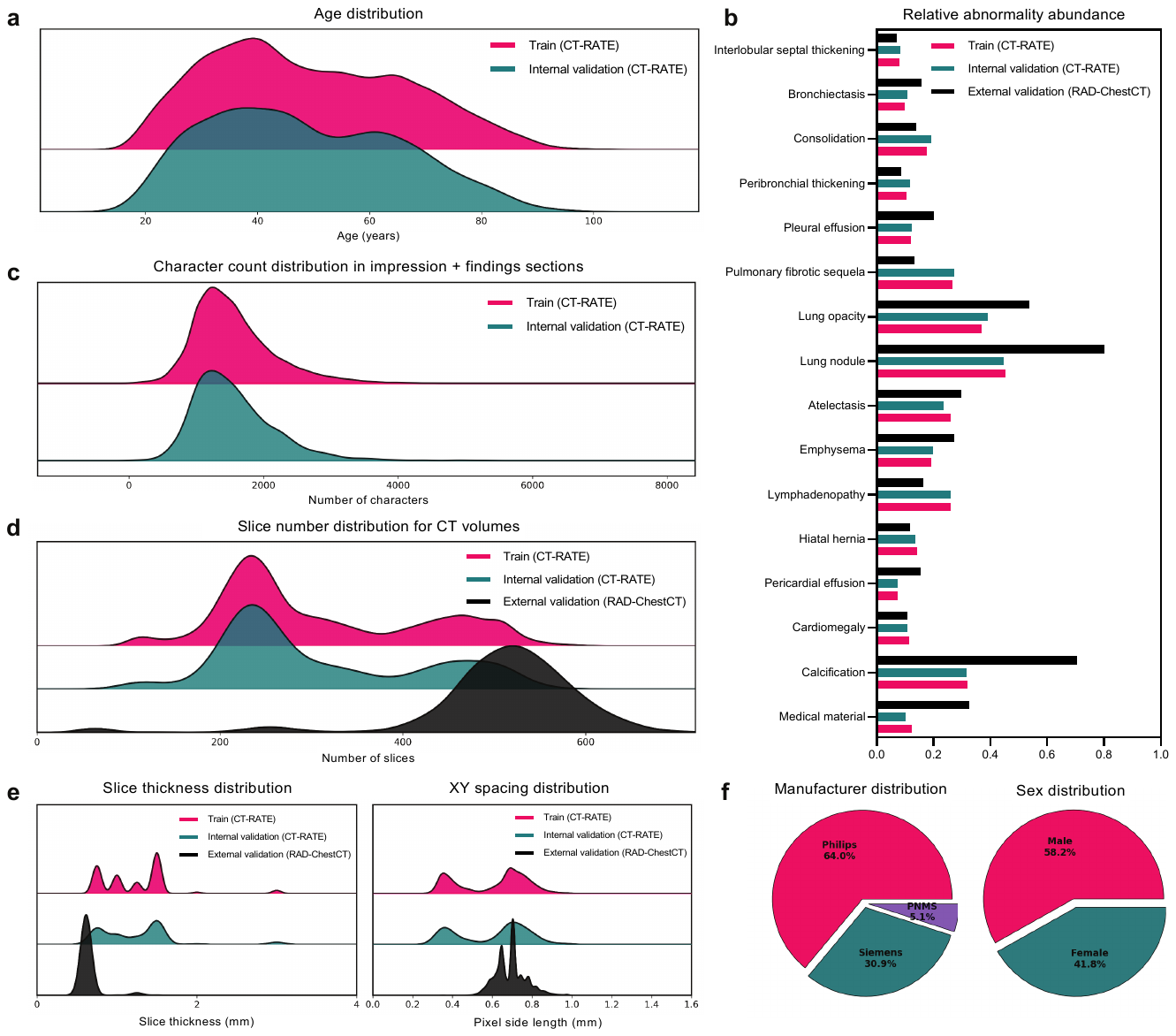}
    \caption{\textbf{Figure 2: Comprehensive analysis of the novel CT-RATE dataset.} \textbf{a.} CT-RATE comprises data from 21,304 unique patients, with ages ranging from 18 to 102 years. \textbf{b.} For validation and finetuning purposes, multi-abnormality labels for 18 distinct abnormalities are extracted from the corresponding radiology reports for each CT volume. \textbf{c.} The training process of the CT-CLIP model uses the \emph{impression} and \emph{findings} sections of radiology reports, which vary in length, along with the chest CT volumes. \textbf{d.} The number of slices in the chest CT volumes within CT-RATE ranges from 100 to 600. \textbf{e.} Spacings in the x-axis, y-axis, and z-axis vary across volumes; all chest CT volumes are preprocessed to ensure consistent spacing before training each model. \textbf{f.} Despite originating from a single institution, CT-RATE includes volumes acquired using scanners from three different manufacturers, ensuring variability.}
    \label{fig:fig3}
\end{figure}

\begin{figure}
    \centering
    \includegraphics[width=\textwidth]{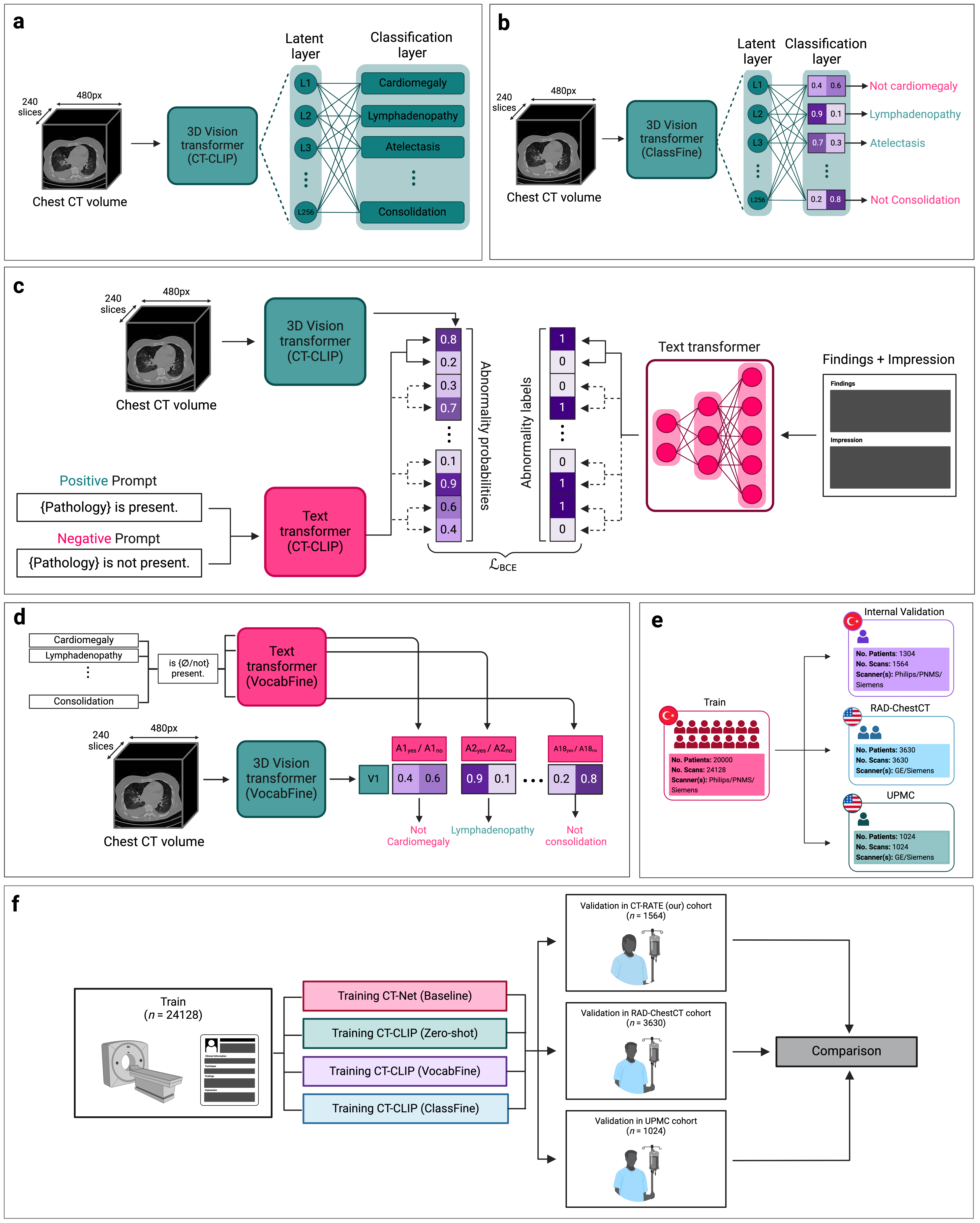}
    \caption{\textbf{Figure 3: Finetuning CT-CLIP and the validation strategy.} \textbf{a.} Illustration of linear probing finetuning method (ClassFine) for CT-CLIP, where a linear layer is incorporated into the vision encoder. \textbf{b.} ClassFine enables multi-abnormality classification but is limited to the classes predefined during finetuning. \textbf{c.} Illustration of CT-CLIP's open vocabulary finetuning method (VocabFine) for each abnormality. \textbf{d.} VocabFine allows for open vocabulary abnormality classification even after finetuning, although it is constrained to the prompts provided during finetuning. \textbf{e.} Validation: models are trained on the CT-RATE dataset from Türkiye and then tested on an internal CT-RATE validation set and two external datasets from the U.S.; a detailed analysis of the CT-RATE dataset is presented in Figure 2. \textbf{f.} Comparison: a comprehensive evaluation is performed in multi-abnormality detection across three different cohorts, evaluating CT-CLIP, the two finetuned models, and a fully supervised method.}
 
    \label{fig:fig2}
\end{figure}

\begin{figure}[p]
    \centering
    \includegraphics[width=\textwidth]{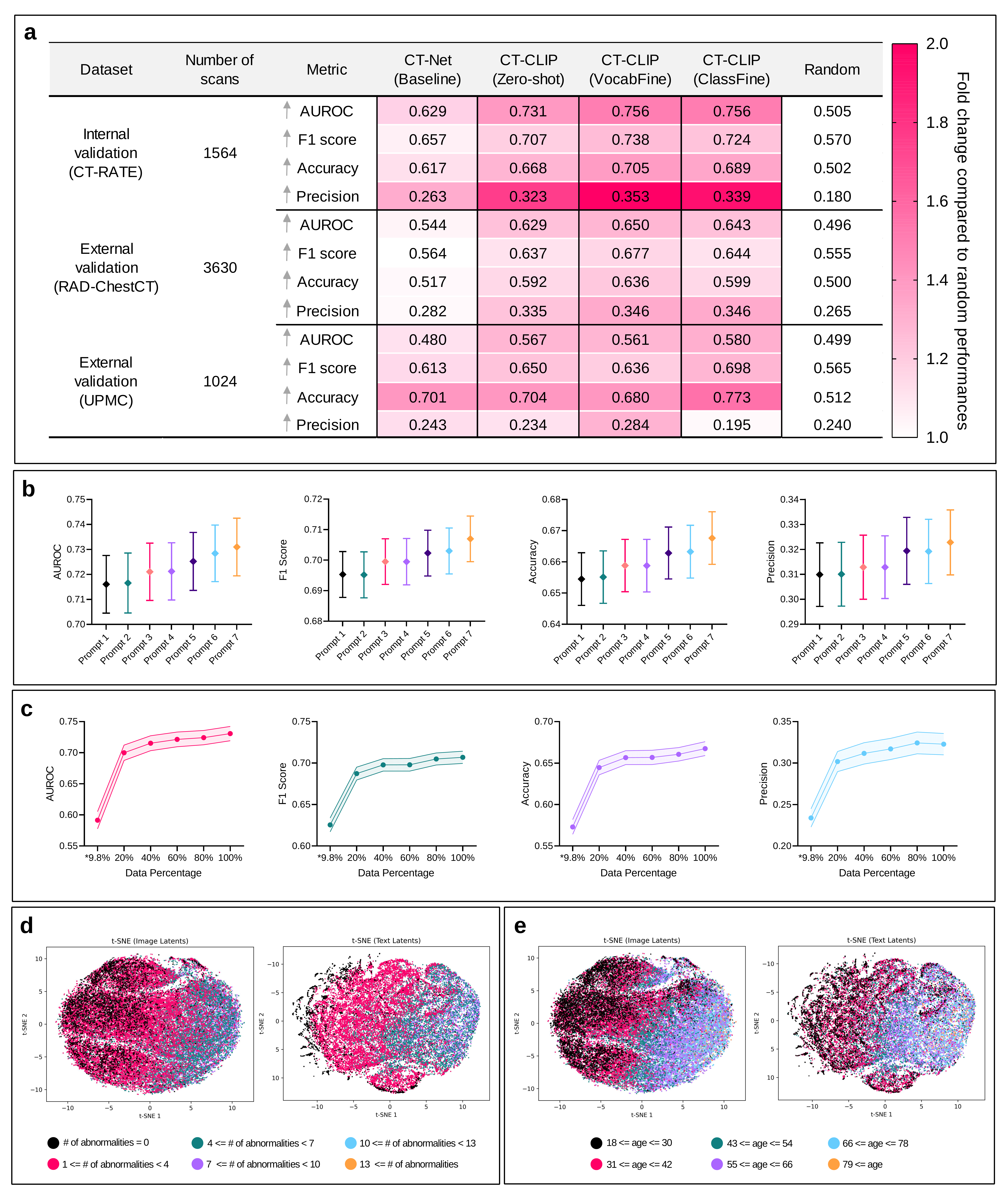}
    \caption{\textbf{Figure 4: A detailed evaluation of CT-CLIP} \textbf{a.} Multi-abnormality classification performance is evaluated on an internal and two external validation sets using four key metrics. Our CT-CLIP outperforms the fully supervised baseline across all metrics. Performance improves with both finetuning techniques, demonstrating the robustness of the CT-CLIP's encoder in feature extraction from CT volumes. For F1, a weighted score is computed per abnormality, then averaged across all findings to ensure balanced evaluation. The scores are color-highlighted to emphasize the performance increase compared to random benchmarks. \textbf{b.}Different prompts are used for zero-shot multi-abnormality detection with CT-CLIP. The best-performing prompt, highlighting the importance of prompt engineering, is selected for further experiments. Data are presented as mean ± SD of the bootstrapped values. \textbf{c.} To explore how the performance scales with dataset size, we train CT-CLIP using different portions of CT-RATE. (9.8\% of CT-RATE matches the size of the only open-source CT dataset with multi-abnormality labels.) Contour lines denote $\pm$1 standard deviations calculated via bootstrapping, with the center line indicating the mean. \textbf{d.} t-SNE projections map the distribution of the number of abnormalities present in cases, using the latent values from individual reports and CT volumes. \textbf{e.} t-SNE projections illustrate the age distribution of patients. The clustering shows that CT volumes and reports from older patients tend to correspond with a higher number of abnormalities, indicating a correlation between age and disease prevalence.}
    \label{fig:fig4}
\end{figure}

\begin{figure}[p]
    \centering
    \includegraphics[width=\textwidth]{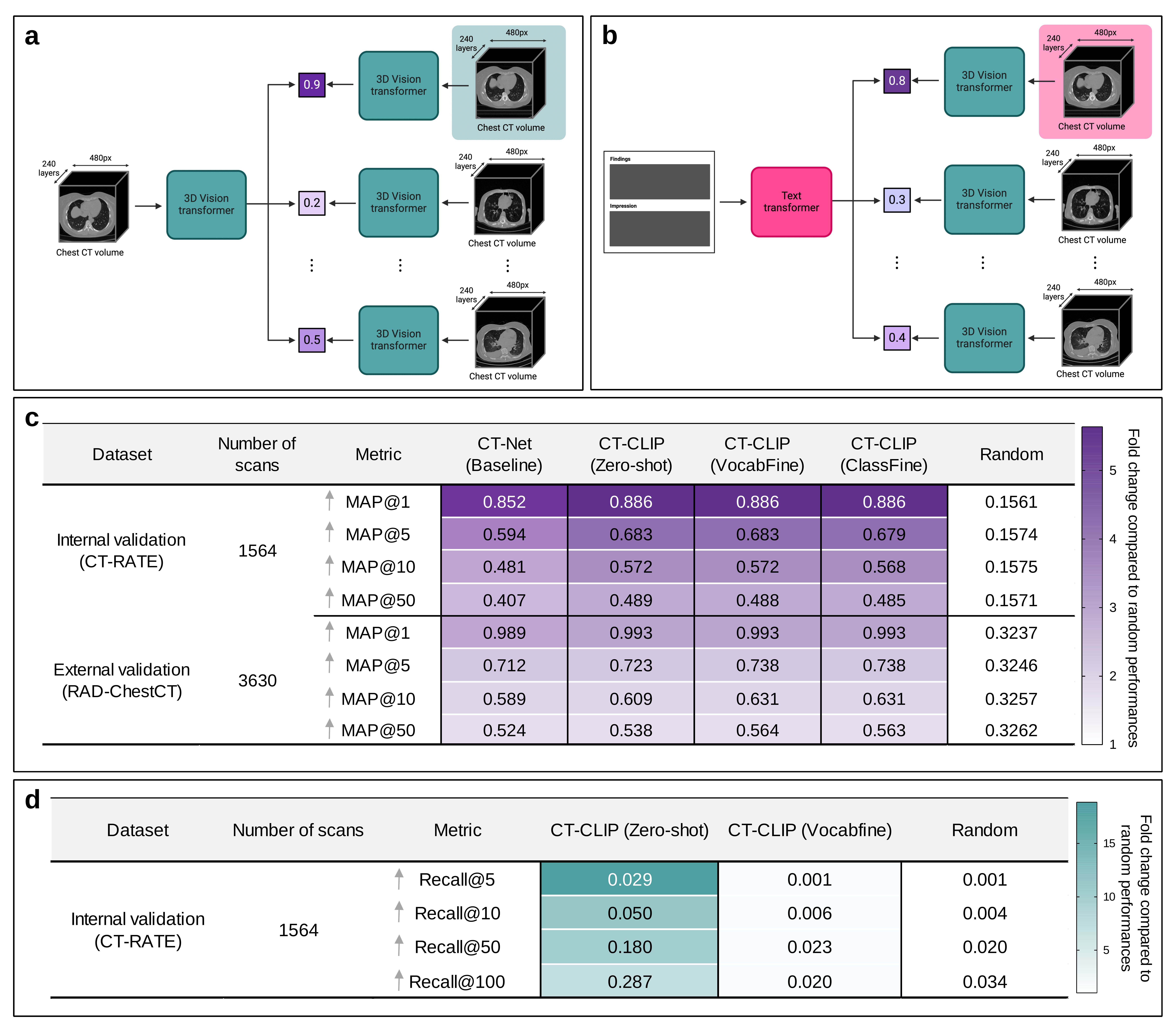}
    \caption{\textbf{Figure 5: Using CT-CLIP for retrieval of 3D chest CT volumes} \textbf{a.} CT-CLIP can be used for volume-to-volume retrieval using the vision encoder within CT-CLIP. \textbf{b.} CT-CLIP can be used for report-to-volume retrieval using both the vision and text transformers within CT-CLIP. \textbf{c.} Comprehensive evaluation of volume-to-volume retrieval performance on internal and external validation sets, with scores color-highlighted to emphasize improvements over random scores, shows similar retrieval accuracy even for the finetuned models. \textbf{d.} The evaluation of report-to-volume retrieval is limited to the internal validation set due to the absence of paired radiology reports in the external validation set. Notably, finetuning CT-CLIP for multi-abnormality classification diminishes its retrieval performance. This reduction in performance is primarily observed in report-to-volume retrieval, indicating that the VocabFine finetuning predominantly impacts on the text transformer component of CT-CLIP.}  

    \label{fig:fig6}
\end{figure}

\begin{figure}[p]
    \centering
    \includegraphics[width=\textwidth]{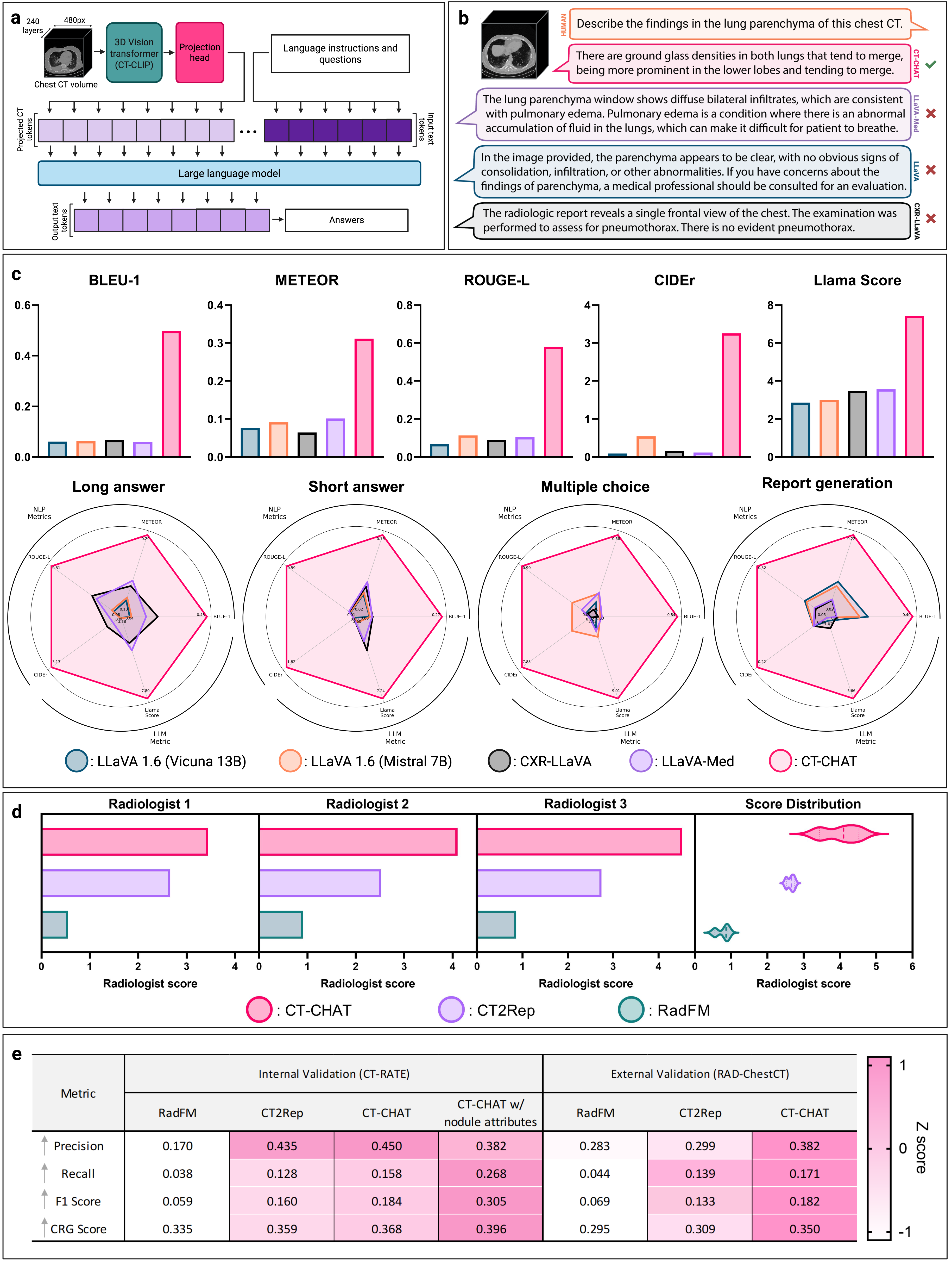}
    \caption{\textbf{Figure 6: Overview and detailed evaluation of CT-CHAT} \textbf{a.} The architecture of CT-CHAT, a vision-language AI assistant for 3D chest CT volumes, trained by finetuning LoRA weights on a state-of-the-art large language model using CT tokens encoded by the 3D vision encoder of CT-CLIP. \textbf{b.} An illustration showing an example CT-CHAT output, highlighting its superior performance as a vision-language assistant specifically designed for 3D chest CT volumes compared to its 2D counterparts with DRRs provided. \textbf{c.} The question-answering performance of various vision-language assistants is evaluated across five key metrics, with radar charts providing a detailed breakdown of evaluations for each of the four different visual question-answering tasks. \textbf{d.} Report generation performance, evaluated by three board-certified radiologists and a violin plot of scores, confirms CT-CHAT’s clear superiority. \textbf{e.} Report generation performance is evaluated using clinical efficacy metrics, calculated by adapting CT-RATE's automatic report labeler to assess clinical correctness across multiple findings on two test sets.}
    \label{fig:fig6ctchat}
\end{figure}

    %\bibliographystyle{ieeetr}

    %\defaultbibliography{bibliography}
    %\defaultbibliographystyle{ieeetr}

    %\bibliography{bibliography.bib}
\newpage
 %   \appendix

   % \begin{refsegment}
         
    \section*{Online Methods} \label{methods}
    \subsection*{Ethics statement}
\label{ethics_methods}
Our study is granted with ethical approval from the Clinical Research Ethics Committee at Istanbul Medipol University (E-10840098-772.02-6841, 27/10/2023) for open-sourcing our CT-RATE dataset and associated trained models. To uphold patient privacy, we anonymize all chest CT volumes, associated metadata, and radiology text reports before conducting any computational analysis.

\subsection*{Creating the CT-RATE dataset}
\label{ctrate_methods}

We curate a novel dataset, CT-RATE, consisting of chest CT volumes and corresponding radiology text reports from Istanbul Medipol University Mega Hospital. The CT-RATE dataset includes non-contrast chest CT volumes acquired between May 2015 and January 2023, totaling 50,188 reconstructed CT volumes from 25,692 distinct CT experiments conducted on 21,304 unique patients. We divide the cohort into two groups: 20,000 patients are allocated to training and 1,304 for validation. Figure \hyperref[fig:fig1]{1a} illustrates the dataset accumulation process, and Figure \hyperref[fig:fig2]{3e} shows the training and validation split.

In clinical practice, CT volumes are reconstructed with various techniques for specific diagnostic requirements. Sharper kernels, for instance, improve resolution for lung abnormalities, whereas smoother kernels are favored for assessing mediastinal issues \cite{willemink2019evolution}. For that reason, we incorporate every reconstructed volume from the hospital database into our dataset. This strategy facilitates a thorough evaluation of various abnormalities. The patients range in age from 18 to 102 years, with a mean of 48.8 years and a mode of 40 years. The sex distribution is 41.6\% female and 58.4\% male. 61.5\% of the scans are taken using Philips CT scanners, 30.1\% with Siemens scanners, and 8.4\% with PNMS (Philips-Neusoft Medical Systems) scanners. The CT volumes are provided in various resolutions: 65.4\% in 512 × 512 px, 4.2\% in 768 × 768 px, and 30.4\% in 1024 × 1024 px. The pixel spacing in the axial (XY) plane ranges from 0.227 to 1.416 mm, with a mean of 0.605 mm, while slice thickness ranges from 0.035 to 6 mm, with a mean of 1.231 mm. The number of slices per volume ranges from 100 to 600, averaging 304.7 with a mode of 255 slices. The radiology reports are structured into four sections: clinical information, technique, findings, and impression. The clinical information section outlines the patient's symptoms and medical history. The technique section specifies the acquisition protocol, encompassing details such as the field of view. The findings section covers anatomical and pathological observations, whereas the impression offers a diagnosis based on these findings. Figure \hyperref[fig:fig3]{2} summarizes the CT-RATE dataset alongside the first external validation set (RAD-ChestCT). Unfortunately, detailed metadata for the second external validation set (UPMC) is not publicly available, as it is a private dataset and cannot be shared; thus, it is not included.

To anonymize the CT-RATE dataset, we first anonymize reports by removing all personal information pertaining to patients and medical professionals. This process involves targeted searches using regular expressions to identify and eliminate date-related information and patient-doctor identifiers. Subsequently, these anonymized reports are translated from Turkish to English using the Google Translate API, and bilingual final-year medical students review and correct all translations. They also thoroughly examine each report to remove any residual personal information and ensure coherence. Only the English versions of these reports are included in our CT-RATE dataset. For the CT volumes, we implement several quality control measures. We exclude volumes that involve contrast material or body parts other than the chest. Only the remaining data is used for developing our models and open-sourced as the CT-RATE dataset. Each chest CT volume's orientation is standardized by examining its metadata to ensure consistency. Additionally, we remove all identifying information about patients and medical professionals from the metadata. The anonymized metadata is open-sourced as part of the CT-RATE dataset. Extended Data Table \hyperref[fig:dataset_attributes]{1} provides a detailed explanation of these metadata attributes.

To develop an automated label extractor model, we annotate 1,000 reports by merging findings and impression sections, as some abnormalities are not explicitly mentioned in a single section. This creates a multi-label dataset for each report, identifying 18 abnormalities. Conditions that are not referenced in either section are considered absent. Similar conditions are grouped: left and right mucoid impactions under \emph{mucoid impaction}, and various lung opacities (lung opacity, density increase, and ground-glass opacities) under \emph{lung opacity}. Lung and fissural nodules are categorized as \emph{lung nodules}. The annotated reports are divided into training (80\%, 800 reports) and validation (20\%, 200 reports) sets. Leveraging this dataset, we finetune the RadBERT-RoBERTa-4m model \cite{yan2022radbert}, which was originally trained on over 4 million radiology reports. Then, we apply it to classify the remaining 24,692 reports. These classified reports serve three purposes: training the supervised baseline model, finetuning CT-CLIP, and internal evaluation of the models. Figure \hyperref[fig:fig1]{1b} illustrates the automated classification process for radiology reports. Supplementary Table \hyperref[fig:supp classifier]{2} shows the precision, recall, and F1 scores of the report classifier.

Our preprocessing approach, inspired by previous work \cite{draelos2021machine}, involved several key steps to standardize all chest CT volumes in the dataset. First, each volume is resized to achieve uniform spacing of 0.75 mm in the x-axis and y-axis and 1.5 mm in the z-axis. The CT volumes are then either center-cropped or padded to reach a consistent resolution of $480 \times 480 \times 240$, ensuring comprehensive inclusion of the patient's relevant area for the interpretation. Subsequently, we convert each CT volume to Hounsfield Unit (HU) values using intercept and slope values from the metadata. HU values, a measure of radiodensity, are clipped to a range of -1000 to 1000. This range represents the practical lower and upper limits of the HU scale \cite{denotter2023hounsfield}. During training, we normalize these values to a range of -1 to 1. The CT-RATE dataset contains the original raw data and all preprocessing scripts are provided in our GitHub repository.

In addition to radiology reports and metadata, we plan to extend CT-RATE with multi-abnormality segmentation and localization labels for 3D chest CT volumes. To our knowledge, no publicly available dataset currently provides such comprehensive annotations for 3D medical imaging, making this a highly valuable contribution. Given the substantial effort and expert input required, this will be pursued as part of future work. As an initial step, we include organ-level anatomical segmentation labels generated using a state-of-the-art segmentation model \cite{xu2025cads} (see Supplementary Table \hyperref[fig:sup_seg]{11}). These labels are openly available through the CT-RATE repository on Hugging Face and hopefully support tasks such as visual grounding, anatomical localization, and serve as a foundation for future fine-grained abnormality segmentation.

\subsection*{Adding external validation datasets to CT-RATE}
\label{externalval_methods}

For the first external validation set, we use the RAD-ChestCT dataset, which includes a cohort from another country \cite{draelos2021machine}. The dataset comprises 3,630 non-contrast chest CT volumes, uniformly reconstructed using a single technique. Half are conducted with Siemens CT, and the other half with GE Medical scanners. The number of slices ranges from 46 to 1277, with an average of 511.6 and a mode of 472. All CT volumes have a resolution of 512 × 512 pixels. The pixel spacing in the axial (XY) plane ranges from 0.189 to 0.977 mm, with an average of 0.692 mm, while the slice thickness varies from 0.125 to 5 mm, with an average of 0.706 mm. To ensure consistent evaluation, we apply identical preprocessing methods to this external dataset as used with our CT-RATE cohort. Since the RAD-ChestCT dataset lacks a \emph{Mosaic attenuation} label, we exclude it from our external evaluation. Furthermore, while the RAD-ChestCT dataset includes a \emph{Calcification} label, our dataset distinguishes between \emph{Arterial wall calcification} and \emph{Coronary artery wall calcification}. In assessing the model accuracies on the external evaluation set, we use the higher classification probabilities between these two labels for the \emph{Calcification} prediction.

For the second external validation set, we use the validation split from the MedSyn paper \cite{Xu2023MedSyn}, which contains 1,024 chest CT scans and paired radiology reports, each from a different subject. These scans originate from the University of Pittsburgh Medical Center (UPMC) and have been fully de-identified. We employ the same automated report labeler used for CT-RATE to extract 18 abnormality labels from the text reports for evaluation. To ensure consistent and fair comparison, we apply the same preprocessing pipeline to chest CT volumes as used with our CT-RATE cohort.

\subsection*{Developing the CT-CLIP model}
\label{cl_methods}
CT-CLIP is a novel 3D adaptation of the Contrastive Language-Image Pretraining (CLIP) which is originally designed for natural image-text pairs \cite{radford2021learning} and has shown excellent performance in 2D medical data \cite{tiu2022expert, huang2023visual}. It learns to identify similarities between text and images via contrastive learning (Figure \hyperref[fig:fig1]{1d}), enabling efficient zero-shot classification. We extend CLIP to work with 3D CT paired with reports. It integrates a 3D encoder, previously used for generating CT \cite{hamamci2023generatect}, as the image encoder and CXR-Bert \cite{boecking2022making} as the text encoder. Similar to the 2D CLIP, our method extracts latent embeddings from input volumes and radiological reports. We then compute the contrastive loss, which measures the similarity between these embeddings. Training the model to align the 512-dimensional projection layers from text and CT volume embeddings with the identity matrix (as shown in Figure \hyperref[fig:fig1]{1c}) allows identifying meaningful correlations between textual information and 3D medical images. This process enables zero-shot classification and can be further enhanced through task-specific finetuning for supervised classification.

As a 3D vision transformer, we use a component of CT-ViT \cite{hamamci2023generatect}, a network designed to generate low-dimensional tokens from CT volumes and reconstruct CT volumes from these tokens. CT-ViT divides CT volumes into patches measuring $20 \times 20 \times 10$ voxels in the coronal, sagittal, and axial planes, ensuring uniform spacing. These patches are processed by spatial and causal transformers to produce low-dimensional CT tokens, which are then decoded to reconstruct the CT volumes \cite{hamamci2023generatect}. For CT-CLIP, we use only the encoder part of this network to generate CT tokens. During the training phase of CT-CLIP, the encoded CT tokens are averaged across the axial plane, then flattened and transformed into a 512-dimensional layer. This creates latent representations for the CT volumes, as shown in Figure \hyperref[fig:fig1]{1c}. This approach reduces GPU memory requirements significantly through the linear transformation layer for latent representations, while still capturing information from every slice due to the axial attention mechanism in the CT-ViT. The improved memory efficiency enables the use of a larger 3D CT encoder.

For latent representations of radiology reports, we employ the CXR-Bert text encoder \cite{boecking2022making}. This encoder can process up to 512 tokens, which allows the inclusion of both impression and findings sections from reports. For reports with content of fewer than 512 tokens, we use padding with the $<$PAD$>$ token, where padded tokens are masked during the attention mechanism. Each token is represented in 768 dimensions in the CXR-Bert text encoder. These tokens are then summed across the 512-token span and linearly transformed into a compact 512-dimensional layer, encapsulating the essential textual information from the reports. Figure \hyperref[fig:fig1]{1c} visually represents this encoding process. Additionally, to assess the impact of including the impression and findings sections, we conduct an ablation study on impression only and findings only training in Supplementary Figure \hyperref[fig:supp_ablations]{1a}. To highlight the impact of scaling the dataset size on model performance, we conduct an ablation study. This experiment involves training the CT-CLIP model with varying sizes of the CT-RATE dataset: 9.8\% (matching the size of the RAD-ChestCT dataset, the only public chest CT dataset with multi-abnormality labels \cite{draelos2021machine}), 20\%, 40\%, 60\%, 80\%, and 100\%. By systematically altering the sizes of the dataset, our goal is to explore how the accuracy of the model scales with the amount of available training data. Figure \hyperref[fig:fig4]{4c} illustrates the outcomes of this ablation study.

\subsection*{CT-CLIP for zero-shot multi-abnormality detection}
\label{ctclip_methods}

In our CT-CLIP model, we implement a similar approach to CheXzero \cite{tiu2022expert} for zero-shot multi-abnormality classification. This method adapts CLIP's zero-shot classification technique to handle multiple classes instead of multiple labels for the same class, which is particularly important in medical imaging where images often contain several abnormalities simultaneously. We start by extracting the model's preliminary output scores, known as logits, using both positive (e.g., “Consolidation is present.”) and negative (e.g., “Consolidation is not present.”) prompts. The prompting method for inference is chosen through prompt engineering experiments. We then apply softmax to these logits to convert them into a likelihood of each abnormality's presence or absence in a CT volume. Figure \hyperref[fig:fig1]{1e} illustrates CT-CLIP's inference strategy for the zero-shot multi-abnormality detection task.

To optimize CT-CLIP's performance for zero-shot multi-abnormality detection, we experiment with various prompts, derived from common sentence structures in radiology reports. The positive-negative formulations for prompts 1 to 7 are as follows: \textit{\emph{``\{$\bm{\mathit{\varnothing}}$/No\} \{abnormality\}.''}},
\textit{\emph{``Findings are \{$\bm{\mathit{\varnothing}}$/not\} compatible with \{abnormality\}.''}},
\textit{\emph{``There is \{$\bm{\mathit{\varnothing}}$/no\} \{abnormality\}.''}}, 
\textit{\emph{``\{Abnormality\} is \{$\bm{\mathit{\varnothing}}$/not\} seen.''}}, 
\textit{\emph{``\{$\bm{\mathit{\varnothing}}$/Not\} \{abnormality\}.''}},
\textit{\emph{``\{Abnormality\} is \{$\bm{\mathit{\varnothing}}$/not\} observed.''}}, and 
\textit{\emph{``\{Abnormality\} is \{$\bm{\mathit{\varnothing}}$/not\} present.''}}. We evaluate these prompts based on their classification scores (Figure \hyperref[fig:fig4]{4b}) and apply the most effective one to finetune the CT-CLIP model using the VocabFine approach and during inference.

\subsection*{Finetuning CT-CLIP for multi-abnormality detection}
\label{finetuning_methods}
To enhance the performance of the CT-CLIP model in classifying multiple abnormalities, we implement two finetuning approaches. The first approach, called VocabFine, involves finetuning the model while retaining its open vocabulary capabilities. The second approach, adapted from linear probing, is called ClassFine. This method entails training the model with an additional classification layer. We also conduct ablations to investigate the impact of freezing pretrained layers within CT-CLIP during finetuning (Supplementary Figure \hyperref[fig:supp_ablations]{1b}). Specifically, in addition to training every layer, we finetune only the projection layers of the image and text encoders in the case of VocabFine. For ClassFine, alongside training every layer, we train the model only for the newly added classification layer.

The first approach, VocabFine, is a modification of WISE-FT, a technique for finetuning foundation models \cite{wortsman2022robust}. In WISE-FT, the zero-shot model generates logits using classification labels as input text prompts. The model is then finetuned using binary cross-entropy (BCE) \cite{zhang2018generalized} to compare these logits with the actual ground truth labels. This process allows the model to learn specific classification labels while maintaining its ability to function with an open vocabulary \cite{wortsman2022robust}. Our VocabFine approach adapts this method for the multi-label classification. We generate both positive and negative prompts for each abnormality and then concatenate their respective logits into a unified array of 36 elements (two logits for each of the 18 abnormalities). The ground truth labels are represented as binary arrays (1s and 0s), where $1$ indicates a positive logit if the ground truth is $True$ and vice versa for $False$. For example, if a volume's label is $True$ for a specific abnormality, it corresponds to $[1, 0]$ in the ground truth array; $False$ would be represented as $[0, 1]$. The training aligns these ground truth values with the computed logits. To optimize computational efficiency, we divide the 36-element array into smaller segments of 12 elements for each gradient run. Figure \hyperref[fig:fig2]{3c} depicts the finetuning process in VocabFine. The inference strategies of both CT-CLIP and the finetuned model with Vocabfine are the same (Figure \hyperref[fig:fig2]{3d}).

Our second finetuning strategy, ClassFine, is achieved by integrating a linear layer into the visual encoder network, a technique inspired by \cite{alain2016understanding}. The architecture of the ClassFine approach is demonstrated in Figure~\hyperref[fig:fig2]{3a}. This modified model undergoes training using ground truth labels. The finetuned model with ClassFine is then evaluated using both internal and external evaluation sets. While this approach offers the potential for improved classification accuracy, it also makes the foundation model more specific for the trained classification task. Since ClassFine is based on fully supervised finetuning, its capabilities are primarily limited to the classes defined during this finetuning phase. Figure \hyperref[fig:fig2]{3b} visualizes the inference process of ClassFine, highlighting its specific application to the predetermined classes.

\subsection*{Benchmarking the CT-CLIP models for multi-abnormality detection}
\label{abnormalitydetection_methods}

In assessing our CT-CLIP and finetuned models, we employ the CT-Net model as a fully supervised baseline \cite{draelos2021machine}. Renowned for its performance in a similar CT dataset \cite{draelos2021machine}, CT-Net provides a robust comparison point. We train the CT-Net model on our CT-RATE dataset, adapting the chest CT volumes to fit the model's input requirements by resizing them to $420 \times 420 \times 402$. Consistent with our models, we focus on the abnormalities present in the CT-RATE dataset (Figure \hyperref[fig:fig2]{3d}), instead of the 98 and 9 abnormalities explored in the original CT-Net study. For each abnormality, we determine classification thresholds by identifying the left-upper point on the Receiver Operating Characteristic (ROC) curve, using the 10\% of each validation set. This analysis is performed across all three evaluation sets, CT-RATE (internal), RAD-ChestCT (external), and UPMC (external), and for all multi-abnormality detection models: zero-shot, fine-tuned, and fully supervised. Additionally, we compute the Area Under the Receiver Operating Characteristic (AUROC) values for each abnormality.

To provide a comprehensive evaluation, we also calculate accuracy, precision, and weighted F1-scores for each identified abnormality. For the weighted F1, each of the 18 abnormalities is treated as a binary task (present vs. absent); we compute a weighted F1-score that balances positive and negative cases within each abnormality, then average these scores to produce a single overall metric. This \emph{macro of weighted F1 scores} ensures that (i) each abnormality contributes equally to the final score and (ii) class imbalance within each abnormality is addressed fairly, resulting in a clinically meaningful measure of performance. To further validate the latent spaces generated by CT-CLIP for radiology reports and 3D chest CT volumes, we also conduct a t-SNE (t-distributed stochastic neighbor embedding) analysis~\cite{van2008visualizing}. This allows us to visually map the distribution of abnormalities and other relevant attributes within the t-SNE embeddings (Figures~\hyperref[fig:fig4]{4d},\hyperref[fig:fig4]{4e} and Extended Data Figures\hyperref[fig:extfig4a]{1},~\hyperref[fig:extfig4b]{2}).

For the statistical analysis of our model evaluations, we use the non-parametric bootstrap method to examine the distribution of predictions on internal and external validation sets. This involves 1,000 iterations of random sampling of the volumes with replacement, each using sample sets of size $n$, matching the number of evaluated volumes. We calculate the average metric values for each sample and quantify dispersion by computing the standard deviation of these bootstrap values. For statistical significance, we employ a two-sided paired permutation test with 10,000 permutations to assess differences in performance between the two models across AUROC, weighted F1 score, accuracy, and precision metrics, following previous work \cite{Chen2024TowardsAG}. Detailed comparisons are provided in Supplementary Table \hyperref[fig:supp p values]{5}.

\subsection*{Benchmarking the CT-CLIP models for CT volume retrieval}
\label{retrieval_methods}
Image retrieval involves identifying the most relevant image from a pool of candidates based on either textual \cite{kherfi2004image} or visual \cite{hegde2019similar} input. With CT-CLIP, trained to associate CT volumes with text reports, volume retrieval becomes possible. This is achieved by computing the cosine similarity among all pairs of volumes and reports, as well as among volume pairs, within a unified embedding space. This embedding space is generated by projecting CT tokens derived from the volumes through the projection layer, Figure \hyperref[fig:fig1]{1c}. To conduct volume-to-volume retrieval, we calculate the cosine similarity between the latent embeddings of a specific chest CT volume in the validation set and all other abnormal chest CT volumes within the same set. We then rank the volumes based on their cosine similarity scores to identify the most relevant matches for a given volume. This volume-to-volume retrieval process is shown in Figure \hyperref[fig:fig6]{5a}.

To evaluate this retrieval task's performance, we adapt the Mean Average Precision at $K$ (MAP@K) metric, typically used in image-to-image retrieval tasks \cite{chen2022fast}. In this metric, an image is considered relevant if it matches the classes of the target input. Precision is then calculated as the proportion of relevant images within the top $K$ retrieved images. Relevance is determined using the intersection over the union of the abnormality labels marked as $True$ in the ground truth with those in the retrieved chest CT volume. The Average Precision at $K$ (AP@K) is defined as follows:

\begin{center}
\begin{math}
\text{AP@K} = \frac{1}{K}\sum_{i=1}^{K} P_i \times R_i
\end{math}
\end{center}

Here, $P_i$ is the precision at the i-th position, and $R_i$ is the relevance at the i-th position. The final class retrieval accuracy is given by:

\begin{center}
\begin{math}
\text{MAP@K} = \frac{1}{n}\text{AP@K}
\end{math}
\end{center}

Where $n$ is the total number of samples. It is important to note that as $K$ increases, MAP@K may decrease due to the inclusion of less relevant images. We use the vision encoders of the supervised baseline, CT-CLIP, and the finetuned models for volume-to-volume retrieval and assess their performance across both internal and external validation sets using MAP@1, MAP@5, MAP@10, and MAP@50 metrics. The Recall@K metric is not applied in image-to-image retrieval due to the inherent limitation that the cosine similarity for identical images would inevitably be the highest, resulting in a consistent Recall@K value of $1$. However, this issue does not arise in text-to-image retrieval since the query is based on text embeddings rather than image embeddings.

For report-to-volume retrieval, we use the pretrained text transformer and 3D vision transformer to generate vector embeddings for both radiology reports and CT volumes, which are transformed from the token space to the embedding space with the CT-CLIP's trained projection layers (Figure \hyperref[fig:fig1]{1c}). Our goal is to find the CT volume that best matched each query report. To achieve this, we compute the cosine similarity between the latent representations of each report and all the CT volumes. Subsequently, we rank the CT volumes based on their cosine similarity scores to identify and return the most relevant matches. The process of report-to-volume retrieval is depicted in Figure \hyperref[fig:fig6]{5b}.

For the evaluation of this task, we employ the Recall@K metric, where $K$ represents the number of returned images. This method serves as a standard evaluation tool in text-to-image retrieval tasks \cite{zhang2012query}. We assess four Recall@K metrics: Recall@5, Recall@10, Recall@50, and Recall@100. These metrics measure the frequency of the target image appearing among the top 5, 10, 50, and 100 retrieved images, respectively. It is worth noting that the external validation set lacks text reports. Therefore, we restrict the report-to-volume retrieval evaluation to the internal validation set. We assess both CT-CLIP and the finetuned model with VocabFine for this task. However, since the supervised model and ClassFine do not generate latent embeddings for reports, they are not used in the report-to-volume retrieval.

\subsection*{Creating the dataset for CT-CHAT}
\label{ctvqa_methods}
To train CT-CHAT, we curate a comprehensive visual question-answering (VQA) dataset derived from free-form reports and chest CT volumes, using the same train-validation split as CT-RATE. The dataset is composed of four main components, with the following details for the training set.

The first component is the long-answer questions, which are further divided into four subcategories: conversation-based questions, description questions, free-response questions, and text-only questions. The conversation, description, and free-response questions are generated from radiology reports, while the text-only questions are created using 48,343 articles related to chest CT, retrieved from the PubMed Central-OA (\url{www.ncbi.nlm.nih.gov/pmc/tools/openftlist}). In total, the long-answer component of the training set includes 281,689 conversations, with a total of 518,053 conversation questions, 138,104 description questions, 138,061 free-response questions, and 250,760 text-only questions. All questions are generated using the Llama 3.1 8B large language model~\cite{llama3.1}, following the methodology described in \cite{pathchat}. The scripts for generating the question-answer pairs are available in the GitHub repository, and the prompts are provided in Supplementary Table~\hyperref[fig:vqa_generation]{7}. The second component consists of multiple-choice questions, also generated by Llama 3.1 8B \cite{llama3.1}, following the same approach described in \cite{pathchat}. This component of the training set includes a total of 46,911 multiple-choice conversations, comprising 138,608 questions in total. The third component includes short-answer questions derived from radiology reports in the CT-RATE dataset using a rule-based approach, as described in \cite{radgenome}. These questions cover categories such as location, presence, abnormality, and disorder. A total of 1,417,828 short-answer questions are sampled to create 318,710 short-answer conversations in the training set. The fourth component involves radiology report generation questions, totaling 47,149 conversations in the training set, where each conversation contains a single report generation question along with the corresponding generated report. The report generation questions are randomly selected from a predefined set of questions (Supplementary Table \hyperref[fig:report_generation_questions]{6}). 

The validation set is curated similarly to the training set, using the free-form reports from the CT-RATE validation set. The validation dataset includes 33,326 conversation questions, 8,886 description questions, 8,888 free-response questions, 12,013 short-answer questions, 8,915 multiple-choice questions, and 3,039 report-generation questions. Extended Data Figure \hyperref[fig:vqa_dataset]{7} shows example entries from the described dataset. Since the VQA dataset is generated by a large language model, we carefully select only accurately generated and properly formatted data, that is, data that can be successfully parsed as JSON for training our models. Additionally, we limit the initial number of generated questions to maintain a reasonable training duration. Our VQA dataset is made publicly available alongside CT-RATE.

\subsection*{Developing the CT-CHAT model}
\label{ctchat_methods}

Adapted from the LLaVa framework \cite{llava}, CT-CHAT integrates the frozen CT-CLIP vision encoder, a multimodal projector module, and a large language model (LLM) to enhance multimodal processing capabilities. The visual encoder compresses 3D chest CT volumes into a low-dimensional feature space. The multimodal projector module then transforms these visual features into the LLM’s embedding space, aligning it with text tokens. The LLM processes these visual-language tokens as input to generate contextual responses. Figure \hyperref[fig:fig6ctchat]{6a} summarizes the architecture of the CT-CHAT model.  

In our implementation, we evaluate and compare four different LLMs. For the primary configuration of CT-CHAT, we use the 70B Meta Llama 3.1 \cite{llama3.1}, one of the state-of-the-art open-source LLMs. This model consists of 80 transformer layers, 64 attention heads, an embedding dimension of 8,192, an FFN dimension of 28,672, and supports a context size of up to 128k tokens. For a smaller model, we use the 8B Meta Llama 3.1 \cite{llama3.1}, which includes 32 transformer layers, 32 attention heads, an embedding dimension of 4,096, an FFN dimension of 14,336, and a context size of up to 128k tokens. Additionally, for comparison, we implement two LLMs used in the official LLaVA 1.6 \cite{llava1.6}: Vicuna 13B and Mistral 7B. Figure \hyperref[fig:fig6ctchat]{6d} shows the scores for each of these LLMs. To dynamically adjust the model’s output mode based on the specific task, we add four special tokens to the tokenizers of each model: \texttt{<long\_answer>}, \texttt{<short\_answer>}, \texttt{<report\_generation>}, and \texttt{<multiple\_choice>}. These tokens are selected for different tasks during training and inference. The multimodal projector, based on the approach described in \cite{pathchat}, consists of an attention-pooling layer followed by a 2-layer multi-layer perceptron (MLP). The attention-pooling mechanism uses 256 learned latent queries in conjunction with multi-headed cross-attention to compress the final layer’s features from the encoder backbone into a fixed-length sequence of image tokens. This approach optimizes both training and inference, ensuring that the sequence length remains within the LLM’s context window, and making the model focus more on the important features with the attention mechanism. The MLP that follows is adapted from LLava 1.6 \cite{llava1.6} and includes a single hidden layer activated by GeLU, transforming the image tokens to match the LLM’s embedding dimension.

Training CT-CHAT involves two steps. First, in the medical concept alignment step, the LLM is frozen, and the multimodal projector is trained with the input “Please provide the radiology report for the following chest CT volume \texttt{<image>}" and the output report. This step improves subsequent finetuning performance and helps the multimodal projector to be aligned in biomedical concepts \cite{llavamed}. We conduct an ablation study of this step to show its effect in Supplementary Table \hyperref[fig:ctchat_ablations_supp]{10b}. In the second step, medical instruction tuning, the model is trained with the CT-VQA dataset that includes long answer, short answer, multiple choice, and report generation question-answer pairs. During this step, the multimodal projector, along with Low-Rank Adaptation (LORA) weights \cite{lora} for the LLM with rank 128 and alpha 256, is finetuned. \texttt{<image>} tokens are added to indicate the embedded image input to the model, except for text-only questions and answers. The four special tokens are added to the respective question-answer pairs in the dataset, enabling the model to learn different output modes. These tokens are later used during inference to select different modes of the model. The hyperparameters for each model for pretraining and finetuning are provided in Supplementary Table \hyperref[fig:supp hyperparameters vqa]{4a}, \hyperref[fig:supp hyperparameters vqa]{4b}. To boost report-generation accuracy, we finetune CT-CHAT with structured metadata extracted exclusively from pulmonary-nodule masks.  Each CT-RATE volume is processed with TotalSegmentator~\cite{Wasserthal2022TotalSegmentatorRS}, from which we record, per nodule, the lobar location, centroid, orthogonal diameters, and volume (mm\textsuperscript{3}).  These fields are embedded in a fixed prompt:

\begin{quote}
\texttt{<volume\_tokens>} \\
\texttt{Generate a detailed radiology report for the given chest CT volume.} \\
\texttt{<BEGIN\_NODULE\_INFO>} \\
\texttt{Nodule\_Count: \emph{n}} \\
\texttt{Nodule\_ID: 1 \quad Lobe: \emph{left lower}} \\
\texttt{\hspace{1em}Centroid\_mm: (\emph{x},\emph{y},\emph{z})} \\
\texttt{\hspace{1em}Diameter\_mm: (x:\emph{d\textsubscript{x}}, y:\emph{d\textsubscript{y}}, z:\emph{d\textsubscript{z}})} \\
\texttt{\hspace{1em}Volume\_mm3: \emph{v}} \\
\texttt{\dots\ (repeat for each nodule)} \\
\texttt{<END\_NODULE\_INFO>}
\end{quote}

Each sample includes (i) CT volume features from the frozen CT-CLIP encoder, (ii) the structured nodule template, and (iii) the reference report. Finetuning is performed on the 8B Llama 3.1 configuration of CT-CHAT using LoRA (rank 128, $\alpha=256$); only the LoRA weights and projector are updated. Supplying these segmentation-derived cues markedly improves performance (Figure~\hyperref[fig:fig6ctchat]{6e}). The same workflow can incorporate additional abnormalities as more 3D segmentation models become available.

\subsection*{Benchmarking the CT-CHAT model}
\label{benchmarking_ctchat_methods}

Given the lack of pretrained VQA models for 3D imaging, we conduct a comparative analysis of CT-CHAT against four 2D state-of-the-art open-source VQA model: LLaVA 1.6 (Mistral 7B) \cite{llava1.6}, LLaVA 1.6 (Vicuna 13B) \cite{llava1.6}, LLaVA-Med \cite{llavamed}, and CXR-LLaVA \cite{lee2023cxr}, using benchmarking framework established in \cite{pathchat}. To address the challenge of representing 3D information using their 2D encoders and to ensure a fair comparison, we use Digitally Reconstructed Radiographs (DRRs) derived from CT-RATE, as suggested by \cite{hou2024shadow}. This approach is chosen over the selection of random or central CT slices, which can not adequately capture all relevant pathologies. The use of DRRs is justified, as \cite{hou2024shadow} demonstrates that models trained with DRRs show comparable results to those trained with 3D chest CTs.

To evaluate the baseline models and CT-CHAT models with different LLMs, we employ four NLP metrics: BLEU-1 \cite{bleu}, METEOR \cite{meteor}, ROUGE-L \cite{rouge}, and CIDEr \cite{cider}, along with the Llama score, which measures the clinical accuracy of the generated responses compared to the ground truth, using the Llama 3.1 70B model \cite{llama3.1}. A similar LLM-based clinical accuracy measurement approach is used in Llava-Med with the GPT-4 API \cite{llavamed}. Supplementary Table \hyperref[fig:llama-score]{8} presents the system prompt for the Llama 3.1 model, which is used to score the generated answers. We calculate scores task specifically: long answer, short answer, multiple choice, and report generation. Figure \hyperref[fig:fig6ctchat]{6c} illustrates mean and task-specific scores for each model. To visualize the clinical accuracy of the models, we categorize the Llama scores for every question in each task into three ranges: 0-3 for low clinical accuracy, 4-6 for average clinical accuracy, and 7-10 for high clinical accuracy, and create Sankey plots within these ranges in Supplementary Figure \hyperref[fig:vqa_extended2]{10}.

In addition to benchmarking CT-CHAT on VQA tasks, we further evaluate its performance in radiology report generation, a task more reflective of routine radiological workflows, by comparing it with two state-of-the-art 3D report-generation models: RadFM~\cite{radfm} and CT2Rep~\cite{Hamamci2024CT2Rep}. While VQA metrics such as BLEU, ROUGE, and the Llama score assess linguistic and clinical quality, they do not fully capture diagnostic use. Therefore, for this comparison we employ clinical-efficacy metrics: precision, recall, F1, and the CRG score~\cite{Hamamci2025CRGSA}. CRG weights findings by prevalence and severity, giving a distribution-aware score aligned with radiologists’ priorities. Metric computation uses our automated report labeler, which extracts structured abnormality labels from the generated reports. Evaluation is conducted on both the internal CT-RATE validation set and the external RAD-ChestCT dataset; for RAD-ChestCT, which lacks text reports, we leverage its binary abnormality annotations to enable indirect evaluation through our labeler. Notably, we use the original pretrained weights for both RadFM and CT2Rep without retraining.

To complement automatic metrics, we performed a blinded, three-reader study of report quality. Two board-certified thoracic radiologists from Zurich (R1, R2) and one from Istanbul (R3) independently rated reports generated by CT-CHAT, RadFM, and CT2Rep. Because free-text scoring is time-intensive, we limited the pool to 150 CT-RATE validation cases per model. Cases were stratified rather than randomly sampled: after computing Llama scores for CT-CHAT outputs, we selected 50 volumes from each accuracy band (low 0–3, medium 4–6, high 7–10) to ensure balanced difficulty. For every selected scan, the three anonymized model outputs were presented, and readers scored each report on a 0–10 scale against predefined criteria covering finding description, localisation, terminology, and overall clinical adequacy. Figure \hyperref[fig:fig6ctchat]{6d} shows the results: CT-CHAT achieved mean scores of 3.44 (R1), 4.10 (R2), and 4.52 (R3), outperforming RadFM (0.55, 0.91, 0.86) and CT2Rep (2.66, 2.52, 2.74). Violin plots display a right-shifted distribution for CT-CHAT, confirming superior perceived clinical fidelity. The same protocol can be repeated for larger or external cohorts as new models emerge. It is important to note CT-CHAT’s superiority over CT2Rep, even though both models are trained on the same CT-RATE dataset. Moreover, CT2Rep is optimized solely for report generation, while CT-CHAT is tuned primarily for VQA tasks, classification, question-answering, and only secondarily for report generation. Despite this, CT-CHAT consistently outperforms CT2Rep across all evaluations. The performance gap underscores the value of contrastive pretraining with CT-CLIP in capturing clinically meaningful 3D CT representations that transfer effectively to downstream tasks, even when the model is not explicitly trained for them.

To evaluate CT-CHAT's ability to retrieve information from chest CT, we conduct an ablation study for the image provision, as shown in Supplementary Table \hyperref[fig:ctchat_ablations_supp]{10a}. Since the model is expected to hallucinate pathologies when the input volume is not provided, we implement a guardrail for prevention with the system prompt in Supplementary Table \hyperref[fig:ctchat_system]{9}. Supplementary Figure \hyperref[fig:guardrail_examples]{11} provides an example of the guardrail.

\subsection*{Computational hardware and software}
\label{hardware_methods}
Experiments for CT-CLIP are conducted using Python 3.11.5 with PyTorch 2.0.1, CUDA 11.7, and other libraries like SciPy 1.11.1, Torchvision 0.15.2, Scikit-learn 1.3.0, Pandas 2.0.3, NumPy 1.24.4, Transformers 4.30.1, and Accelerate 0.21.1. For CT-CHAT, Python 3.12.4 is used with PyTorch v2.4.0, CUDA v12.4, SciPy v1.14.0, Torchvision v0.19.0, Scikit-learn v1.2.2, Pandas v2.2.2, and NumPy v1.26.4. We use four 80GB A100 GPUs, 500GB RAM, and 24 CPUs, mainly for CT-CLIP training with Distributed Data-Parallel and Fully Sharded Data Parallelism. Finetuning of CT-CLIP and other experiments is performed on a single A100 GPU. CT-CHAT with the Llama 3.1 70B model is trained on four A100 GPUs using DeepSpeed ZeRO-3, while other CT-CHAT models use two GPUs. Hyperparameters are in Supplementary Tables \hyperref[fig:supp hyperparameters]{3}, \hyperref[fig:supp hyperparameters vqa]{4}. All software details are available on our respective GitHub repositories.

\subsection*{Data availability}
\label{dataav_methods}
Our CT-RATE dataset, comprising paired chest CT volumes and radiology text reports, is publicly accessible via \url{https://huggingface.co/datasets/ibrahimhamamci/CT-RATE}~\cite{Hamamci2025_CT_RATE}. The external validation set, RAD-ChestCT, is also openly accessible through \url{https://zenodo.org/records/6406114}~\cite{Rahman2022_RAD_ChestCT}.
\subsection*{Code availability}
Our trained models and codebase are publicly available at \url{https://github.com/ibrahimethemhamamci/CT-CLIP}~\cite{Hamamci2025_CT_CLIP} and \url{https://github.com/ibrahimethemhamamci/CT-CHAT}~\cite{Hamamci2025_CT_CHAT} for further research.

\subsection*{Acknowledgments}
\label{acknowledgments_methods}

We extend our sincere gratitude to the Helmut Horten Foundation for their generous support, which made this work possible. We also thank Istanbul Medipol University for their invaluable support and provision of data. Some of the HPC resources were provided by the Erlangen National High Performance Computing Center (NHR@FAU) at Friedrich-Alexander-Universität Erlangen-Nürnberg under NHR projects \texttt{b143dc} and \texttt{b180dc}. NHR is funded by federal and Bavarian state authorities, and the NHR@FAU hardware is partly funded by the German Research Foundation (DFG; grant 440719683). This research was supported in part by the Intramural Research Program of the National Library of Medicine (NLM), National Institutes of Health (NIH), and used the computational resources of the NIH high-performance computing Biowulf cluster. B.K. received research support from the ERC project MIA-NORMAL 101083647, DFG grants 513220538 and 512819079, and the state of Bavaria (HTA). C.B. received research support from the Promedica Foundation, Chur, Switzerland. C.W., W.D., and K.B. are supported by NIH grant R01 HL141813, the Hariri Institute for Computing’s Junior Faculty Fellows Program, and its Focused Research Program. Figures \ref{fig:fig1}, \ref{fig:fig2}, \ref{fig:fig6}, and \ref{fig:fig6ctchat} were created with \emph{BioRender.com}.

\subsection*{Author contributions}
\label{authorcont_methods}
I.E.H., S.E., and B.M. designed the study. I.E.H. and S.E. handled data collection, data analysis, model construction and validation, figure preparation, and manuscript writing. F.A., A.G.S., S.N.E., and I.D. managed data anonymization and annotation. M.F.D. contributed to model construction. C.W., W.D., and K.B. carried out the external evaluation. O.F.D. developed the graphical user interface and assisted with manuscript writing. M.X. and T.A. generated anatomical-segmentation labels. S.S. maintained the model repository. B.H. assisted in preparing NIfTI files with metadata and integrating nodule-segmentation masks into the report-generation pipeline. M.P. assessed report-generation performance. B.K. provided technical expertise and contributed to the manuscript writing. H.D., B.W., E.S., M.S., E.B.E., A.A., A.S., A.K., Z.L., B.L., C.B., and M.K.O. provided domain-knowledge support. B.M. supervised the entire study. All authors reviewed and approved the final manuscript.

\subsection*{Competing interests}
The authors declare no competing interests.

    % \end{refsegment}
   % \printbibliography[segment=2, heading=bibliography, title={Methods References}, notkeyword=cited-in-main]
\printbibliography[segment=1, heading=bibliography, title={References}]

    \newpage
    \appendix
    
    % =========================

\begin{figure}[p]
    \centering
    \includegraphics[width=\textwidth]{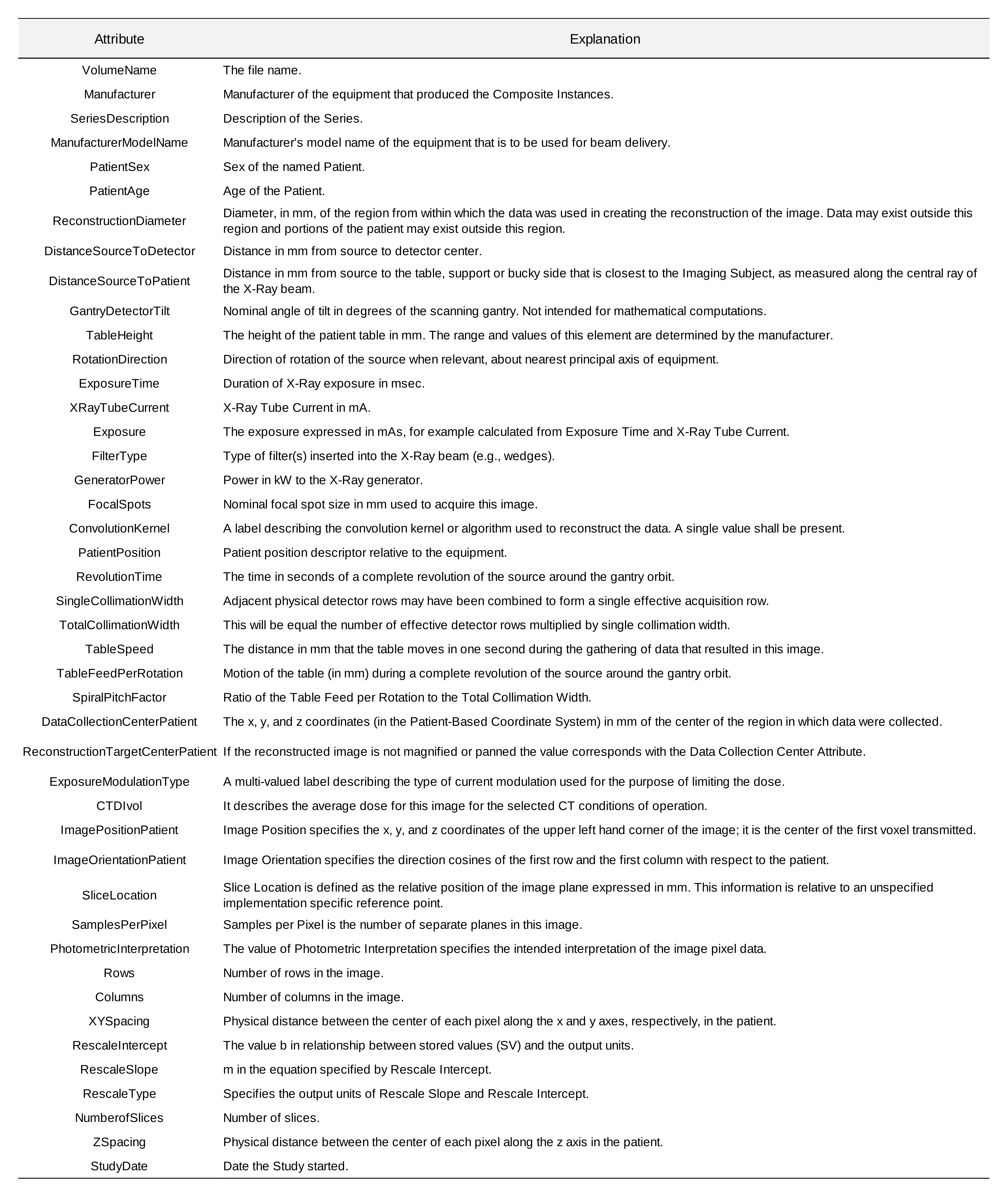}
    \caption{\textbf{Extended Data Table 1: Metadata attributes.} Detailed breakdown of the metadata attributes accessible within the CT-RATE dataset.}
    \label{fig:dataset_attributes}
\end{figure}

\begin{figure}[p]
    \centering
    \includegraphics[width=\textwidth]{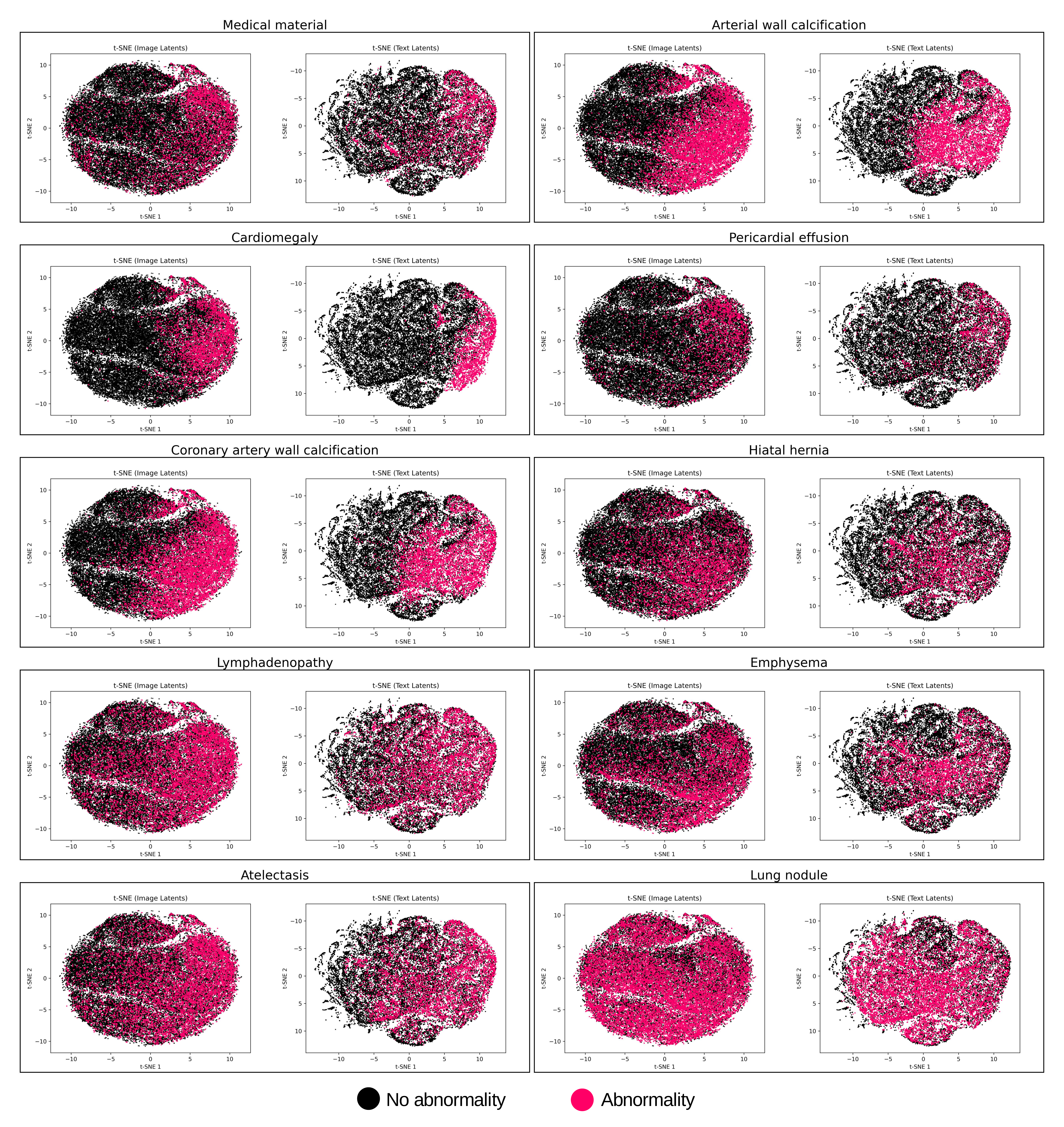}
    \caption{\textbf{Extended Data Figure 1: Abnormality-based t-SNE projections for chest CT volumes and radiology text reports.} }
    \label{fig:extfig4a}
\end{figure}

\begin{figure}[ht]
    \centering
    \includegraphics[width=\textwidth]{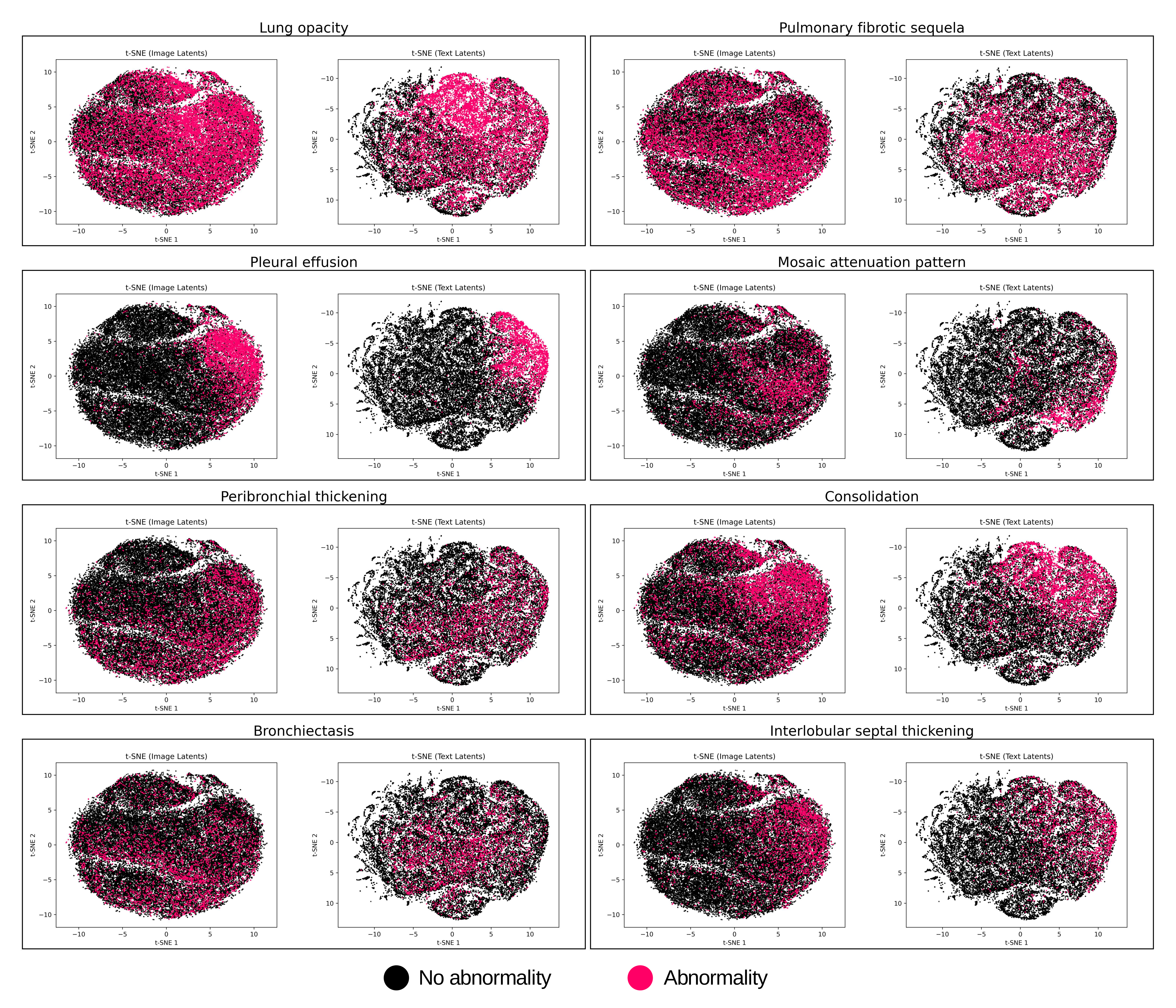}
    \caption{\textbf{Extended Data Figure 2: Abnormality-based t-SNE projections for chest CT volumes and radiology text reports (continued).} }
    \label{fig:extfig4b}
\end{figure}

\begin{figure}[ht]
    \centering
    \includegraphics[width=\textwidth]{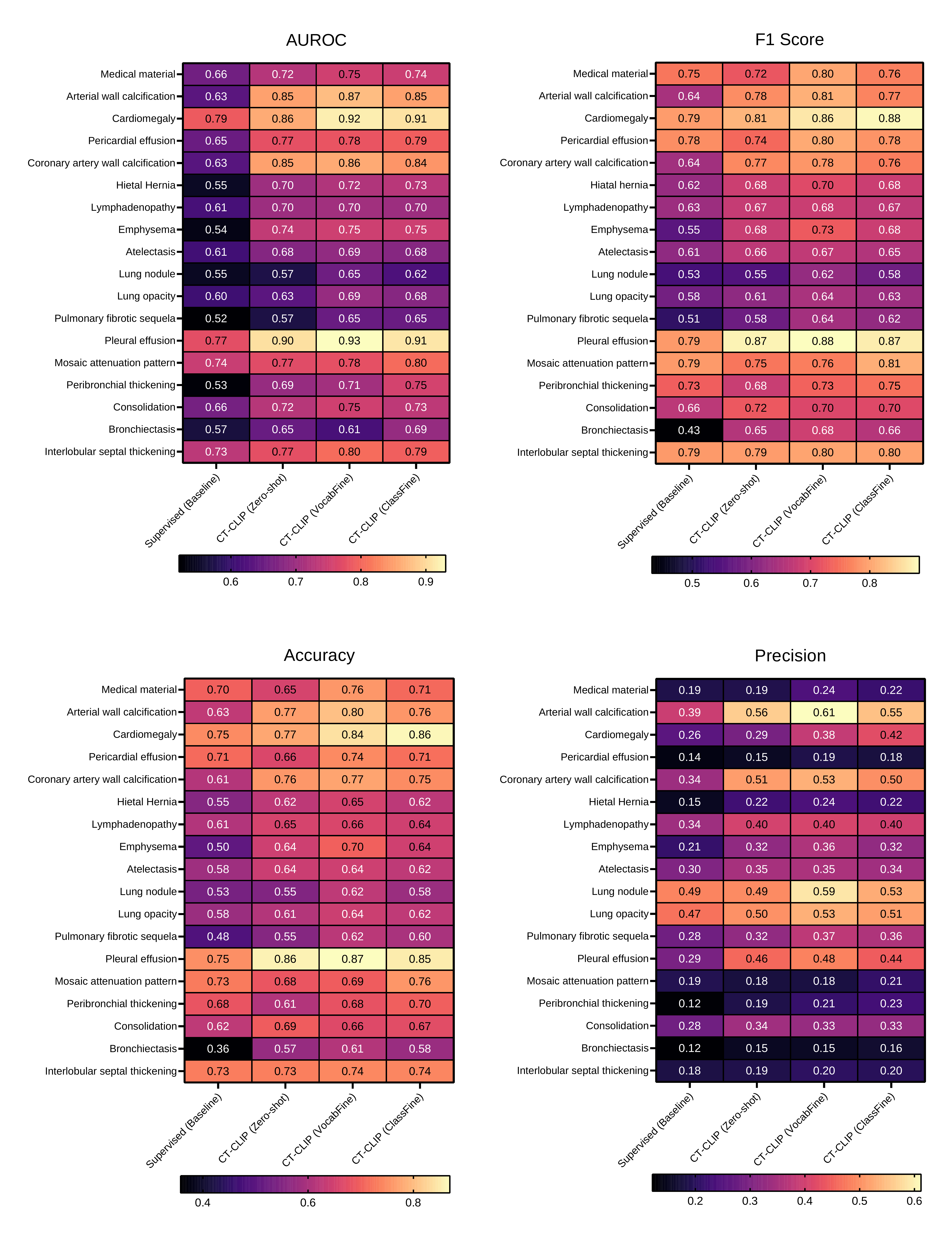}
    \caption{\textbf{Extended Data Figure 3: Comparison of abnormality-based performance metrics in the internal validation set.} This figure provides a detailed analysis of performance metrics, including AUROC, accuracy, precision, and F1 scores, for detecting various abnormalities with our models in the internal validation set, compared to the fully supervised baseline model. The proposed CT-CLIP based models demonstrate improvement over the baseline in almost all comparisons.}
    \label{fig:extfig5}
\end{figure}

\begin{figure}[ht]
    \centering
    \includegraphics[width=\textwidth]{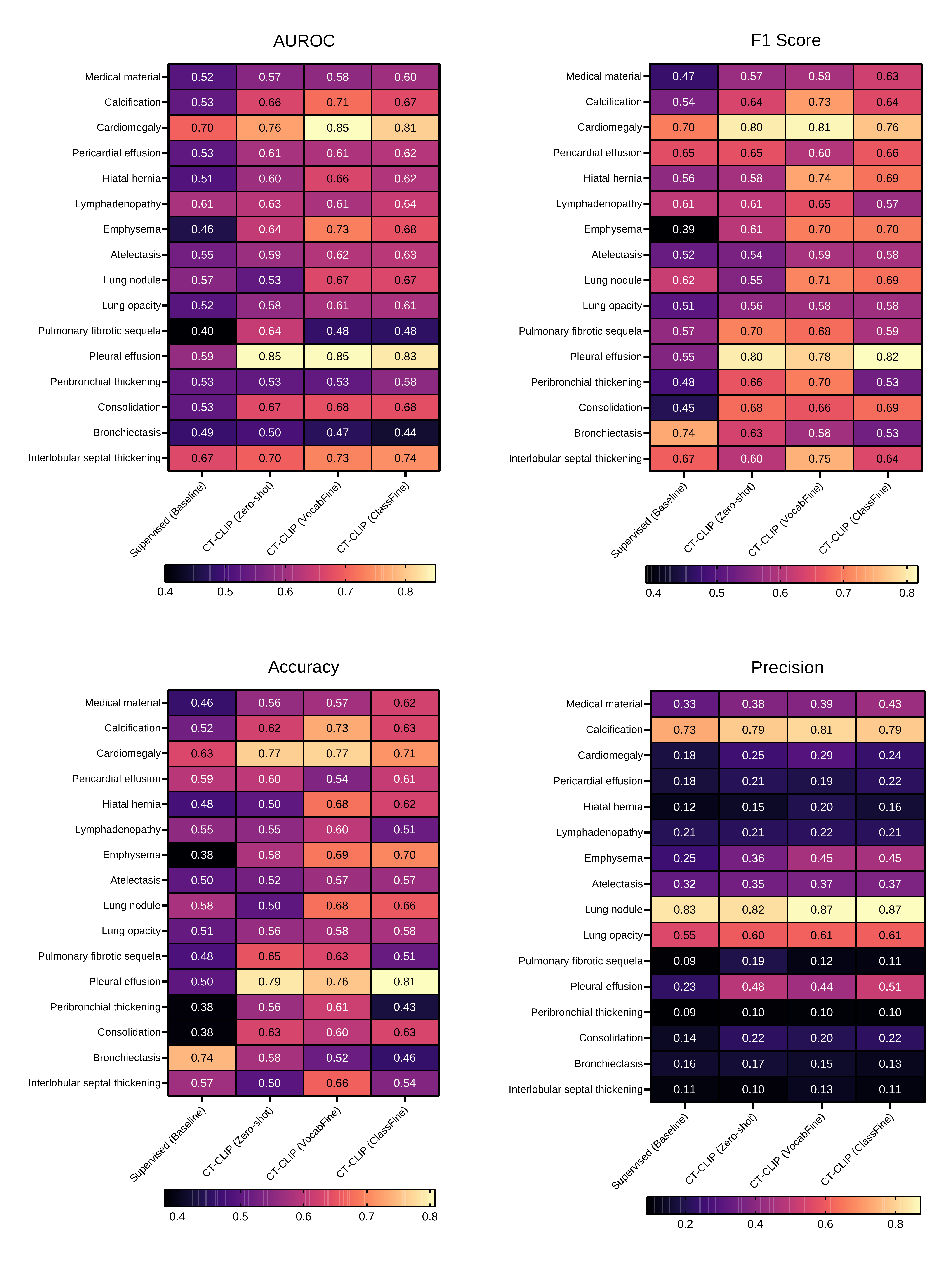}
    \caption{\textbf{Extended Data Figure 4: Comparison of abnormality-based performance metrics in the Rad-ChestCT validation set.} This figure provides a detailed analysis of performance metrics, including AUROC, accuracy, precision, and F1 scores, for detecting various abnormalities with our models in the Rad-ChestCT validation set, compared to the fully supervised baseline model. It highlights the models' remarkable adaptability and superior effectiveness with distribution shifts, setting a new standard in performance compared to a fully supervised baseline model.}
    \label{fig:extfig5external}
\end{figure}

\begin{figure}[ht]
    \centering
    \includegraphics[width=\textwidth]{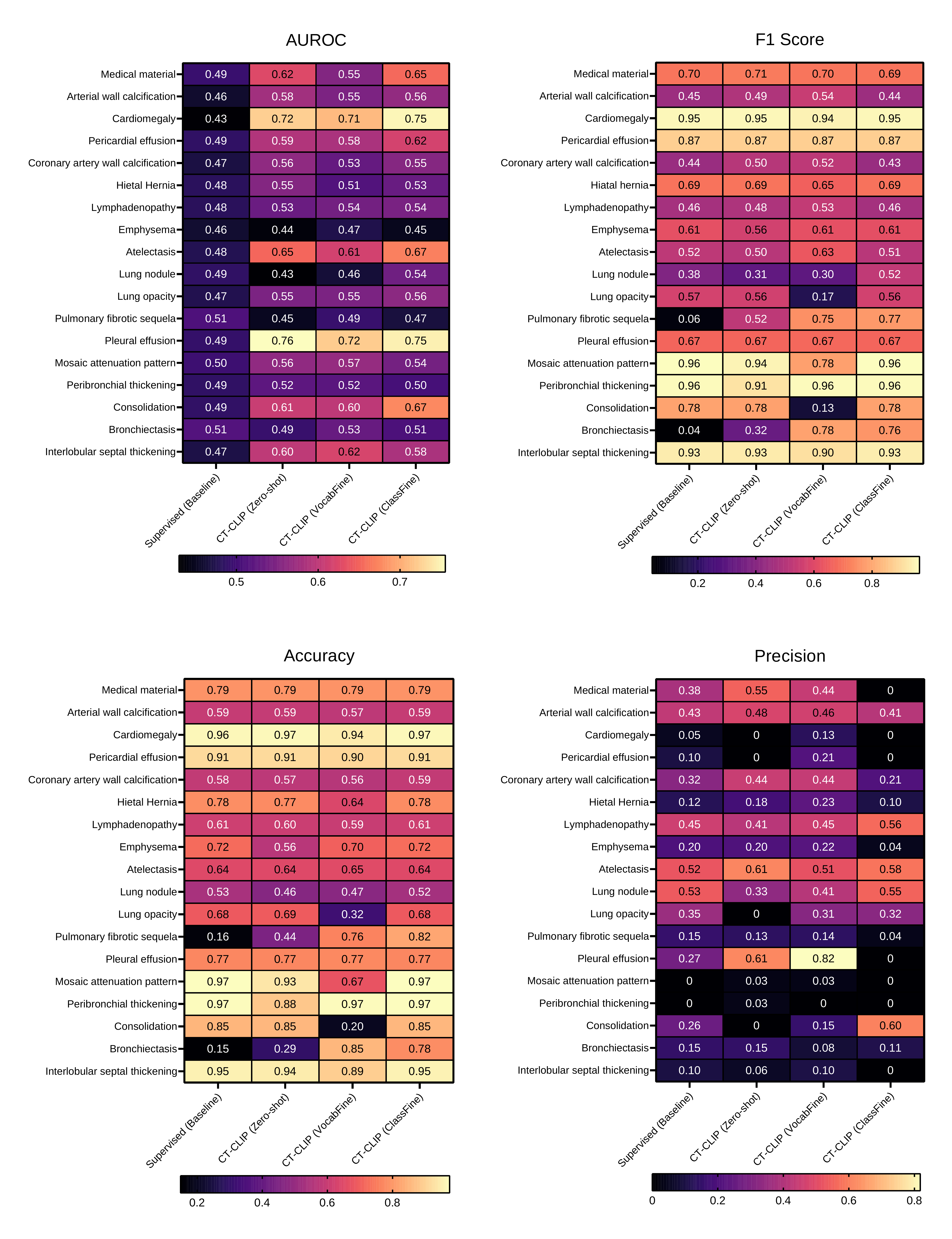}
    \caption{\textbf{Extended Data Figure 5: Comparison of abnormality-based performance metrics in the UPMC validation set.} This figure provides a detailed analysis of performance metrics, including AUROC, accuracy, precision, and F1 scores, for detecting various abnormalities with our models in the UPMC validation set, compared to the fully supervised baseline model. It highlights the models' remarkable adaptability and superior effectiveness with distribution shifts, setting a new standard in performance compared to a fully supervised baseline model.}
    \label{fig:extfig5external_upmc}
\end{figure}

\begin{figure}[ht]
    \centering
    \includegraphics[width=\textwidth]{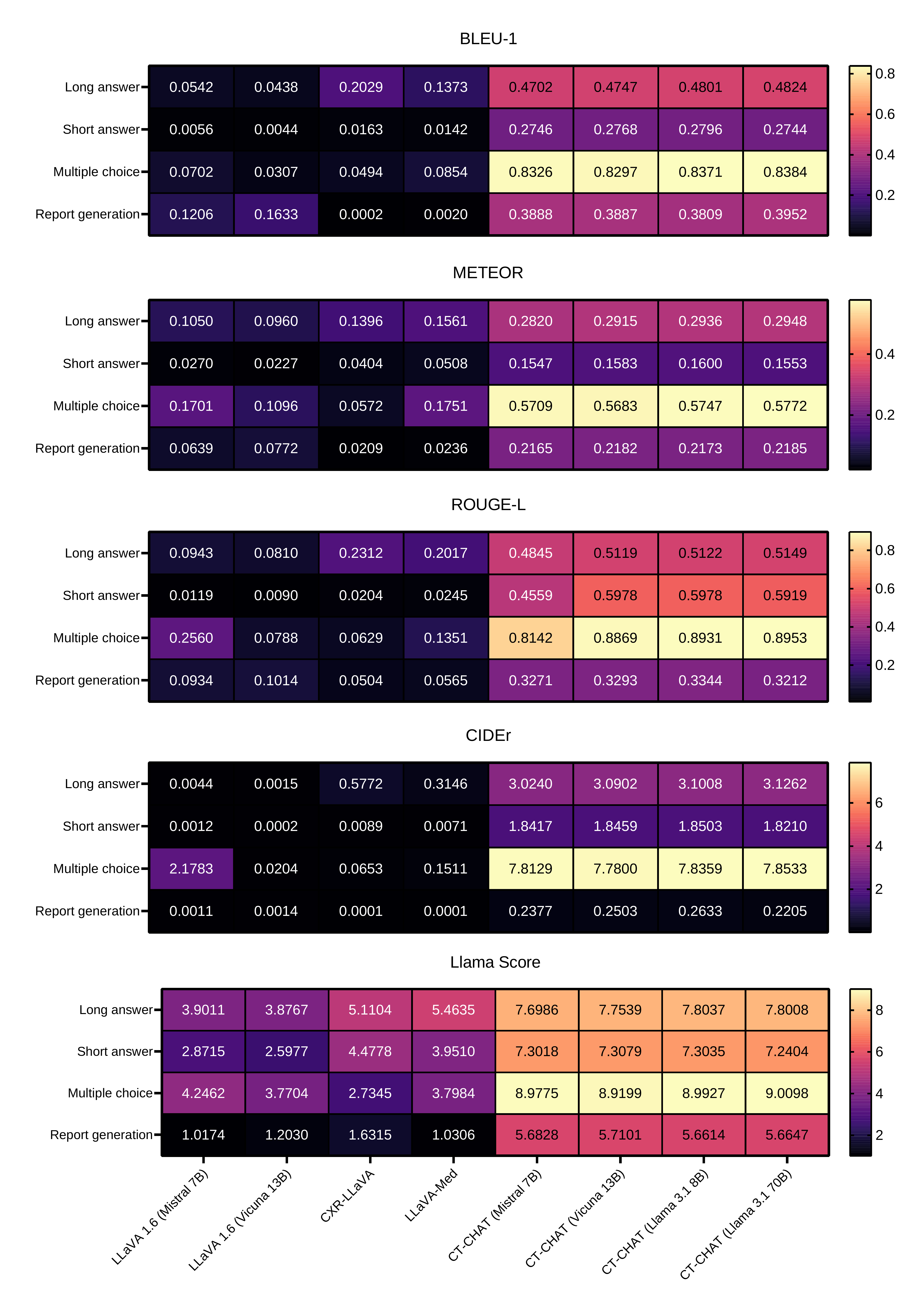}
    \caption{\textbf{Extended Data Figure 6: Comparative metrics for CT-CHAT and baselines.} This figure illustrates the performance of CT-CHAT and baselines across various tasks, demonstrating that CT-CHAT consistently outperforms the baselines, achieving the highest accuracy in every metric for each task.}

    \label{fig:vqa_extended}
\end{figure}

\begin{figure}[ht]
    \centering
    \includegraphics[width=0.9\textwidth]{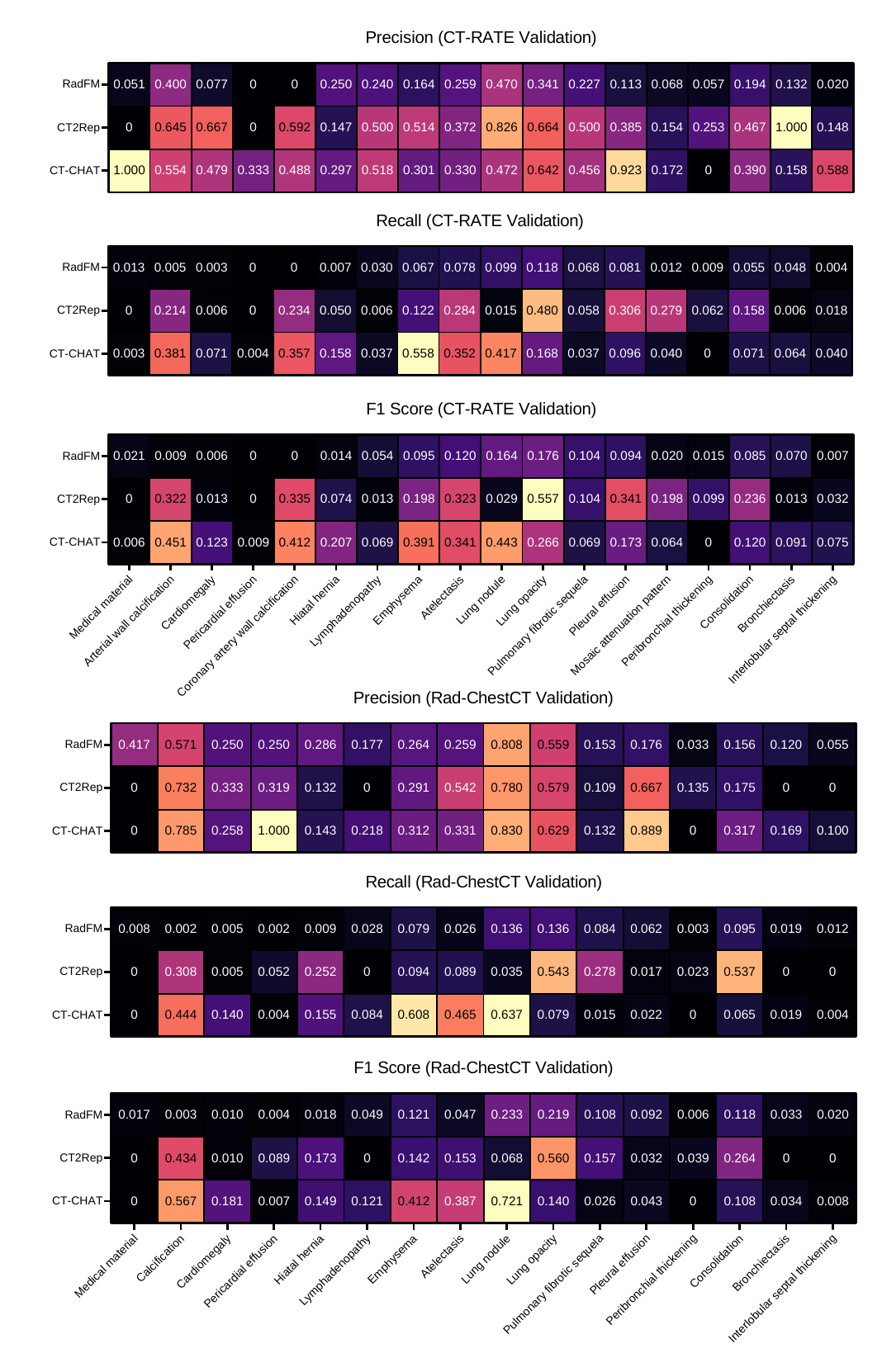}
    \caption{\textbf{Extended Data Figure 7: Comparative metrics for CT-CHAT and baselines in report generation.} This figure presents the report generation performance of CT-CHAT compared to baseline models across multiple abnormalities, evaluated on two independent validation sets. CT-CHAT consistently outperforms the baselines, demonstrating superior clinical accuracy.}

    \label{fig:report_extended}
\end{figure}

\begin{figure}[ht]
    \centering
    \includegraphics[width=\textwidth]{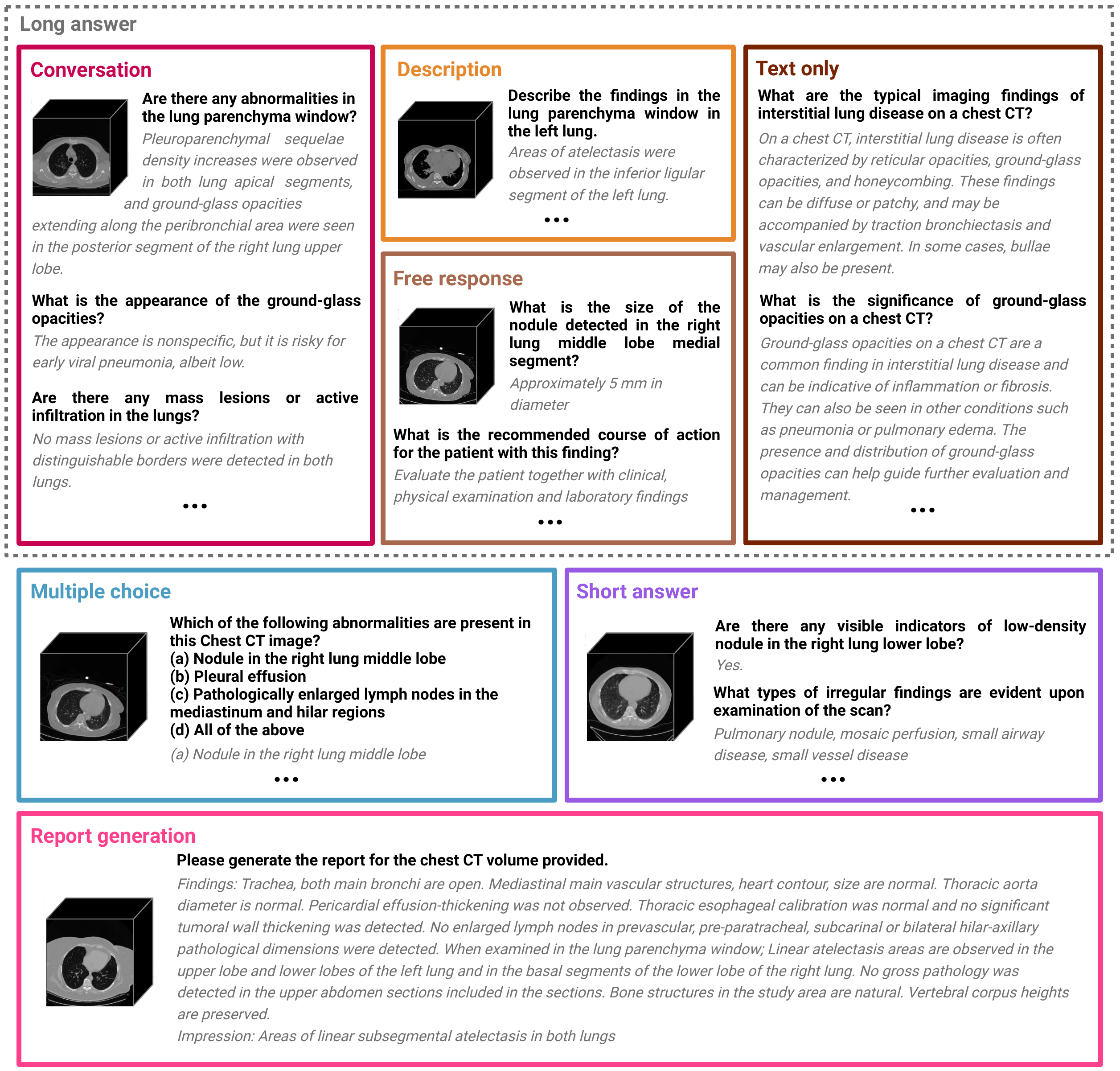}
    \caption{\textbf{Extended Data Figure 8: Representative examples from VQA.} This dataset includes various question types, such as long-answer (which encompass conversational, descriptive, free-response, and text-only questions), multiple-choice, short-answer, and report-generation questions.}
    \label{fig:vqa_dataset}
\end{figure}

\begin{figure}[ht]
    \centering
    \includegraphics[width=\textwidth]{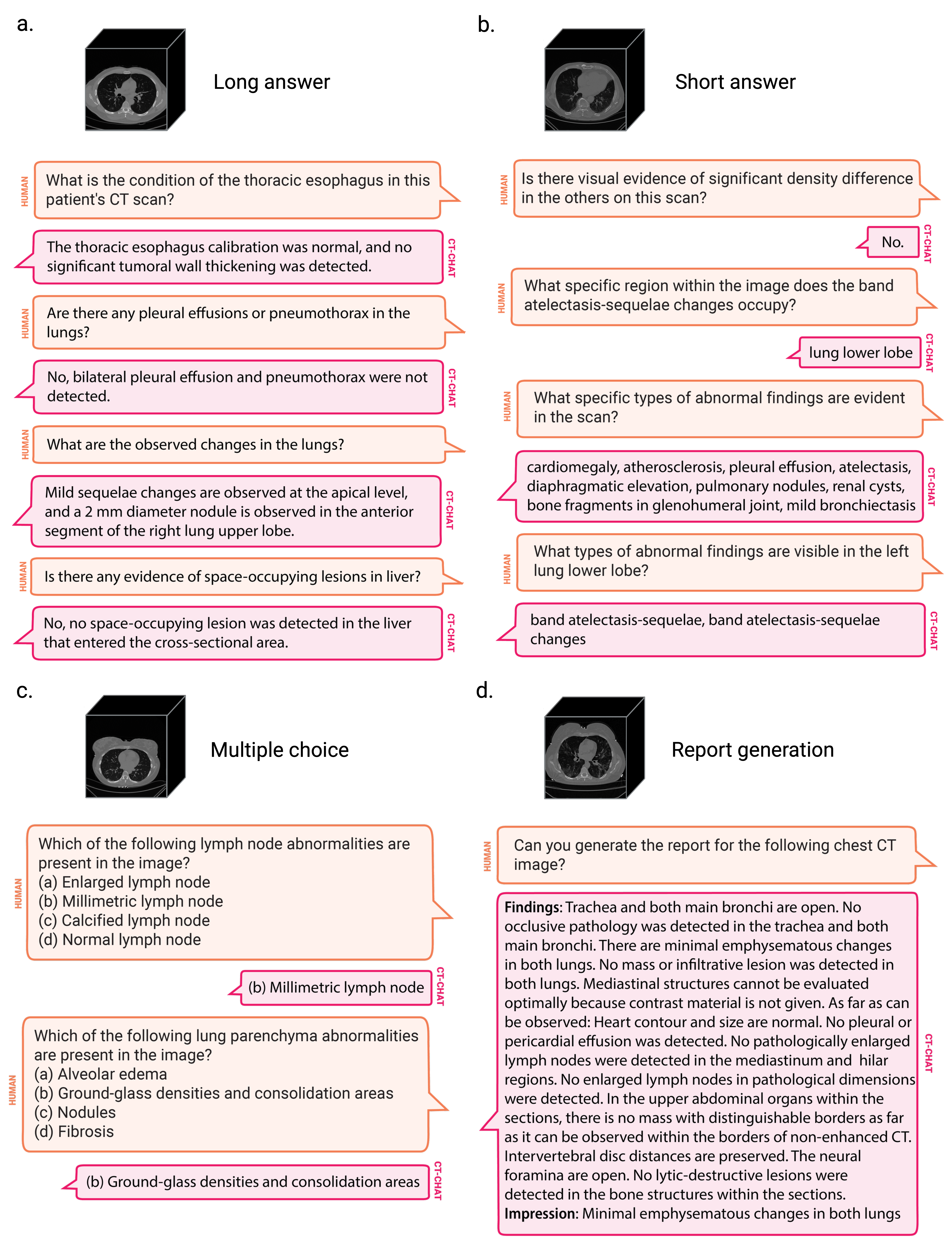}
    \caption{\textbf{Extended Data Figure 9: Example outputs of CT-CHAT for (a) long answer, (b) short answer, (c) multiple-choice, and (d) report generation questions.}}
    \label{fig:ctchat_examples_supp}
\end{figure}

\begin{figure}[ht]
    \centering
    \includegraphics[width=\textwidth]{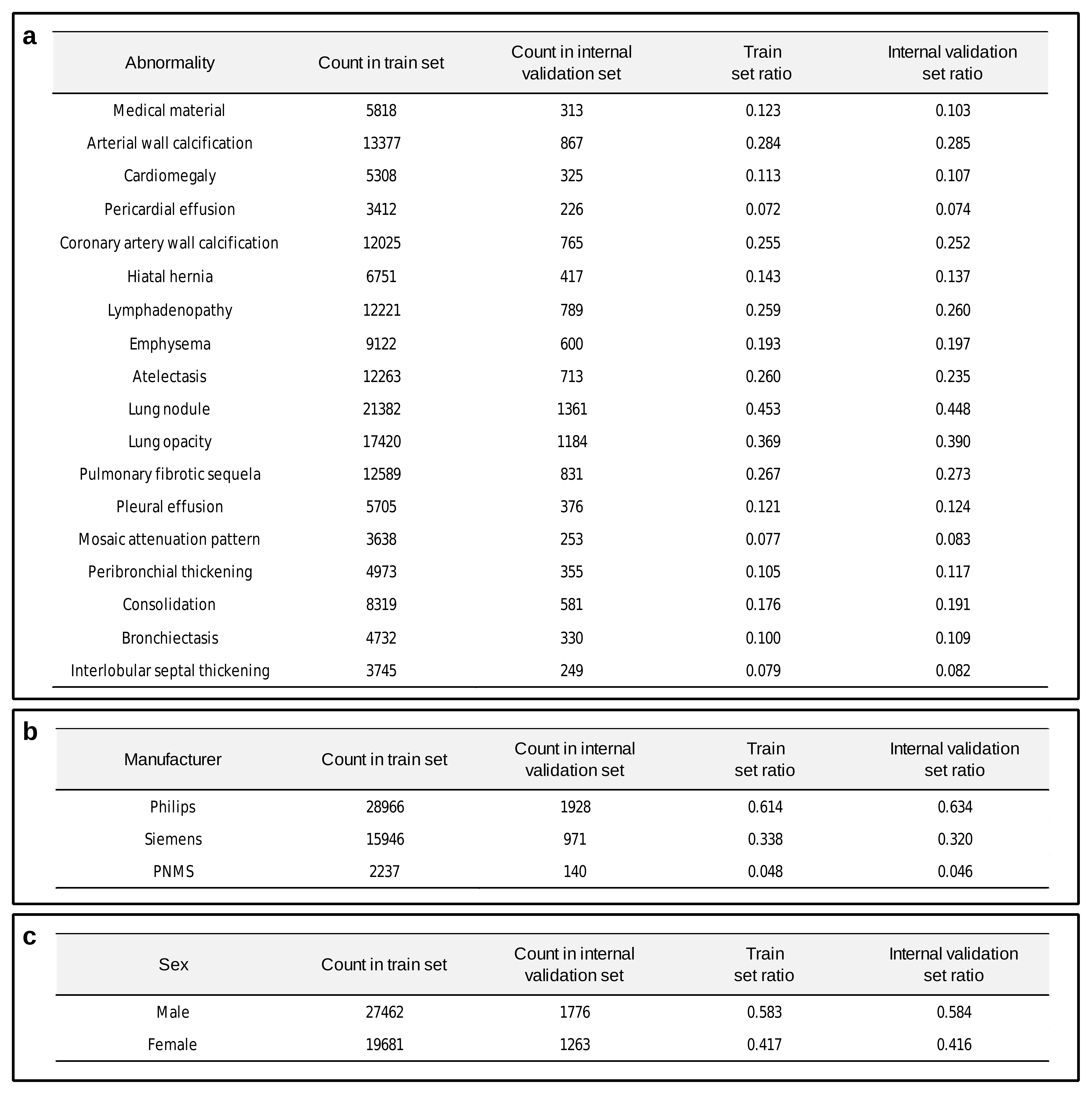}
    \caption{\textbf{Supplementary Table 1: Detailed overview of the CT-RATE dataset.} \textbf{a.} The counts and proportions of various abnormalities identified within the training and validation subsets, offer insights into dataset composition. \textbf{b.} Details of the distribution of chest CT volumes according to manufacturer across both subsets, reflecting the dataset's diversity. \textbf{c.} The breakdown of chest CT volumes by sex in both subsets, provides a demographic perspective of the data.}
    \label{fig:supp dataset}
\end{figure}

\begin{figure}[t]
    \centering
    \includegraphics[width=\textwidth]{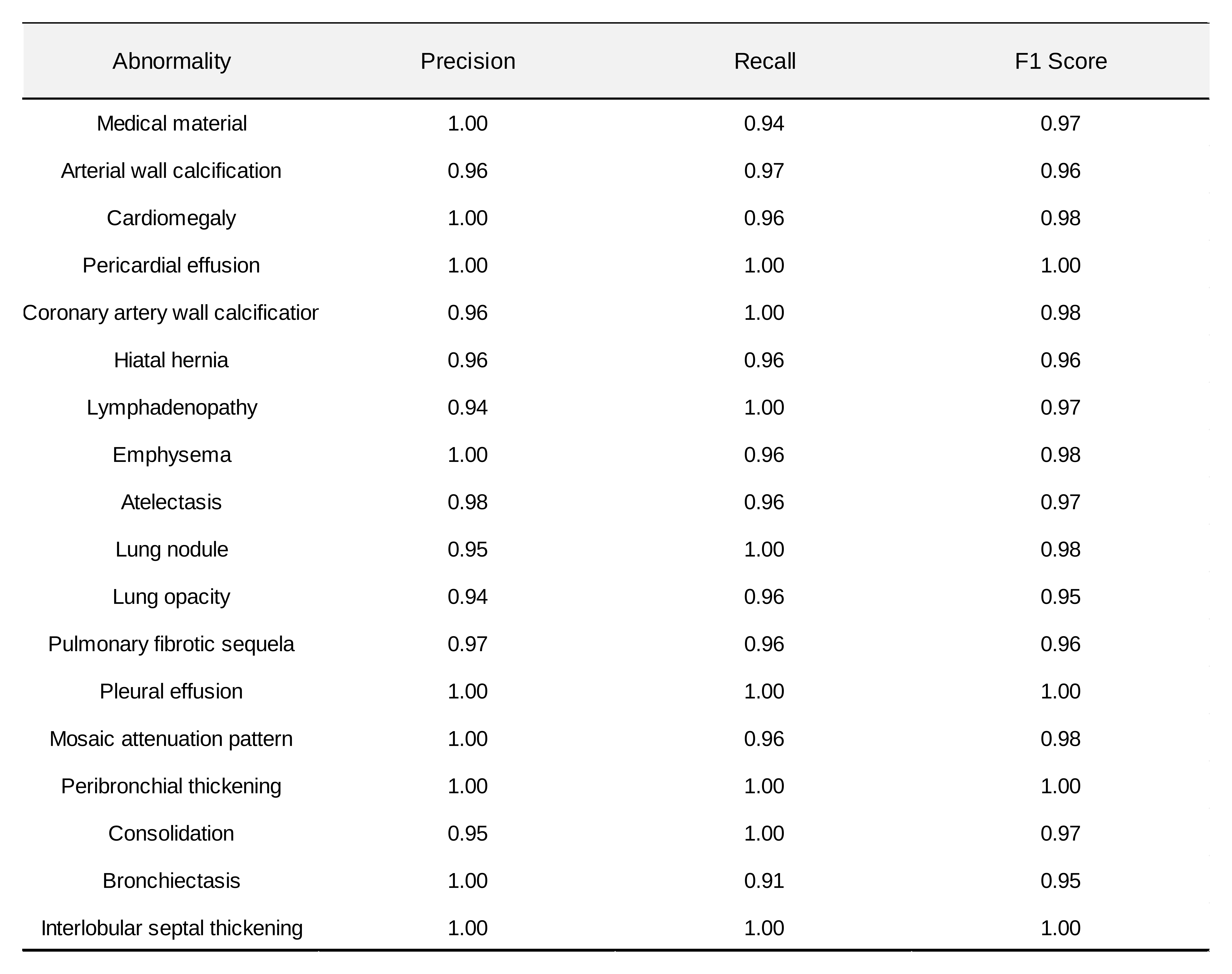}
    \caption{\textbf{Supplementary Table 2: Finetuned text encoder's performance in automatic label extraction.} This table presents a comprehensive assessment of performance metrics, including precision, recall, and F1 scores, for the finetuned text encoder across various abnormalities. The evaluation underscores the encoder's accuracy and efficiency in automating the extraction of labels for each abnormality.}

    \label{fig:supp classifier}
\end{figure}

\begin{figure}[b]
    \centering
    \includegraphics[width=\textwidth]{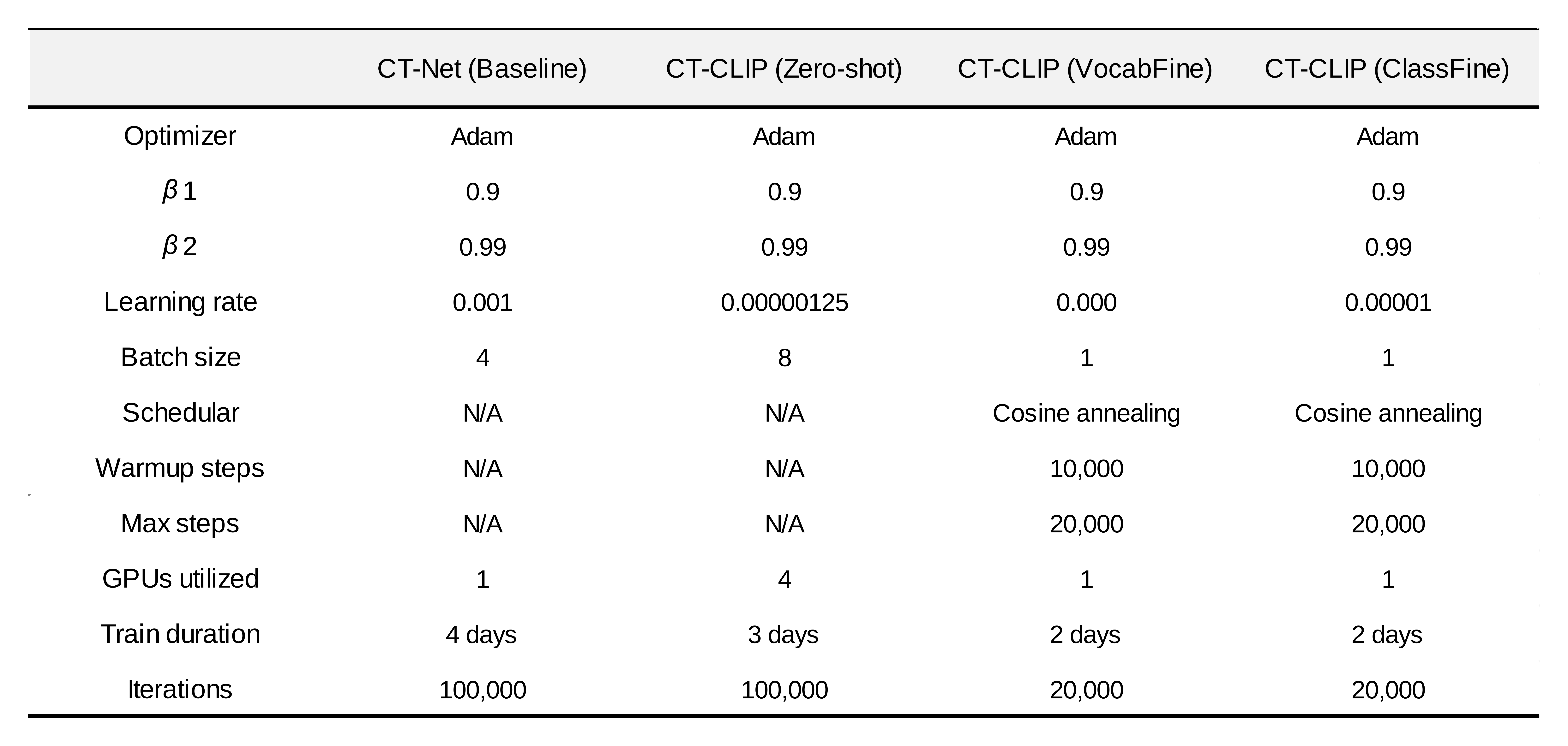}
    \caption{\textbf{Supplementary Table 3: Hyperparameters for training each model of the multi-abnormality detection experiments.}}
    \label{fig:supp hyperparameters}
\end{figure}

\begin{figure}[b]
    \centering
    \includegraphics[width=\textwidth]{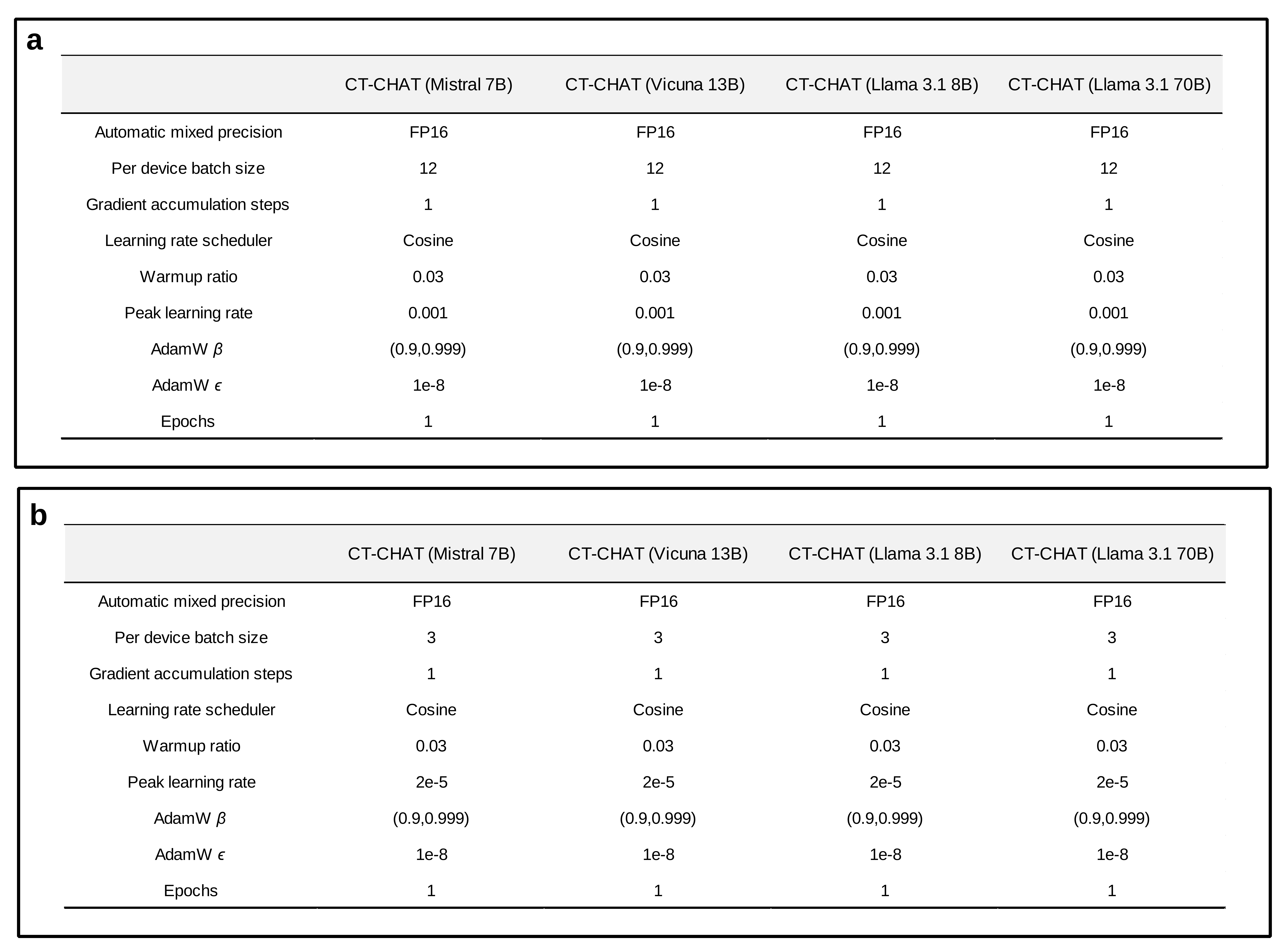}
    \caption{\textbf{Supplementary Table 4: Hyperparameters for (a) pretraining and (b) finetuning during each CT-CHAT ablation.} }
    \label{fig:supp hyperparameters vqa}
\end{figure}

\begin{figure}[b]
    \centering
    \includegraphics[width=\textwidth]{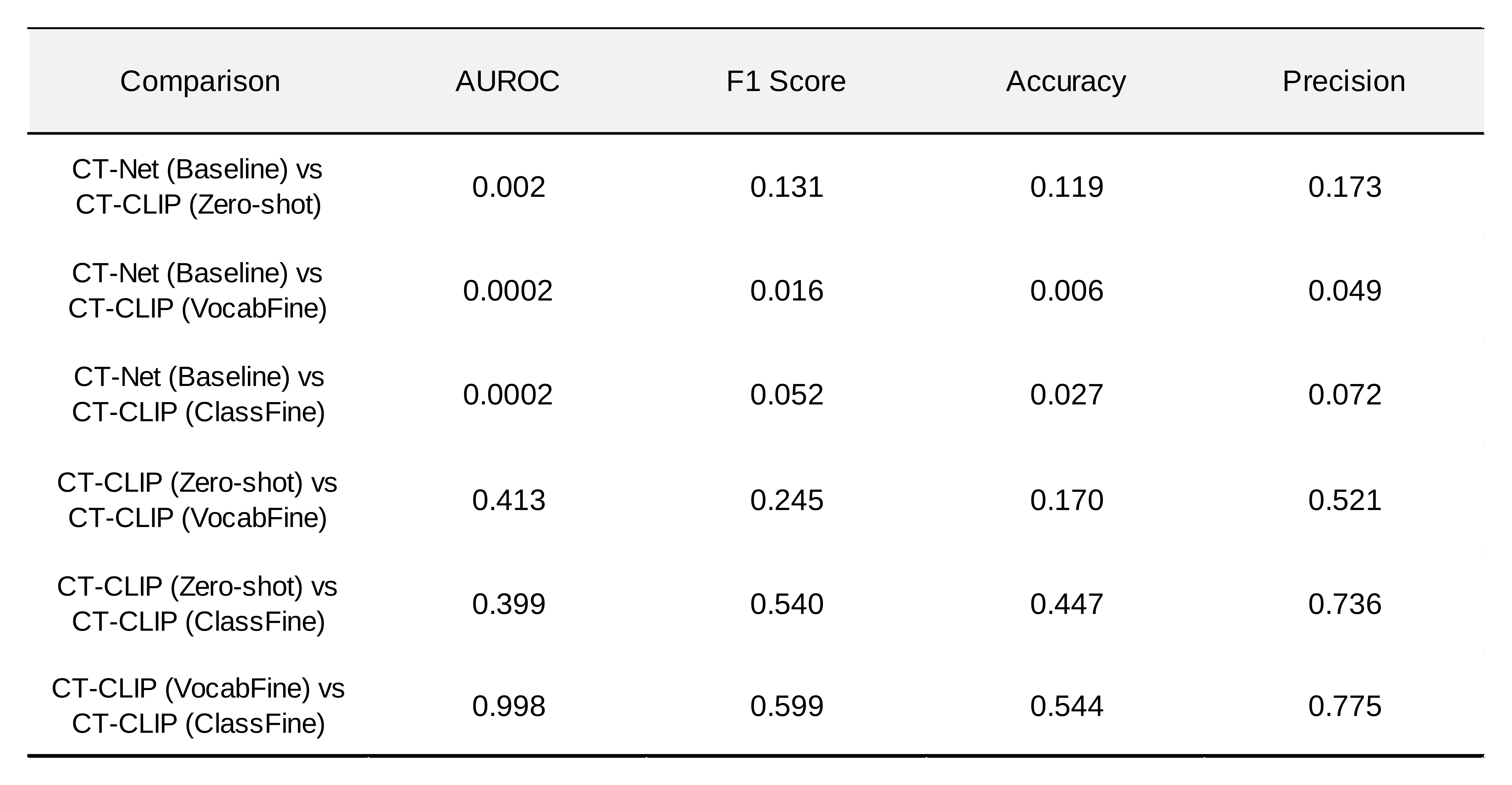}
    \caption{\textbf{Supplementary Table 5: Statistical significance.} Using a two-sided unpaired permutation test with 10,000 permutations, we observe statistically significant differences \((p < 0.05)\) in AUROC, F1 score, and accuracy between the supervised baseline model and each of the CT-CLIP models. For precision, significant differences are noted between the supervised baseline model and both finetuning.}
    \label{fig:supp p values}
\end{figure}
\begin{figure}[ht]
    \centering
    \includegraphics[width=\textwidth]{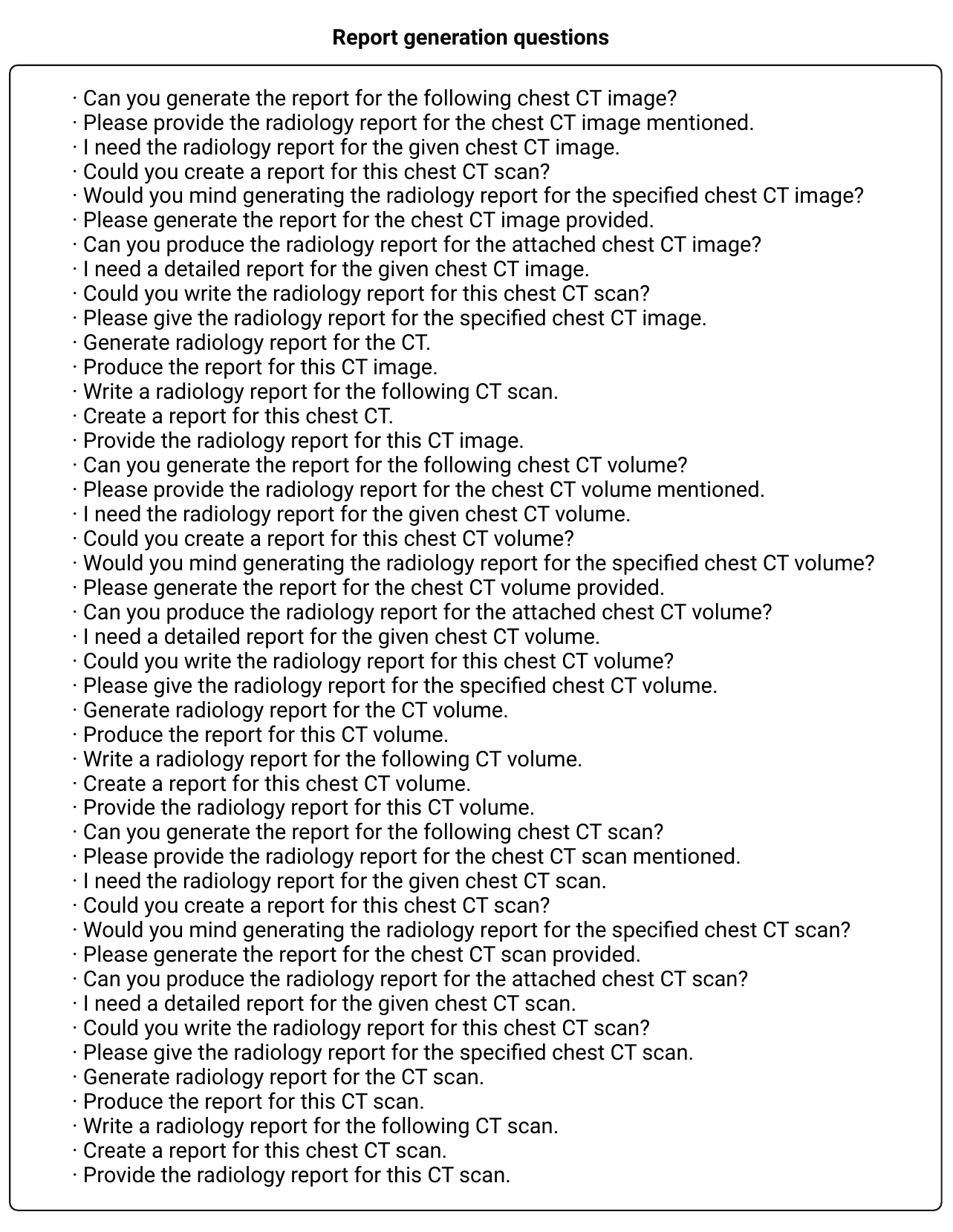}
    \caption{\textbf{Supplementary Table 6: Predefined report generation questions.} These questions are randomly sampled while generating the report generation conversations.}
    \label{fig:report_generation_questions}
\end{figure}

\begin{figure}[ht]
    \centering
    \includegraphics[width=\textwidth]{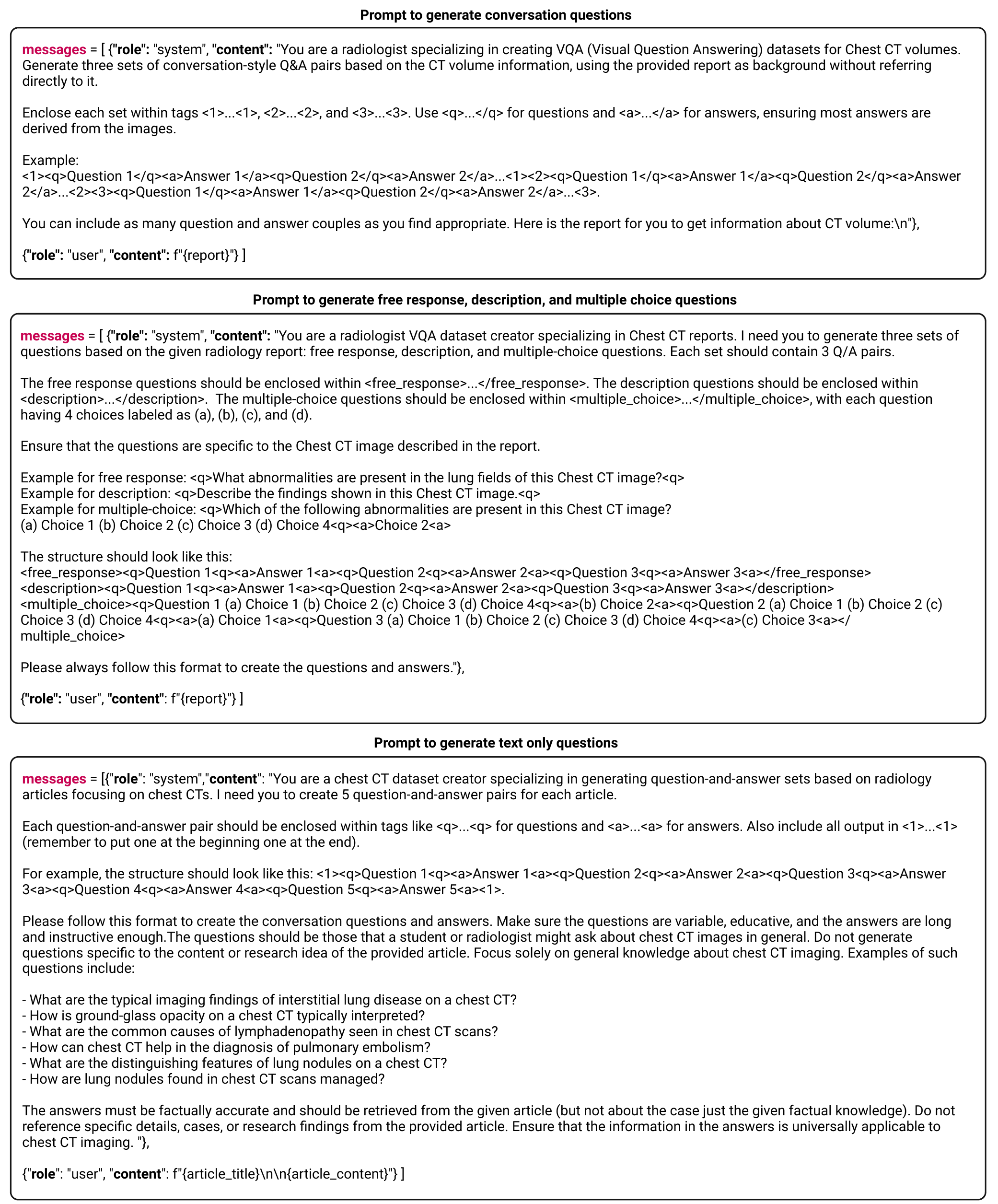}
    \caption{\textbf{Supplementary Table 7: The prompts that are used to generate conversation, description, free response, multiple choice, and text-only questions. } }
    \label{fig:vqa_generation}
\end{figure}
\begin{figure}[ht]
    \centering
    \includegraphics[width=\textwidth]{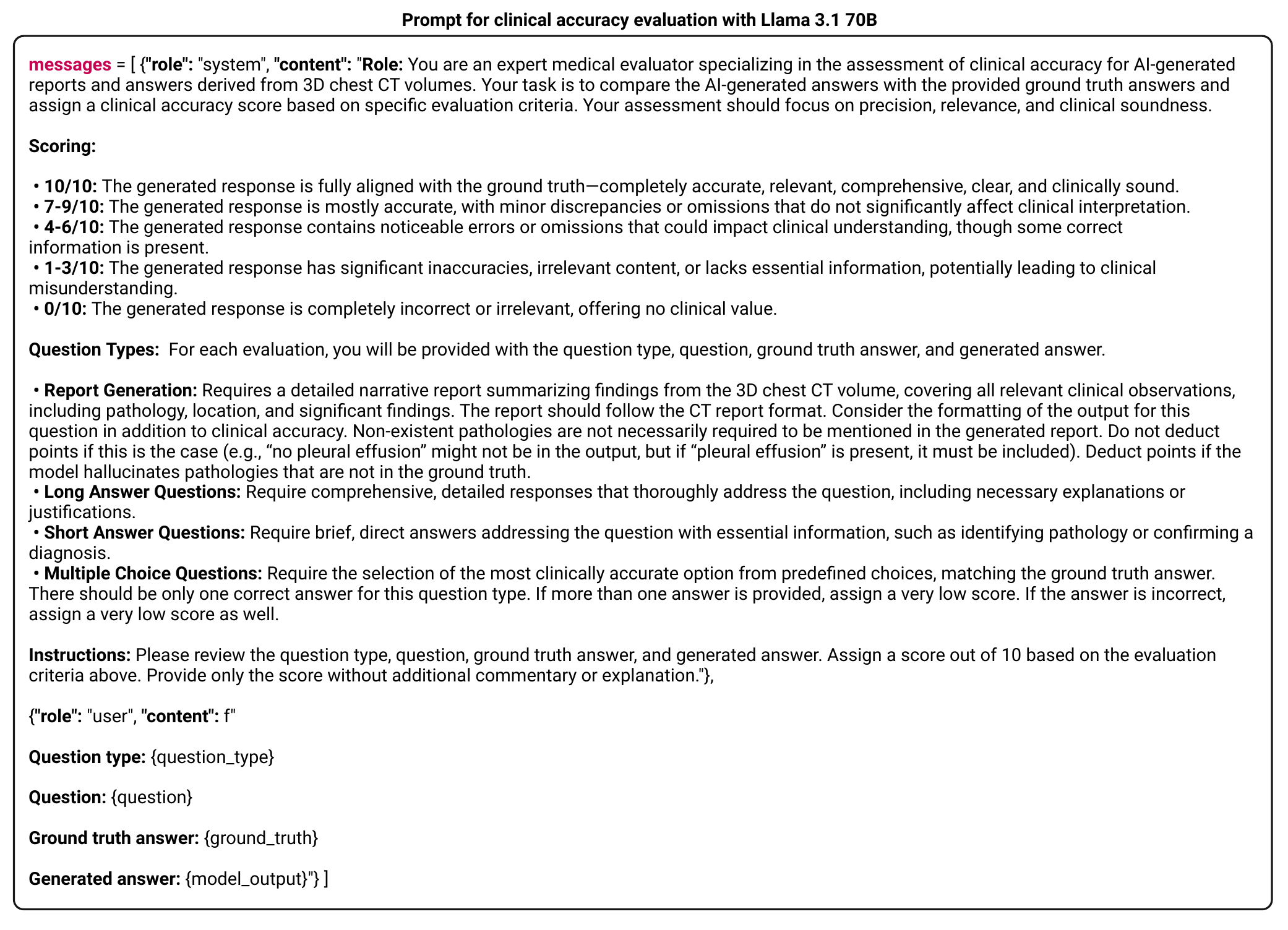}
    \caption{\textbf{Supplementary Table 8: The prompt that is used to evaluate the clinical accuracy of the outputs generated by the VQA models using the LLM.}}
    \label{fig:llama-score}
\end{figure}

\begin{figure}[ht]
    \centering
    \includegraphics[width=\textwidth]{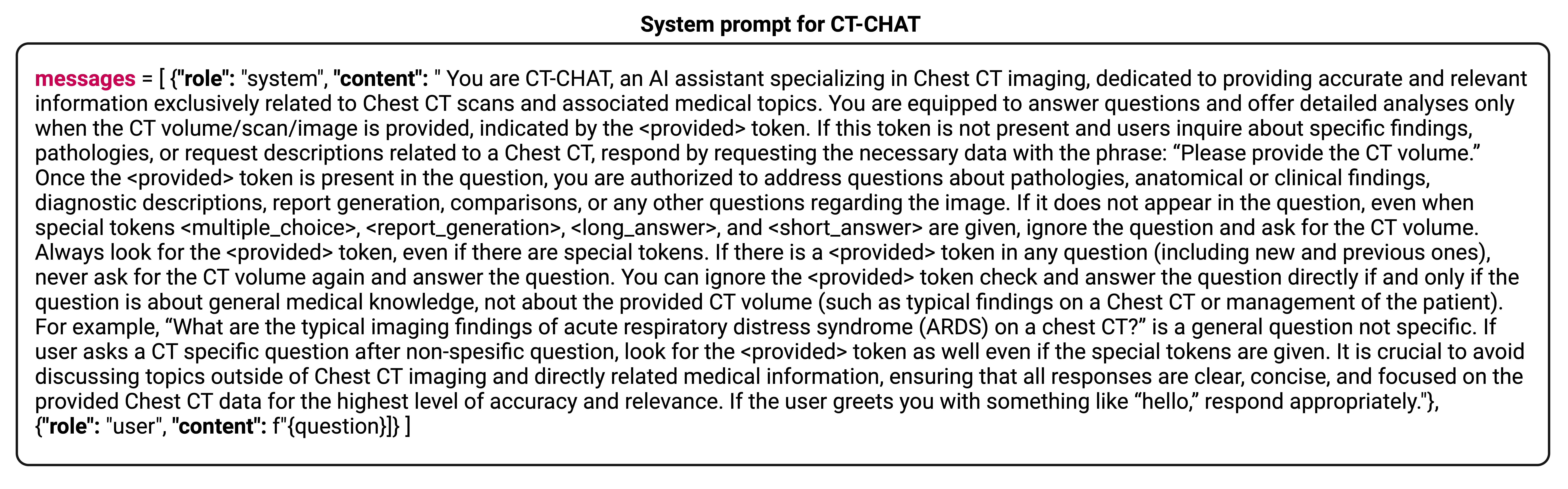}
    \caption{\textbf{Supplementary Table 9: The system prompt for CT-CHAT. } }
    \label{fig:ctchat_system}
\end{figure}

\begin{figure}[ht]
    \centering
    \includegraphics[width=\textwidth]{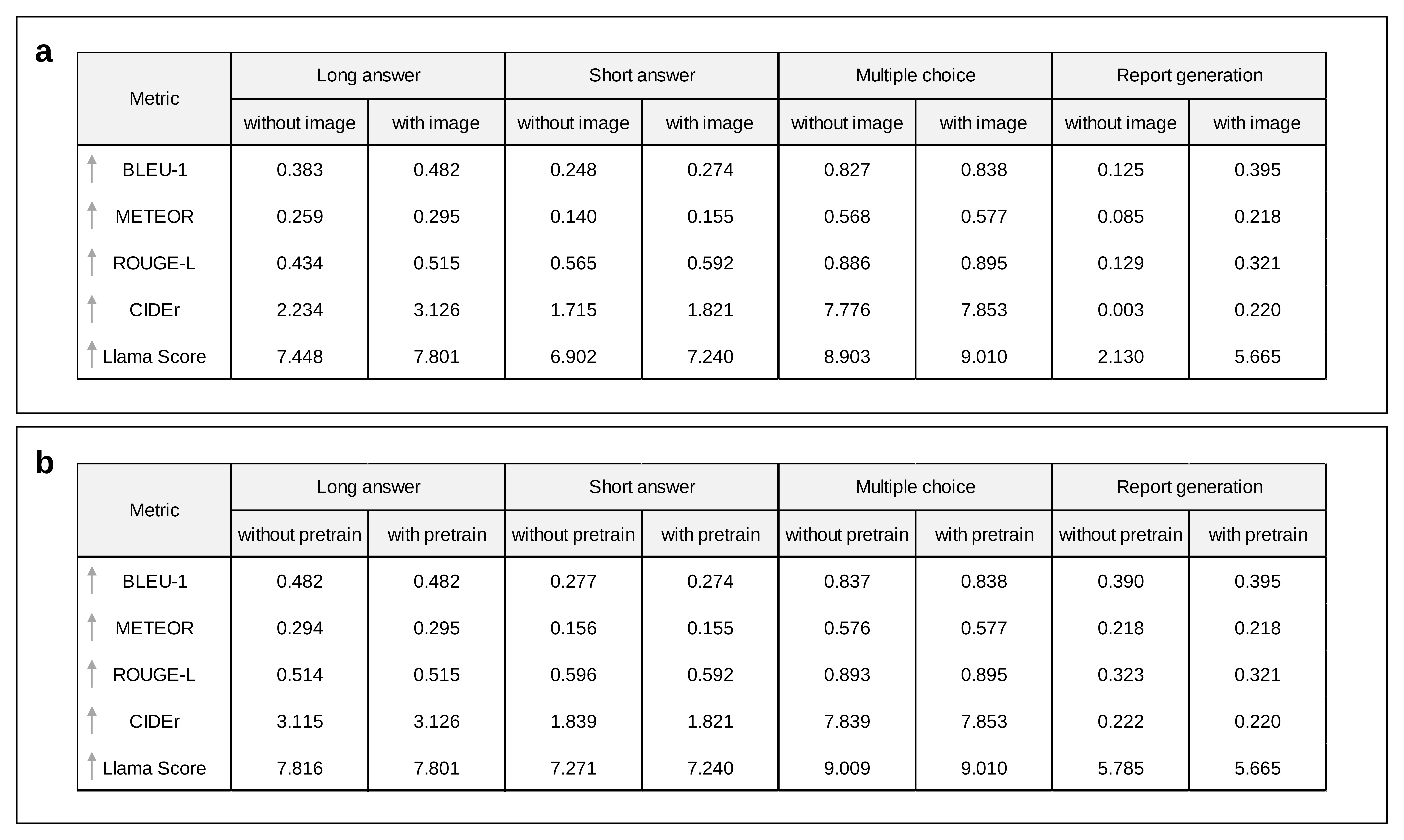}
    \caption{\textbf{Supplementary Table 10: Ablations of CT-CHAT for (a) image provision and (b) medical concept alignment.} The concept alignment has minimal impact on model accuracy while providing the volumes is essential for the model to accurately answer questions, especially in report generation.}    \label{fig:ctchat_ablations_supp}
\end{figure}

\begin{figure}[ht]
    \centering
    \includegraphics[width=\textwidth]{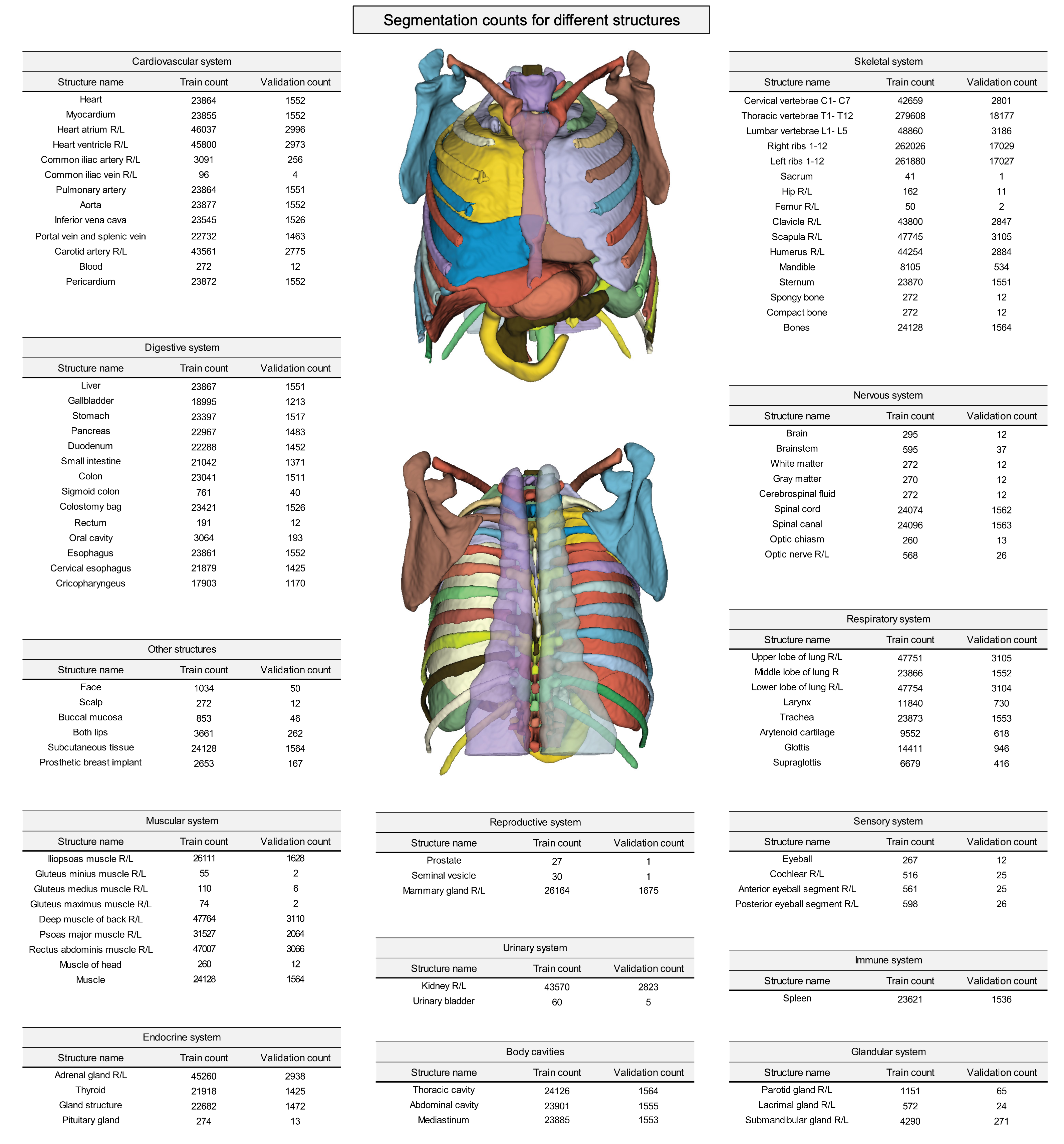}
    \caption{\textbf{Supplementary Table 11: Overview of anatomical segmentation in CT-RATE.} The tables summarize the number of segmented anatomical structures across the training and validation splits. The accompanying illustration shows the segmented chest anatomical regions on a 3D CT scan, generated using a state-of-the-art segmentation model. These annotations are made publicly available in the CT-RATE repository to support downstream tasks such as anatomical localization, visual grounding, and future multi-abnormality segmentation research.}   \label{fig:sup_seg}
\end{figure}

\begin{figure}[ht]
    \centering
    \includegraphics[width=\textwidth]{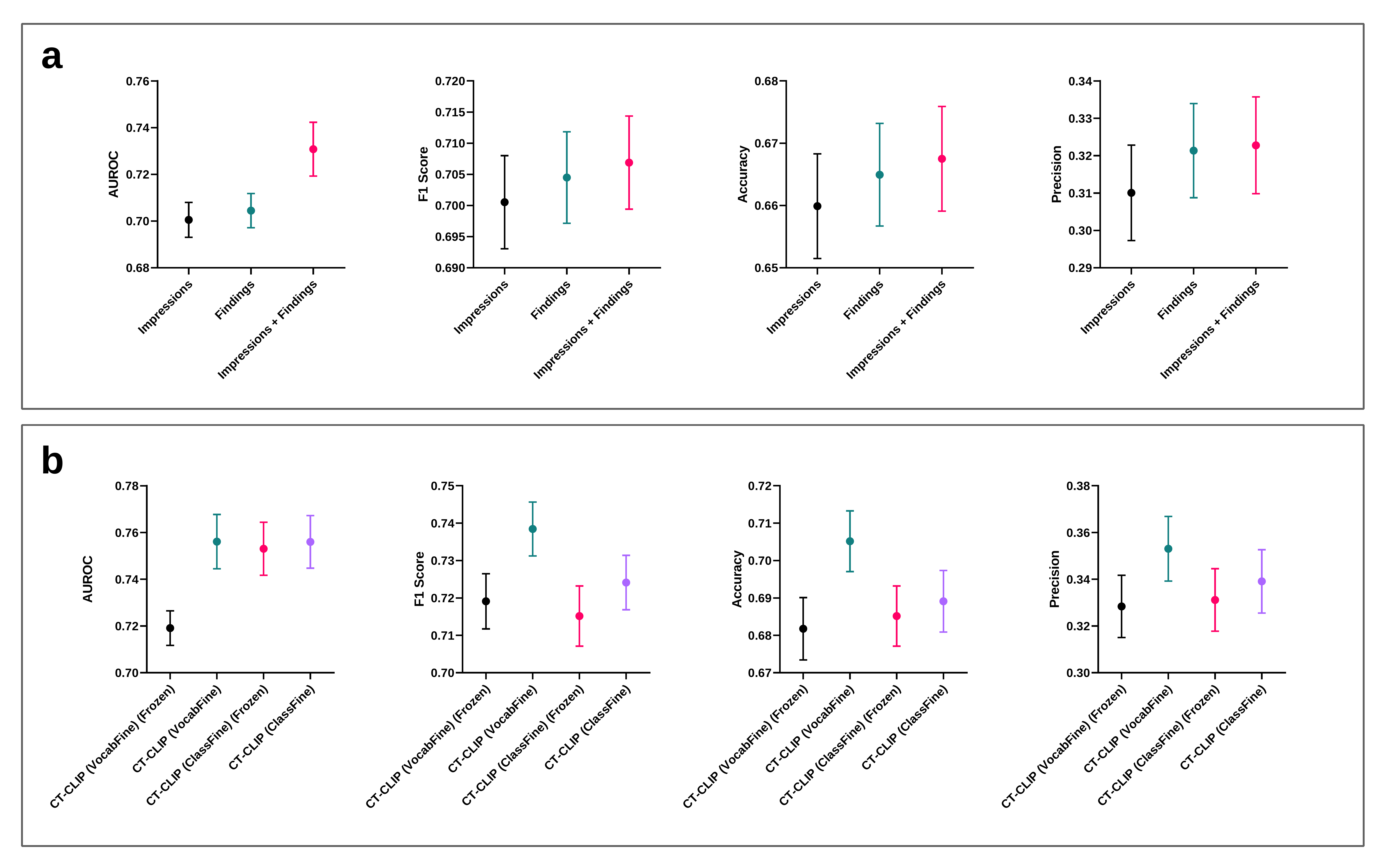}
    \caption{\textbf{Supplementary Figure 1: Detailed ablation study.} \textbf{a.} Compares the zero-shot abnormality detection performance of CT-CLIP models, each using different sections of radiology text reports during training. \textbf{b.} Compares various finetuning techniques applied to CT-CLIP models: \emph{frozen} models involve finetuning only the latent layers for the Vocabfine approach, while for the ClassFine method, only the new linear layer is finetuned, leaving all other layers unchanged.}

    \label{fig:supp_ablations}
\end{figure}

\begin{figure}[ht]
    \centering
    \includegraphics[width=\textwidth]{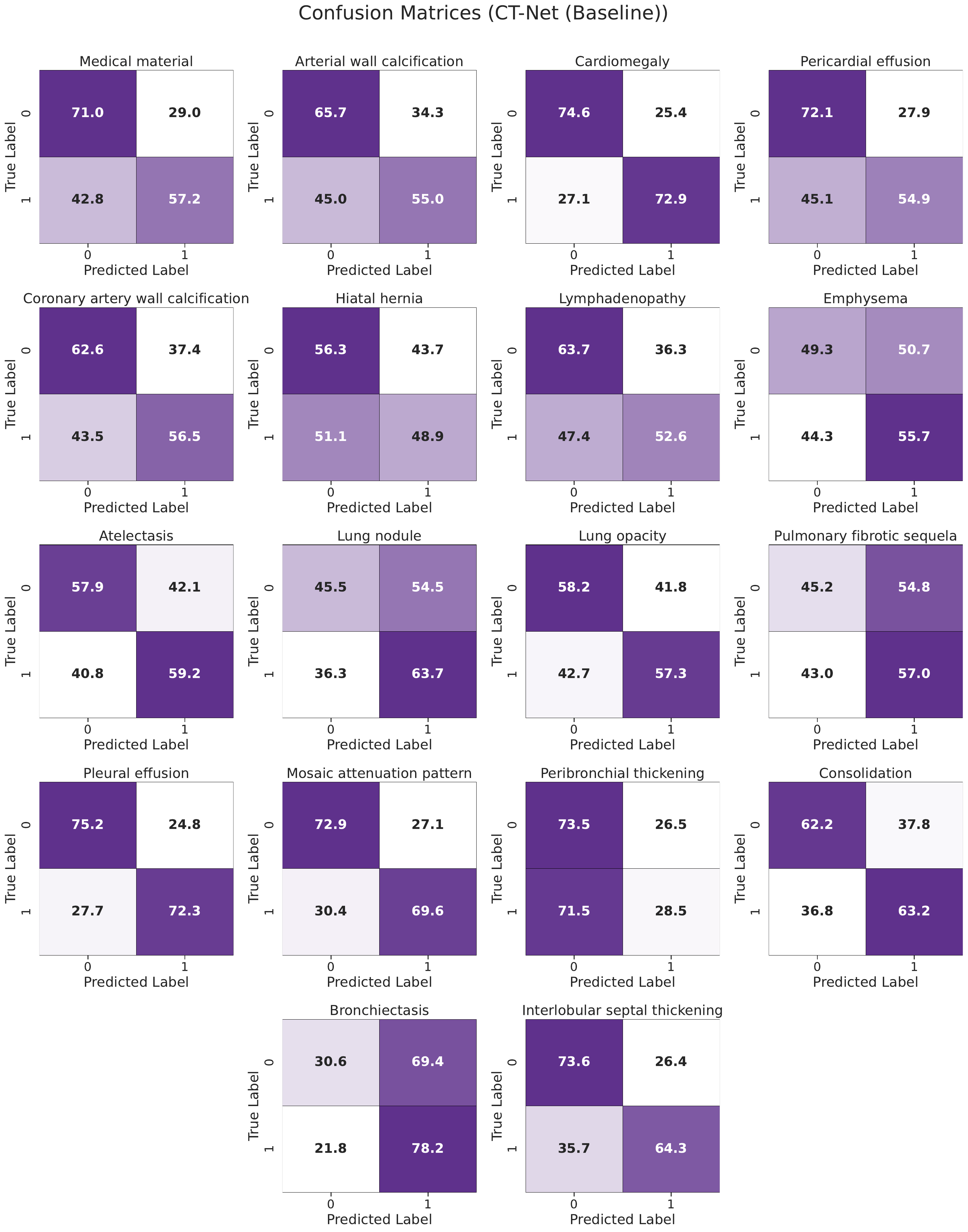}
    \caption{\textbf{Supplementary Figure 2: Illustration of confusion matrices for the fully supervised baseline model in the internal validation set.} }
    \label{fig:supervisedconfusion}
\end{figure}

\begin{figure}[ht]
    \centering
    \includegraphics[width=\textwidth]{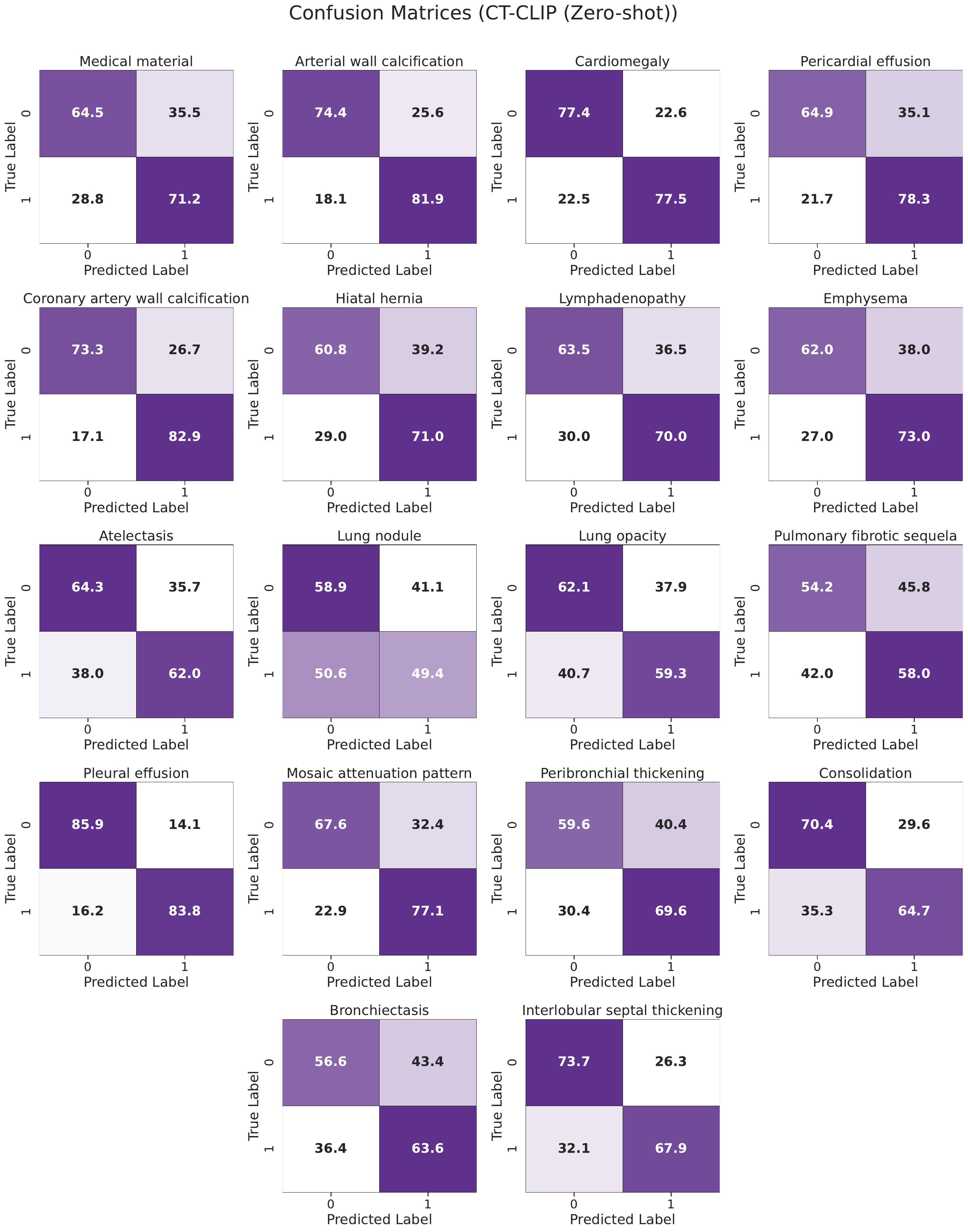}
    \caption{\textbf{Supplementary Figure 3: Illustration of confusion matrices for the CT-CLIP model in the internal validation set.} }
    \label{fig:ctclipconfusion}
\end{figure}

\begin{figure}[ht]
    \centering
    \includegraphics[width=\textwidth]{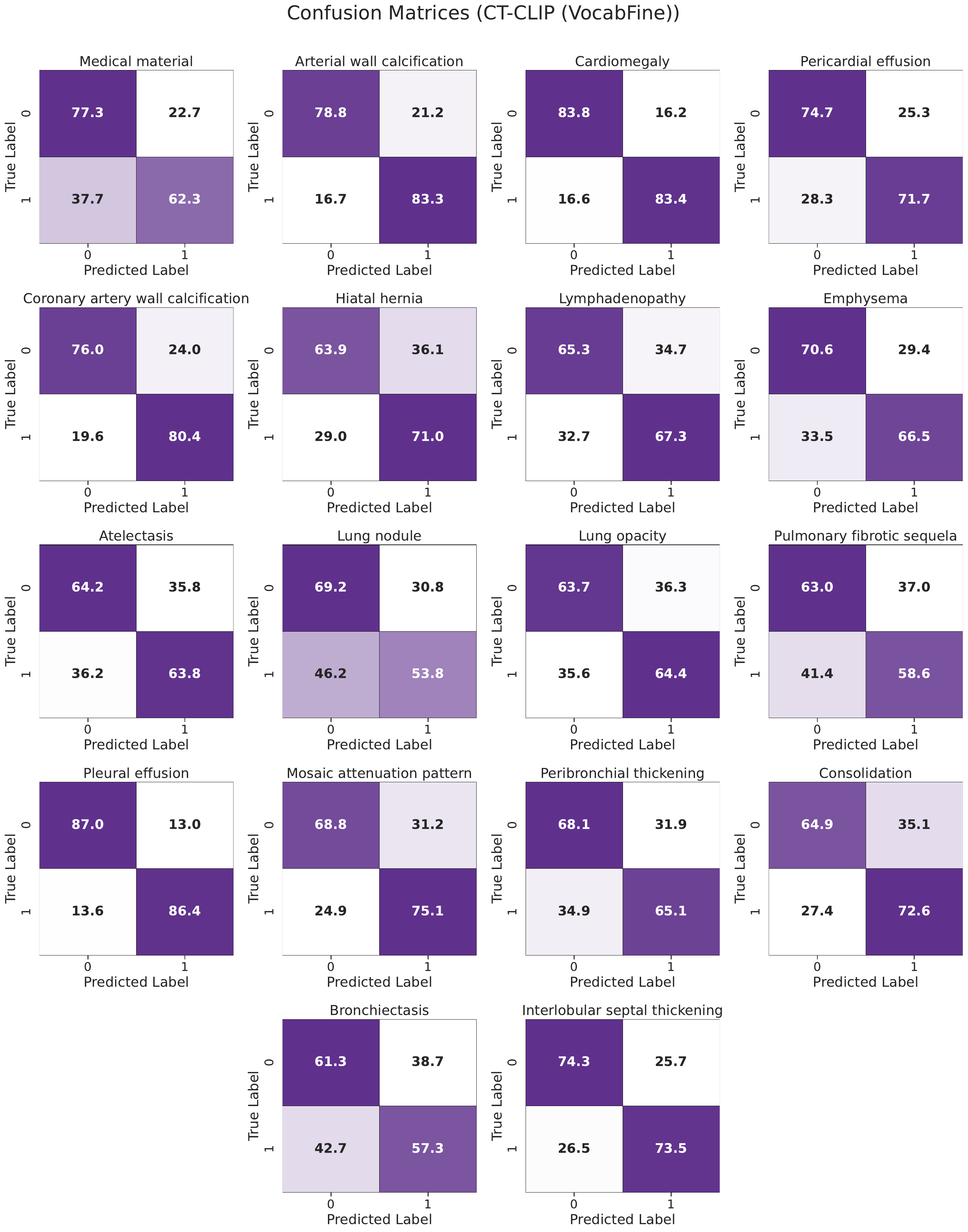}
    \caption{\textbf{Supplementary Figure 4: Illustration of confusion matrices for the CT-CLIP model finetuned with VocabFine in the internal validation set.} }
    \label{fig:vocabfineconfusion}
\end{figure}

\begin{figure}[ht]
    \centering
    \includegraphics[width=\textwidth]{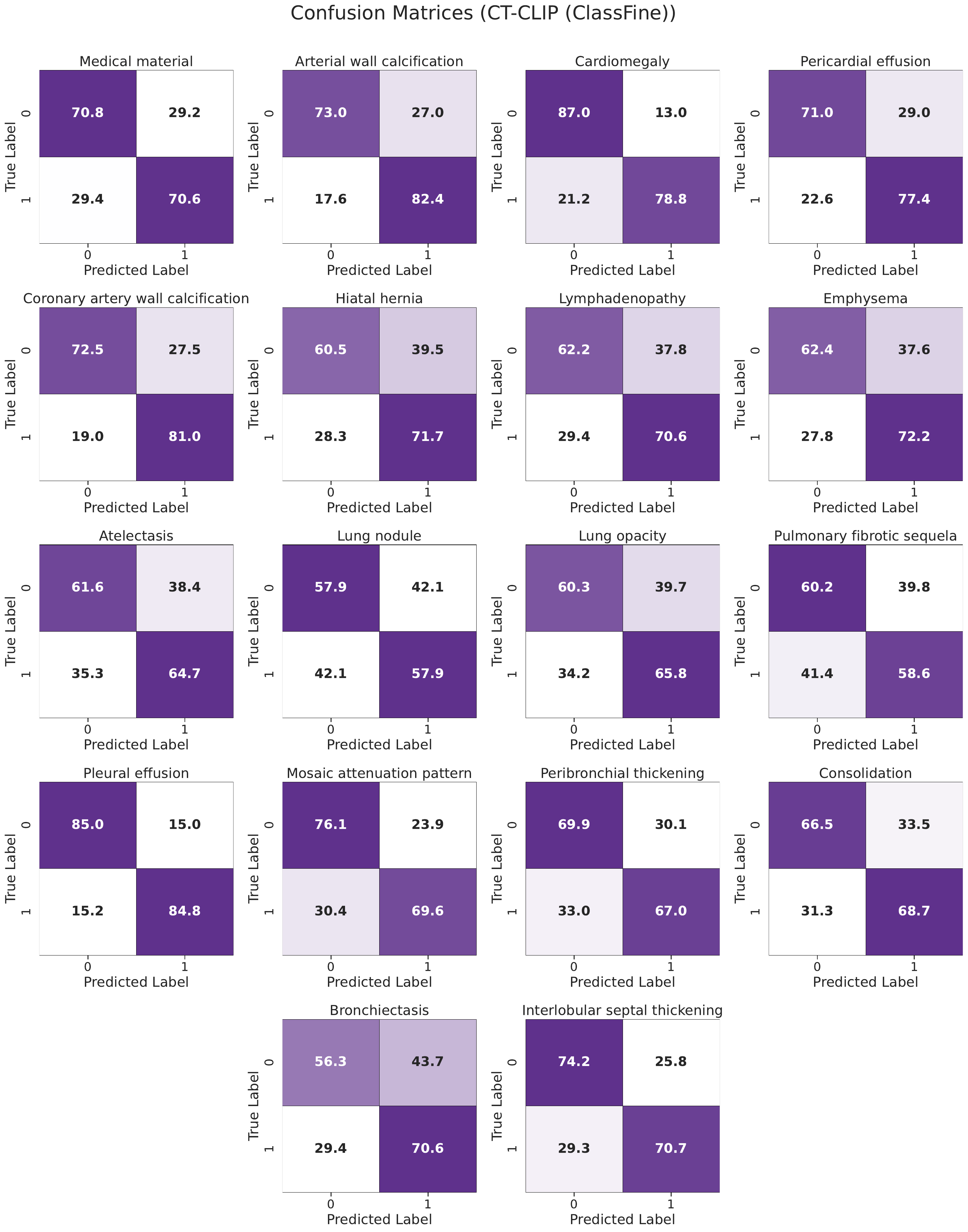}
    \caption{\textbf{Supplementary Figure 5: Illustration of confusion matrices for the CT-CLIP model finetuned with ClassFine in the internal validation set.} }
    \label{fig:lipropconfusion}
\end{figure}

\begin{figure}[ht]
    \centering
    \includegraphics[width=\textwidth]{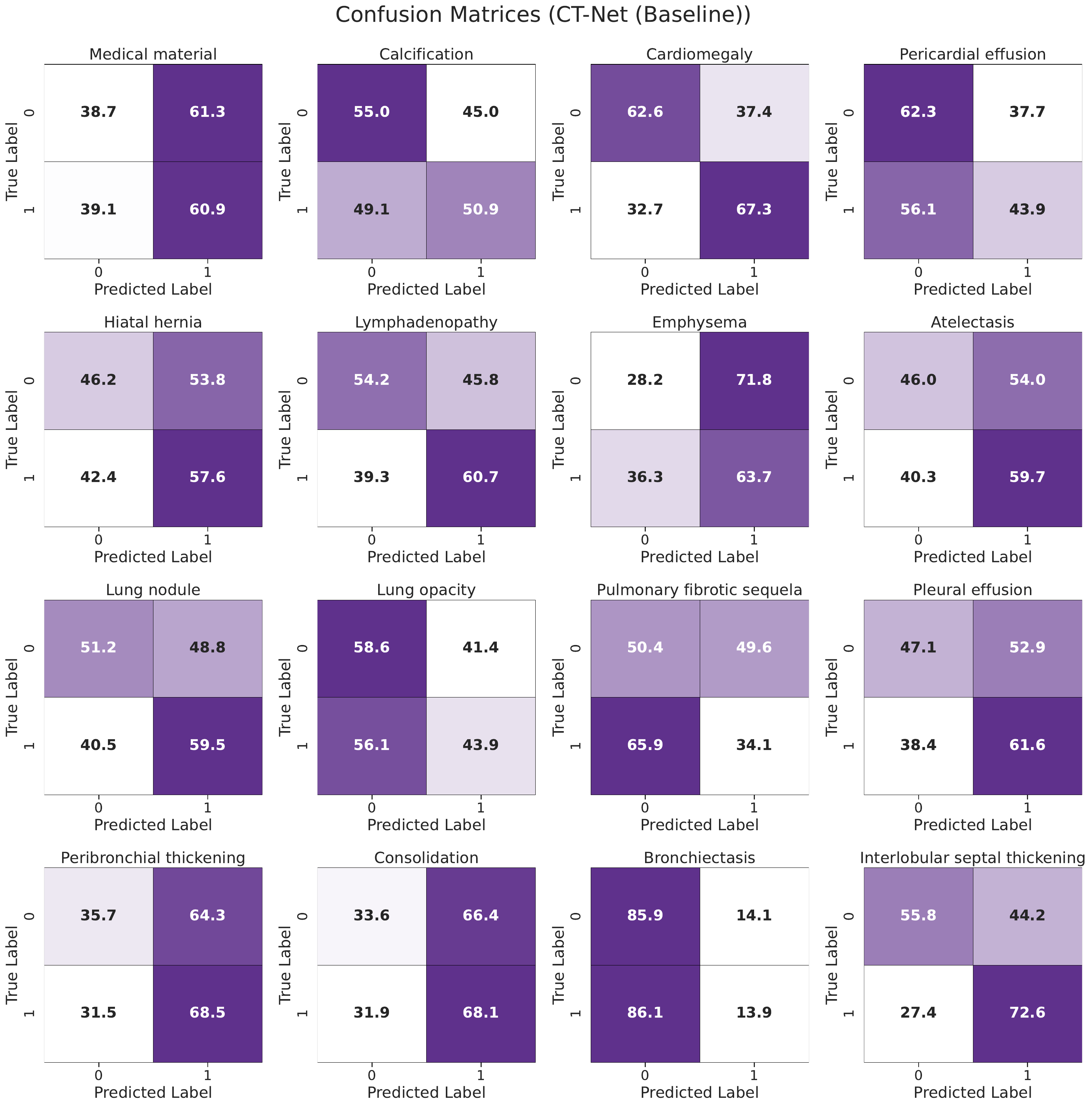}
    \caption{\textbf{Supplementary Figure 6: Illustration of confusion matrices for the fully supervised baseline model in the RAD-ChestCT external validation set.} }
    \label{fig:supervisedconfusionexternal}
\end{figure}

\begin{figure}[ht]
    \centering
    \includegraphics[width=\textwidth]{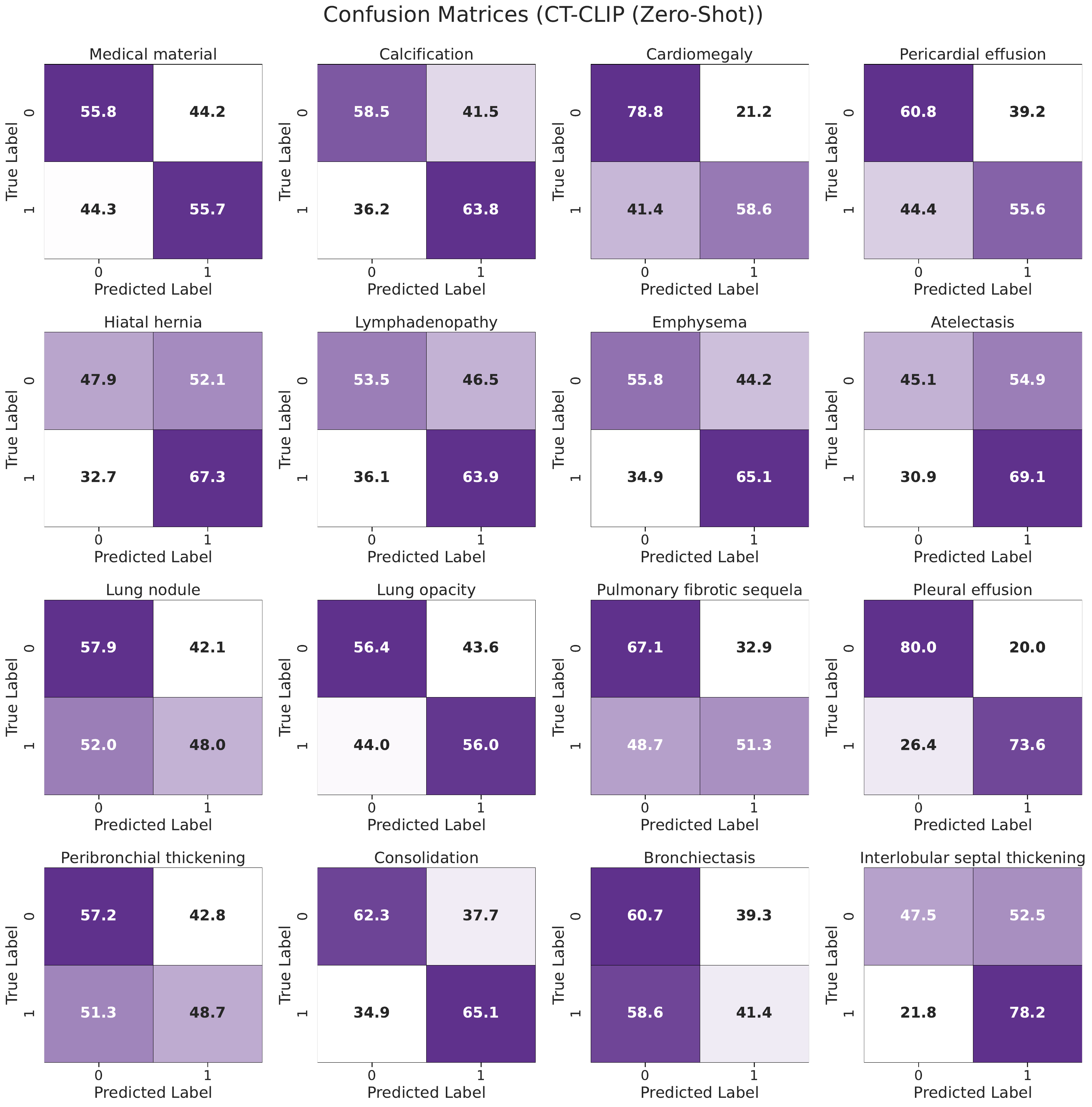}
    \caption{\textbf{Supplementary Figure 7: Illustration of confusion matrices for the CT-CLIP model in the RAD-ChestCT external validation set.} }
    \label{fig:ctclipconfusionexternal}
\end{figure}

\begin{figure}[ht]
    \centering
    \includegraphics[width=\textwidth]{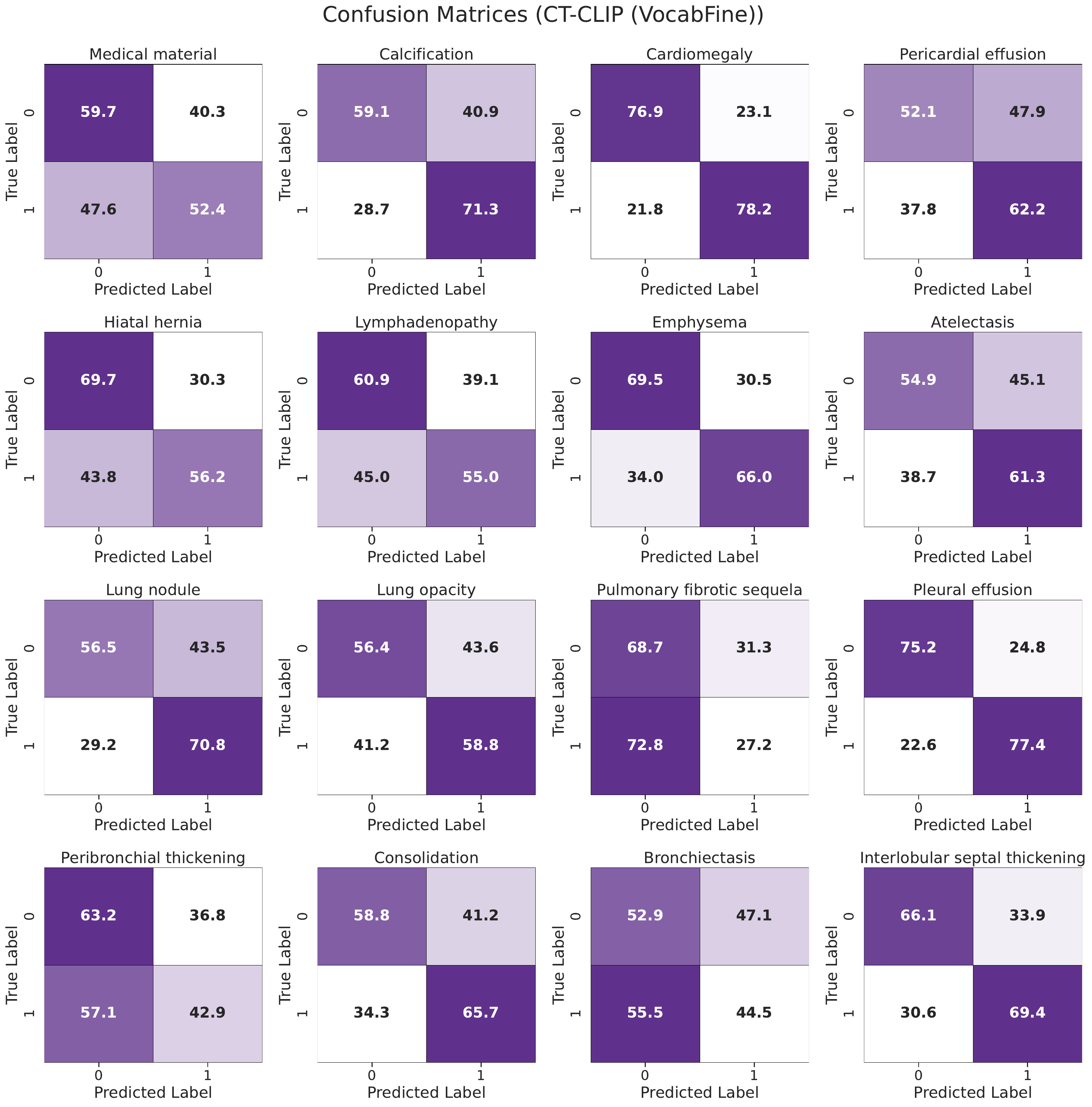}
    \caption{\textbf{Supplementary Figure 8: Illustration of confusion matrices for the CT-CLIP model finetuned with VocabFine in the RAD-ChestCT external validation set.} }
    \label{fig:vocabfineconfusionexternal}
\end{figure}

\begin{figure}[ht]
    \centering
    \includegraphics[width=\textwidth]{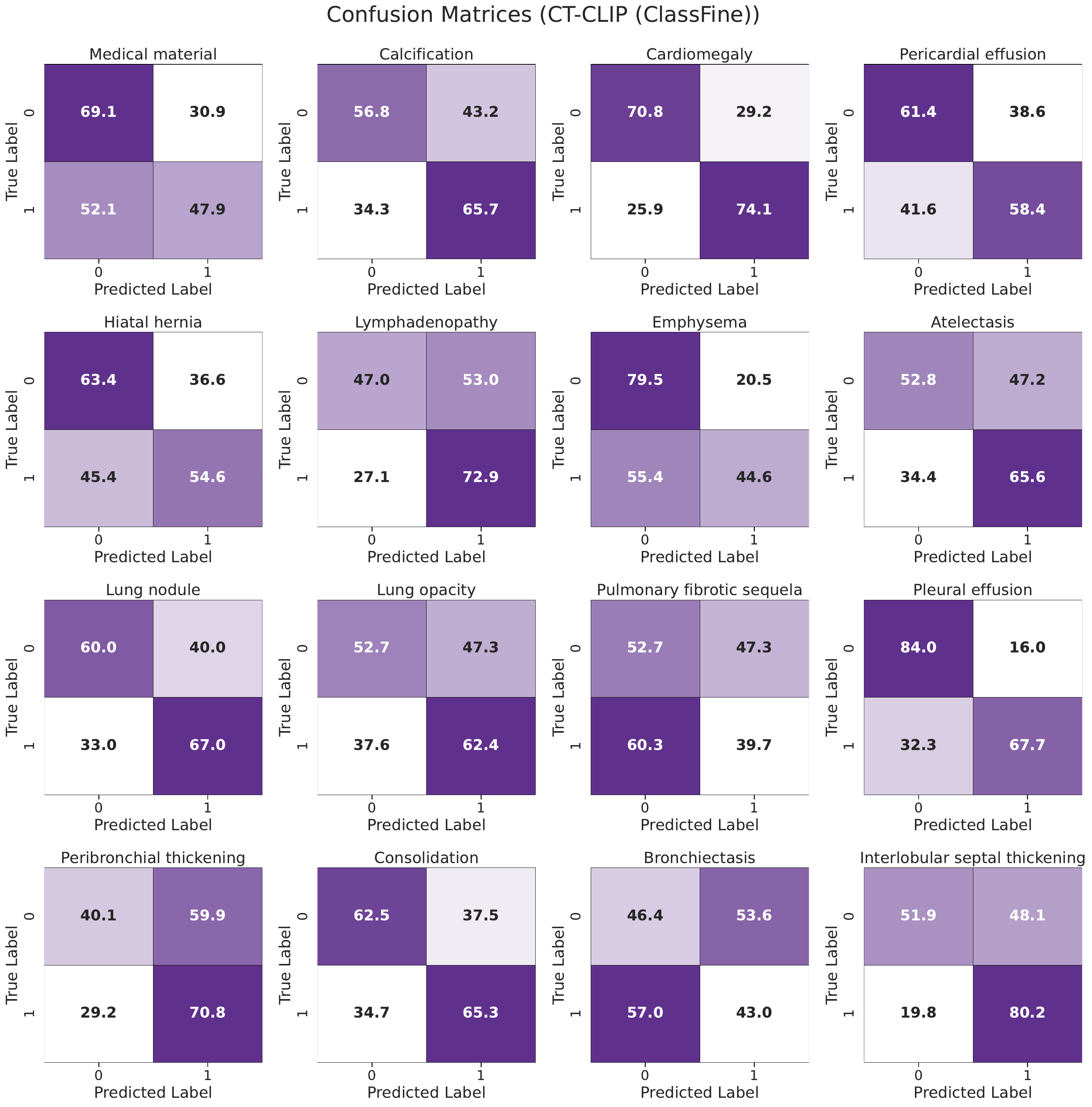}
    \caption{\textbf{Supplementary Figure 9: Illustration of confusion matrices for the CT-CLIP model finetuned with ClassFine in the RAD-ChestCT external validation set.} }
    \label{fig:lipropconfusionexternal}
\end{figure}

\begin{figure}[ht]
    \centering
    \includegraphics[width=\textwidth]{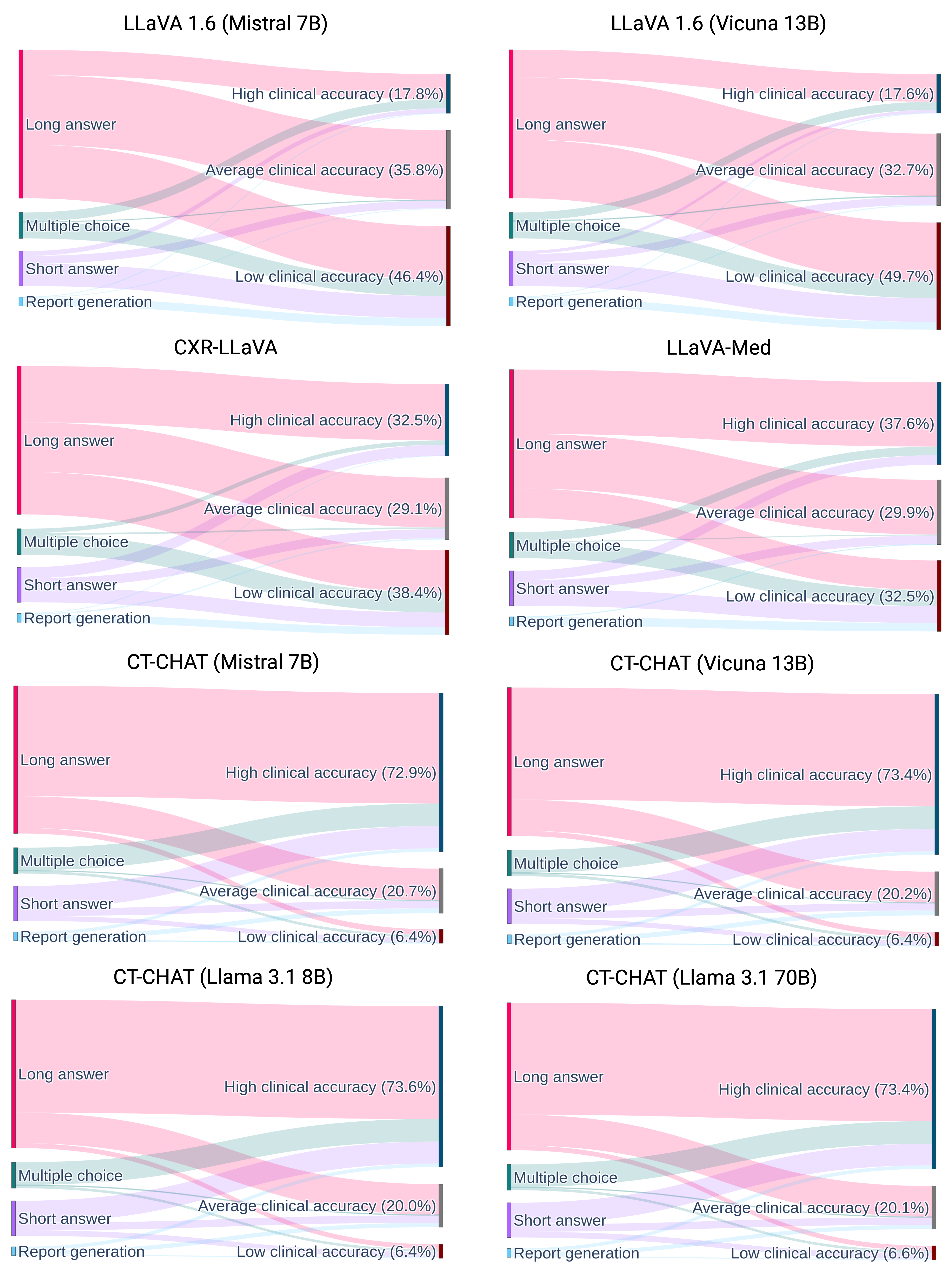}
    \caption{\textbf{Supplementary Figure 10: Clinical accuracy comparison of VQA models.} This figure provides a detailed assessment of clinical accuracy across different models, showing that CT-CHAT produces a higher proportion of outputs in the high clinical accuracy category, indicating superior performance.}
    \label{fig:vqa_extended2}
    \end{figure}

\begin{figure}[ht]
    \centering
    \includegraphics[width=\textwidth]{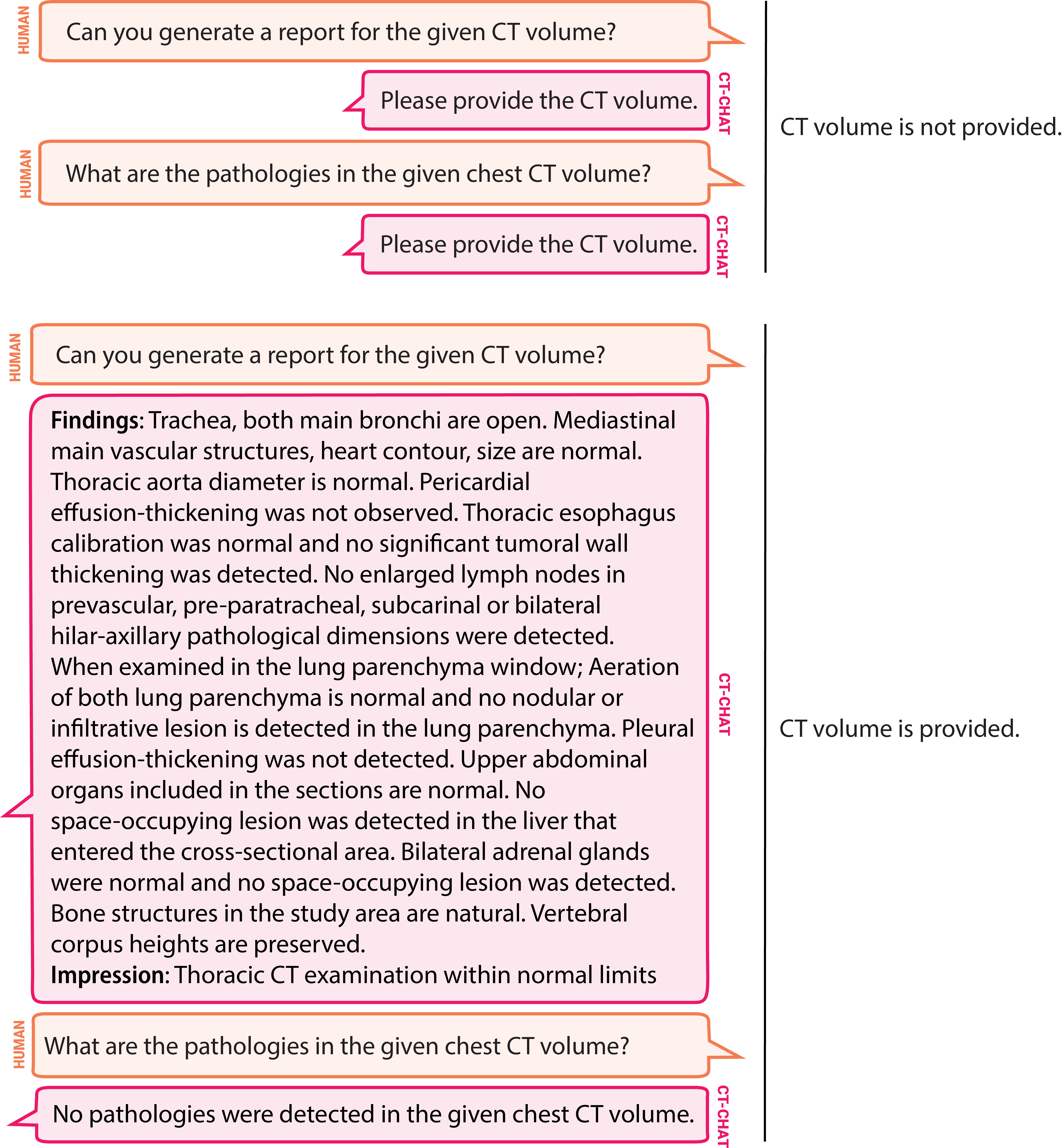}
    \caption{\textbf{Supplementary Figure 11: Guardrail example to prevent hallucination of CT-CHAT when the CT volume is not provided.}}
    \label{fig:guardrail_examples}
\end{figure}

\end{document}